\documentclass[letterpaper, 10 pt, conference]{ieeeconf}  %

\IEEEoverridecommandlockouts                              %

\usepackage{graphicx}
\usepackage{amsmath}
\usepackage{amssymb}
\usepackage{booktabs}

\usepackage[dvipsnames]{xcolor,colortbl}
\usepackage{threeparttable}
\usepackage{tabularx}
\usepackage{multirow}

\usepackage{textcomp}
\usepackage{bm}
\usepackage{mathrsfs}
\usepackage{times}
\usepackage{epsfig}

\usepackage[normalem]{ulem}
\usepackage{outlines}
\usepackage{multirow}
\usepackage{ifthen}
\usepackage{pifont}
\usepackage[noadjust]{cite}
\usepackage[caption=false]{subfig}
\usepackage[pagebackref,breaklinks,colorlinks=true, citecolor=red, urlcolor=RoyalBlue]{hyperref}

\usepackage[capitalize]{cleveref}
\crefname{section}{Sec.}{Secs.}
\Crefname{section}{Section}{Sections}
\Crefname{table}{Table}{Tables}
\crefname{table}{Tab.}{Tabs.}

\newcommand{\ourmodel}{\textsc{GoRela}}

\newcommand{\x}{\bm{x}}
\newcommand{\e}{\bm{e}}
\newcommand{\g}{\bm{g}}

\newcommand{\cmark}{\ding{51}}%
\newcommand{\xmark}{\ding{55}}%
\newcolumntype{s}{>{\raggedleft\arraybackslash}X}
\title{GoRela: Go Relative for Viewpoint-Invariant Motion Forecasting}

\author{
\textbf{Alexander Cui, Sergio Casas, Kelvin Wong, Simon Suo, Raquel Urtasun} \\
Waabi, University of Toronto \\
\texttt {\{acui, sergio, kwong, ssuo, urtasun\}@waabi.ai}
}

\begin{document}

\maketitle
\thispagestyle{empty}
\pagestyle{empty}

\begin{abstract}
    The task of motion forecasting %
    is critical for self-driving vehicles (SDVs) to be able to plan a safe maneuver. Towards this
    goal, modern approaches 
    reason about the map, the agents' past trajectories and their interactions in order to produce
    accurate forecasts. 
    The predominant approach has been to encode the  map and other agents in the reference frame of
    each target agent. However, this approach is computationally
    expensive for multi-agent prediction as inference needs to be run for each agent. 
    To tackle the scaling challenge, the solution thus far has been to encode all agents
    and the map in a shared coordinate frame (e.g., the
    SDV frame). However, this is sample inefficient and vulnerable to domain shift (e.g., when the SDV visits uncommon states). 
    In contrast, in this paper, we propose an efficient shared encoding for all agents and the map without sacrificing accuracy or generalization. 
    Towards this goal, we leverage 
    pair-wise relative positional encodings to represent geometric relationships between
    the agents and the map elements in a heterogeneous spatial graph. 
    This parameterization allows us to be invariant to scene
    viewpoint, and save online computation by re-using map embeddings computed offline. 
    Our decoder is also viewpoint agnostic, predicting agent goals on the lane graph to enable
    diverse and context-aware multimodal prediction. We demonstrate the effectiveness of our approach on the urban Argoverse 2 benchmark as well as a novel highway dataset. %
\end{abstract}

\section{Introduction}

Predicting the future motion of the  traffic participants is critical for SDVs. To drive safely,
predictions have to be not only accurate and generalize across many scenarios, but also made in a timely manner so that the SDV can react appropriately.

Existing methods have made compromises in their accuracy and generalization abilities vs. computation needs. Most approaches \cite{janjovs2022starnet,gao2020vectornet,cui2018multimodal} have prioritized accuracy and generalization at the expense of
runtime by processing the scene from the viewpoint of each target agent. Unfortunately this
approach  does not scale to situations with a large number of agents, which arise in crowded urban
scenes or in  highways  where, due to the agents' high speed,  predictions need to be
performed over a very large area, thus potentially containing many agents. 
Other works \cite{ngiam2021scene,gilles2021thomas,zeng2021lanercnn} have focused on achieving a reasonable inference time at the expense of less
accurate models that are less capable of generalization. These works propose a shared encoding of
the scene for all agents by using a fixed viewpoint for all predictions such as the coordinate frame defined by the
SDV's current pose. 
While achieving low latency, the predictions are no longer invariant to the 
viewpoint from which the scene is encoded, and as a consequence may not generalize to rarely seen or
novel SDV poses (e.g., when the SDV is performing a U-turn).

\begin{figure}[t]
    \centering
    \includegraphics[width=\columnwidth]{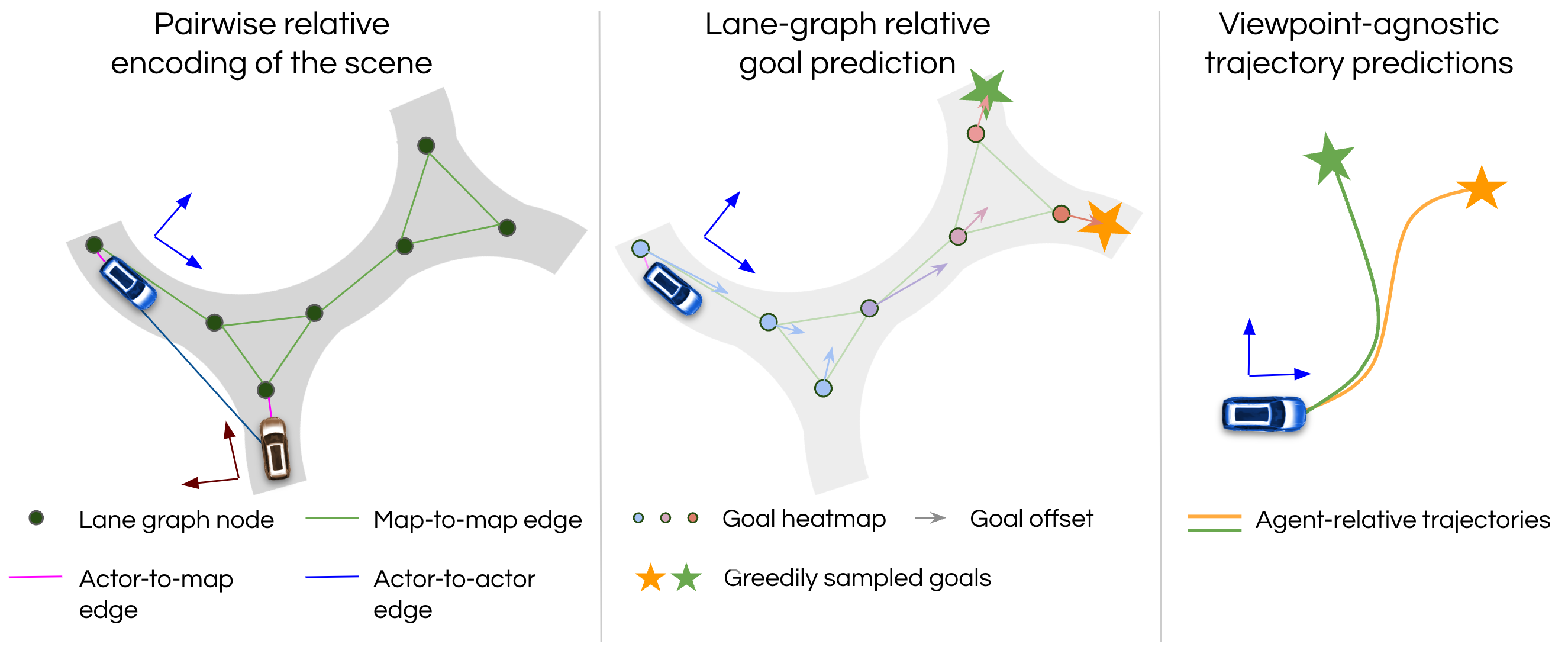}
    \vspace{-20pt}
    \caption{ Our model, \ourmodel{}, encodes the scene by reasoning about pair-wise relative geometric relationships between the agents and the lane graph. Then, it
        predicts a goal distribution using the lane graph nodes as anchors. Finally, a
        trajectory is generated conditioned on the goal. The architecture is viewpoint-invariant, thus allowing for shared computation across agents.}
    \vspace{-15pt}
    \label{fig:motivating}
\end{figure}

In this paper, we propose to encode the interaction between different map entities and the
agents in a viewpoint agnostic representation by modelling  their pairwise relative relationships.
This viewpoint invariance provides several key advantages: (i) it helps the model generalize by greatly
reducing the space of the problem domain, (ii) it makes learning more sample efficient, making
training faster while  requiring less data (e.g., removing the need for any data
augmentation during training in the form of viewpoint translation and rotation), (iii) it keeps inference highly resource-efficient as the
scene encoding (which is the heaviest module) only needs to be processed once, and (iv) it enables caching
of any computation that depends solely on the static parts of the scene, allowing our model to compute
the map embeddings offline, thus saving critical processing time
 on the road. 
Through extensive quantitative evaluation on both urban and highway datasets, we demonstrate that
our model is more effective, generalizes better to novel viewpoints, and it is less data hungry than competing methods. %

We structure the paper as follows: Section \ref{sec:related} offers a survey of motion forecasting methods, Section \ref{sec:hmp} introduces a new pair-wise relative heterogeneous graph neural network that is the backbone for interaction reasoning in our motion forecasting model, presented in Section \ref{sec:pipeline}. Finally, we benchmark our approach in Section \ref{sec:exps}.

\section{Related Work in Motion Forecasting}
\label{sec:related}

In this section we review prior approaches, breaking down their contributions to several motion forecasting modules.

\vspace{0.2cm}
\noindent{\bf Agent history encoding:}
Most works represent the past trajectory as a sequence of 2D waypoints in a global frame. Rasterizing the past trajectory and feeding it through a 2D convolutional
neural network (CNN) was first proposed \cite{cui2018multimodal,chai2019multipath}. Then, fully
vectorized methods \cite{lgn,gao2020vectornet,gilles2021thomas} proposed to transform all past poses
to a common coordinate frame and process them with 1D CNNs, recurrent neural networks (RNN) or
Transformers. 

\vspace{0.2cm}
\noindent{\bf Map encoding:}
Early works encode the map using bird's-eye view (BEV) rasters and 2D CNNs \cite{cui2018multimodal,casas2018intentnet, bansal2018chauffeurnet,Hong_2019_CVPR,
chou2018arxiv,zeng2019end}. However, this design is poorly suited
when the SDV heading is not aligned with the road as the encoding is not rotationally invariant. In
self-driving datasets, the SDV heading will generally align with its lane, and thus heading offsets
will make the inputs out of distribution (OOD) regardless of the behavior of
other agents. Processing a region of interest (RoI) centered and rotated around each agent
\cite{chai2019multipath,casas2020spagnn,casas2020implicit,cui2021lookout,casas2021mp3,zeng2020dsdnet,li2020end}
mitigates this issue, but makes inference computationally demanding as these RoIs need to be large.
Recently, vector-based approaches represent the map as a set of lanes and nodes,
employing graph neural networks (GNNs) or Transformers to fuse map features with the agents. To
encode the scene context,
\cite{gao2020vectornet,lgn,varadarajan2022multipath,gilles2022gohome,mo2022multi}
transform the map elements' coordinates to the target agent frame.
To enable multi-agent prediction while being invariant to SDV pose, many follow \cite{janjovs2022starnet} and repeat
the scene encoding for each agent, linearly increasing the cost as the number of agents grow.
To tackle the linear growth in computational cost,
\cite{gilles2021thomas,casas2020implicit,cui2021lookout,casas2019spatially,zeng2021lanercnn,wang2022ltp}
 encode the scene in a global frame shared by all agents,
but lose viewpoint invariance, thus sacrificing accuracy and generalization and making the model more data hungry. 
In contrast, our lane-graph encoder achieves a shared, viewpoint-invariant scene encoding.

\vspace{0.2cm}
\noindent{\bf Heterogeneous fusion:}
To fuse information between agents and map, LaneGCN\cite{lgn} defines a late fusion where information is propagated in the following order:
agent-to-map, map-to-map, map-to-agent, agent-to-agent. LaneRCNN\cite{zeng2021lanercnn} instead early fuses agents' past trajectories into the lane-graph nodes. 
In an attempt to
simplify the pipeline, SceneTransformer \cite{ngiam2021scene} regards agent waypoints and
lane-graph nodes as tokens and performs multiple rounds of cross-attention. 
Similarly to the map encoding, a shared coordinate frame is assumed in order to process the scene only
once, at the expense of losing invariance to the viewpoint. 
Instead, our model leverages edge attributes in a heterogeneous graph, where geometric
information is expressed as pair-wise relative positional encodings, thus being viewpoint invariant
while maintaining the desired shared scene processing across agents. As far as we know, concurrent work HDGT\cite{jia2022HDGT} is the only other method to also leverage a pair-wise relative encoding. 
However, our work presents several advantages: 
(i) our bounded pair-wise relative positional encoding removes the need for many graph creation heuristics,
(ii) our goal-based trajectory decoder is more powerful than direct regression (as shown by our ablations), and
(iii) it enables map embedding caching due to its late fusion of agents and map, thus reducing model latency.
These differences contribute to a superior performance in Argoverse 2 w.r.t. HDGT
\cite{jia2022HDGT}. %

\vspace{0.2cm}
\noindent{\bf Future trajectory decoding:}
Regressing all the parameters of a mixture of future trajectories 
in parallel \cite{casas2019spatially,chai2019multipath,lgn,jia2022HDGT,mo2022multi}  
has been the default approach to trajectory decoding due to its simplicity.
However, the trajectories output by these methods tend to go off-map
unless prior knowledge losses are imposed as soft constraints \cite{casas2020importance}. 
While methods that model the future trajectory autoregressively over time
\cite{rhinehart2018r2p2,ivanovic2019trajectron} can help with better map understanding,
they bring the additional issue of compounding errors \cite{ross2011reduction}.
Many state-of-the-art methods today factorize trajectory prediction into goal (endpoint) prediction and
trajectory completion \cite{zhao2020tnt,gu2021densetnt,gilles2021home,gilles2021thomas,deo2020trajectory,zeng2021lanercnn, deo2022multimodal}. The intuition is that the goal captures most of the stochasticity in the future and also allows
for simple multimodal reasoning.
Since runtime is not evaluated in motion forecasting benchmarks, ensembles of multiple models followed by clustering
\cite{varadarajan2022multipath,wang2022tenet,nayakanti2022wayformer} have emerged in order to predict a better covering set of modes.
However, this is prohibitive for autonomy where reaction time is critical.
In our method, we use goal-based prediction anchored to lane graph nodes, and propose a simple yet
effective greedy goal sampler that is competitive without incurring any expensive postprocessing
such as ensemble clustering.

\begin{figure*}[t]
    \centering
    \includegraphics[width=\textwidth]{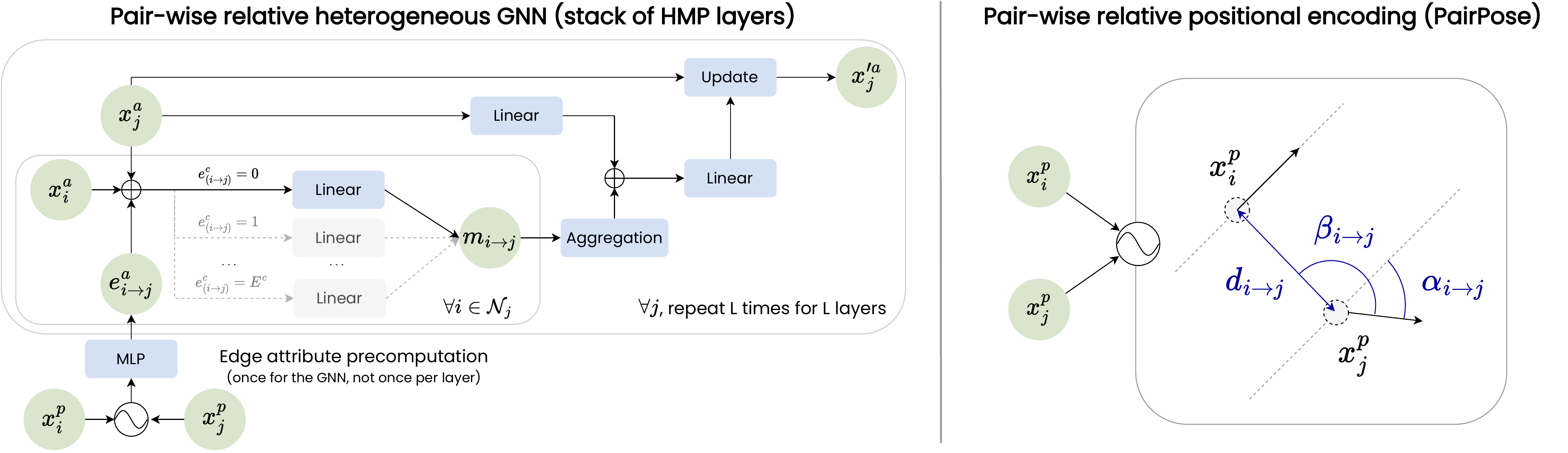}
   \vspace{-20pt}
    \caption{Left: \textbf{Heterogeneous message passing} between two nodes. Right: Zoom-in to visually explain our \textbf{pair-wise relative positional encoding}.}
    \vspace{-15pt}
    \label{fig:hetero_message_passing}
 \end{figure*}

\section{Pair-wise Relative Heterogeneous GNN}
\label{sec:hmp}

Graph neural networks (GNN) process graph structures by exchanging messages between node pairs connected by edges.
In this section, we propose a novel GNN architecture that will be extensively used in the next section as the backbone for encoding graphs composed of both agents and map nodes, and decoding future predictions from them. 
The goal of our architecture is to capture complex interactions between nodes in a heterogeneous spatial graph.
By spatial, we mean that every node $\x_i$ has both a feature vector with general attributes $\x_i^a$ as well as a 3-DOF pose $\x_i^p$ describing its centroid and yaw. 
Heterogeneous here means that nodes may belong to different semantic categories, and their node attributes may have different dimensionality.
This graph neural network models directional relationships via continuous edge attributes $\e^a_{i \to j}$ and discrete edge classes $\e^c_{i \to j}$. 
In the next section, we describe the parameterization of the positional encodings we employ to describe the relative pose between two nodes, which make learning easier by being magnitude bounded and viewpoint-invariant (i.e., agnostic to reference frame rotations and translations). 
We then explain our heterogeneous message passing layer, which exploits these encodings as edge attributes.  We refer the reader to Fig. \ref{fig:hetero_message_passing} for an illustration of this layer and the pair-wise relative positional encodings.

\vspace{0.2cm}
\noindent{\bf Pair-wise relative positional encoding (PairPose):}
Each node in the graph has a pose $\x_i^p$, which is composed of
a centroid $\bm{c_i}$ and a unit vector in the heading direction $\bm{h_i}$.  To represent the directional, pairwise relationship between node $i$
and $j$ (i.e., $i \to j$), we first compute the displacement vector between each node's centroids $\bm{v}_{i
\to j} = \bm{c}_i - \bm{c}_j$ as well as the sine and cosine of the heading difference 
$$\sin(\alpha_{i \to j})=\bm{h}_i \times \bm{h}_j, \,\,\quad \cos(\alpha_{i \to j})=\bm{h}_i \cdot
\bm{h}_j$$ 
Note that the displacement vector $\bm{v}_{i \to j}$ depends on the arbitrary global frame the
centroids are expressed in, and thus is not viewpoint invariant. 
To achieve invariance,
we instead utilize the centroid distance $d_{i \to j} = ||\bm{v}_{i \to j}||_2$, together with the sine and
cosine of the angle between the displacement vector $\bm{v}_{i \to j}$ and the heading  $\bm{h}_j$ (shown in Fig. \ref{fig:hetero_message_passing}-Right)
$$\sin(\beta_{i \to j})=\frac{\bm{v}_{i \to j} \times \bm{h}_j}{|\bm{v}_{i \to j}| |\bm{h}_j|},
\cos(\beta_{i \to j})=\frac{\bm{v}_{i \to j} \cdot \bm{h}_j}{|\bm{v}_{i \to j}| |\bm{h}_j|}$$

To make the centroid distances bounded, we follow \cite{vaswani2017attention} and  map each distance to a vector $\bm{p}_{i \to j} = [p_1, \dots, p_N, r_1, \dots, r_N]$  composed of sine and cosine functions of $N$ different frequencies that represent the range of distances that we are interested in (from a handful of meters to hundreds of meters). More concretely, 
$$p_n = \sin(d_{i \to j} \exp(\frac{4n}{N})), \, \, \quad r_n = \cos(d_{i \to j}
\exp(\frac{4n}{N}))$$
The pair-wise geometric relationship of entities $i$ and $j$ can be summarized as a
concatenation ($\oplus$): 
$$\g^a_{i \to j}=[\sin(\alpha_{i \to j}), \cos(\alpha_{i \to j}), \sin(\beta_{i \to j}),
\cos(\beta_{i \to j})] \oplus \bm{p}_{i \to j}$$
The final positional encoding is then learned as 

\vspace{-5pt}
\begin{equation}
    \label{eq:pos_enc}
    \e^a_{i \to j} = \text{MLP}(g^a_{i \to j})
    \vspace{-5pt}
\end{equation}

\vspace{0.2cm}
\noindent{\bf Heterogeneous message passing (HMP) layer:} The goal of this layer is to update the node features $\x^a_j$ in a directed graph by taking into account all its incoming neighbors' features $\x^a_i \in \mathcal{N}_j$ as well as their pair-wise relative positional encodings $\e_{i \to j}$. The message passing process is depicted in Fig. \ref{fig:hetero_message_passing}-Left.
For every neighbor $i \in \mathcal{N}_j$, we compute a message $m_{i \to j}$ by linearly projecting the concatenation of the source features $\x^a_i$ and edge attributes $\e^a_{i \to j}$. 
Note that the weights of this linear layer are different depending on the discrete edge class $\e^c_{i \to j}$, and that the dimensionality of those weights depends on the dimensionality of the node features, which vary by node class. 
Then, all incoming messages are aggregated using a permutation invariant aggregation function (e.g., max-pooling). Subsequently, aggregated messages are concatenated with a linear projection of the node features, and fused with another linear layer. Finally, an update function (e.g., GRU cell) enables our heterogeneous message passing layer to keep or forget information from the previous features $\x^a_j$ into the updated features $\x'^a_j$.
We compose our pair-wise relative heterogeneous GNN by stacking multiple HMP layers.
We do not use GAT \cite{velickovic2017graph} as it only uses the edge to compute attention weights over neighboring nodes, and does not use the edge embedding to directly update the node embedding, which limits its model capacity.

\section{Viewpoint-invariant Motion Forecasting} 
\label{sec:pipeline}

\begin{figure*}[t]
   \centering
   \includegraphics[width=\textwidth]{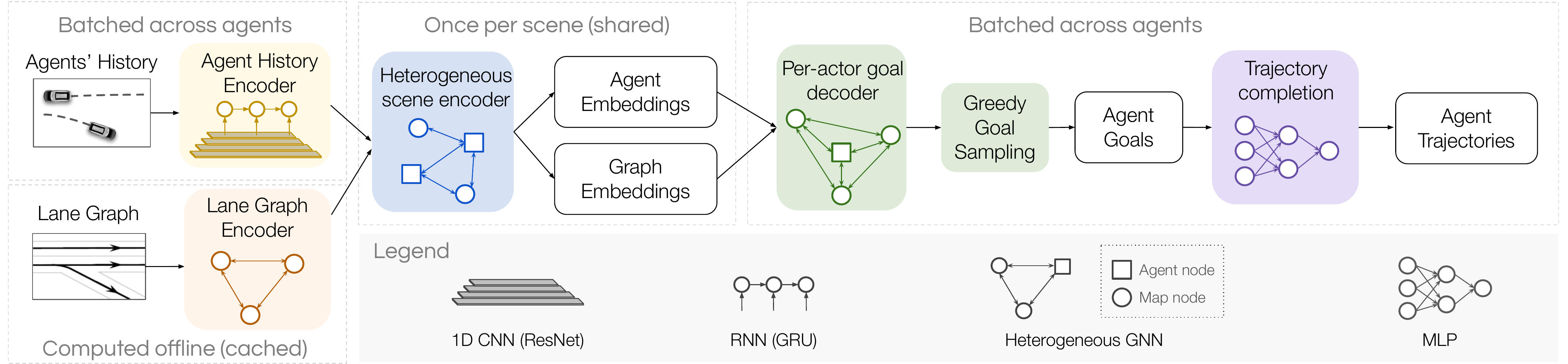}
   \vspace{-20pt}
   \caption{\textbf{\ourmodel{} model pipeline}. Thanks to its viewpoint invariance, our model can
   run the lane-graph encoder offline and cache its activations for online runtime savings, and
   perform heterogeneous scene encoder just once, shared across all agents. }
   \vspace{-15pt}
   \label{fig:overview}
\end{figure*}

In this section, we describe how we efficiently and effectively forecast multi-agent future trajectories based on the map as well as past
trajectories.
Towards this goal, we  first compute agent features and map features separately because (i) the nature of the features is different (spatio-temporal vs. spatial), and (ii) efficiency - the map features can be computed offline as they only rely on the high-definition maps.
We then leverage the heterogeneous message passing (HMP) layers proposed in Section \ref{sec:hmp} to model inter- and intra-class interactions on a single heterogeneous
graph that connects agents and map elements. 
Finally, we predict the future trajectory of each agent by first predicting their goal
(waypoint at the end of the prediction horizon), and subsequently
predicting their full trajectory given the goal.
Fig.
\ref{fig:overview} shows an illustration of our approach. %
We dub our method \ourmodel{} to highlight the fact that we ``go relative'' in the modelling of all pair-wise geometric interactions, achieving a viewpoint agnostic architecture.
In the remainder of this section, we describe every module and the training objective.  

\vspace{0.2cm}
\noindent{\bf Agent history encoder:}
The spatial input features  of current multi-agent prediction systems are typically encoded in a
coordinate frame whose  origin is the SDV position and the x-axis is aligned with its heading \cite{lgn,gilles2021thomas,ngiam2021scene}. This
has the undesired property that the same  trajectory  has different embeddings if the SDV  is at different positions and or orientations in the scene. 
Instead, we design a
viewpoint-invariant representation  to express the past trajectory as a sequence of pair-wise relative positional encodings
between the past waypoints and the current pose. More precisely, we encode the past trajectory as $\tau = [\bm{e}_{{-T} \to 0}, \dots, \bm{e}_{{-1} \to 0}] \in \mathbb{R}^{T \times D}$, where $\bm{e}_{t \to 0}$ refers to the pair-wise relative positional encoding of the pose $x^p_t$ at a past time step $t < 0$ with respect to the current pose $x^p_0$ (as described in Eq. \ref{eq:pos_enc}).
Note that current time is $t=0$, negative times denote the past with $t=-T$ being the history horizon, and $D$ is the dimensionality of our pair-wise relative positional encoding.
We then pass each agent's features into a 1D convolutional network with residual connections followed by a gated recurrent unit (GRU), and use the final hidden state of the GRU as the embedding of the agent's trajectory. Intuitively, this works well because the 1D CNN can extract local temporal patterns and the GRU can aggregate them in a learnable hidden state.
Importantly, the computation for all agents can be batched  and thus inference is very efficient even  in complex crowded scenes.

\vspace{0.2cm}
\noindent{\bf Lane-graph encoder:}
We build our lane graph by sampling the centerlines in the high-definition map  at regular intervals (e.g., $3m$ for urban, $10m$ for highway) to obtain lane segments. Each lane segment is represented as a node in the graph containing features such as its length, width, curvature, speed limit, and lane boundary type (e.g., solid, dashed).
Following \cite{lgn}, these lane nodes are then connected with 4 different relationships: successors, predecessors, left and right neighbors. For successors and predecessors, we also include dilations (i.e., skip connections) for $\{2,3,4,5\}$ hops in order to make the receptive field of our GNN grow faster over layers, which helps particularly with fast moving vehicles. 
On top of these, we find it helpful to sample map nodes uniformly from crosswalk polygons since these are highly relevant for pedestrians and vehicles interacting with pedestrians. To understand the interactions occurring at overlapping map entities (e.g., lanes at intersections, crosswalks and lanes) we add a ``conflict" edge for every pair of nodes in the graph that are less than 2.5 meters apart and belong to distinct map entities.
We use a stack of the previously introduced HMP layers to update the embeddings of every node in the graph.
Importantly, our viewpoint agnostic lane graph encoder allows us to compute map embeddings offline for large-scale maps after training \ourmodel{}.
Thanks to its memory-efficiency and coordinate frame invariance, we can compute the map embeddings for a large tile offline in a single inference pass over the lane-graph encoder, eliminating the need for stitching regions of map embeddings together.
On the onboard side, a map provider serves large map tiles to
autonomy every several seconds which contain the cached lane graph embeddings, improving the SDV reaction time. 

\vspace{0.2cm}
\noindent{\bf Heterogeneous scene encoder:}
We fuse the agent and lane graph features using another stack of HMP layers.
The heterogeneous scene graph contains agents and map nodes, whose input attributes are initialized with the results of the previously described agent history encoder and lane graph encoder respectively. 
In addition to the map-to-map edges used in the lane graph encoder, we include agent-to-agent, agent-to-map and map-to-agent edges in the graph. 
We create agent-to-map and
map-to-agent edges between agents and the lane graph node that is nearest to their position at
$t=0$. For agent-to-agent, we connect every pair of agent nodes that closer than 100 meters.
At every round of message passing we compute messages for all edge types, and update both the agent and lane graph node embeddings with heterogeneous messages incoming from the different edge types.
This is different from LaneGCN\cite{lgn}, which defines a custom ordering for message passing of different types. Unlike HEAT's \cite{mo2022multi} message passing, our HMP layers use edge functions
with specialized parameters for each edge class for improved expressivity.

\begin{table*}[t]
      \centering
      \begin{threeparttable}
          \begin{tabularx}{\textwidth}{ l |  %
                          l | r s s s s s s s s%
                          }
          \toprule
                  Split & Model & BrierFDE K=6  & FDE K=6 & ADE K=6 & MR K=6& FDE K=1 & ADE K=1
                  & MR K=1   \\
              \midrule
    \multirow{2}{2cm}{Test (focal agent)} & THOMAS \cite{gilles2021thomas}  & 2.16  & 1.51 & 0.88 & $\mathbf{0.20}$ & 4.71 & 1.95 & $\mathbf{0.64}$ \\
    & \ourmodel & $\mathbf{2.01}$ & $\mathbf{1.48}$ & $\mathbf{0.76}$ & 0.22 & $\mathbf{4.62}$ & $\mathbf{1.82}$ & 0.66 \\
    \midrule
    \multirow{4}{2cm}{Validation (multi-agent: focal+scored)}
    & MultiPath*\cite{chai2019multipath} & 2.54 & 2.13 & 0.89 & 0.33 & 14.90 & 7.06 & 0.52 \\
    & MTP*\cite{cui2018multimodal}  & 2.05 & 1.54 & 0.68 & 0.24 & 6.66  & 2.74 & 0.47 \\
    & LaneGCN\cite{lgn}   & 1.94 & 1.34 & 0.55 & 0.22 & 4.82  & 1.82 & 0.51 \\
    & SceneTransformer*\cite{ngiam2021scene}  & 1.80 & 1.24 & 0.52 & 0.20 & 4.57  & 1.75 & 0.46 \\
    & \ourmodel   & $\mathbf{1.29}$ & $\mathbf{0.96}$ & $\mathbf{0.42}$ & $\mathbf{0.14}$ & $\mathbf{2.43}$  & $\mathbf{0.95}$ & $\mathbf{0.32}$ \\
        \bottomrule
          \end{tabularx}
      \end{threeparttable}
      \caption{\scriptsize \textbf{Comparison against state of the art on Argoverse 2}. All metrics are minimum error over $K$ modes (lower is better). \\
       * indicates re-implementation as these methods are not open-sourced.\\
      }
      \vspace{-25pt}
      \label{table:main_argoverse}
  \end{table*}

  \begin{table}[t]
      \centering
      {
        \setlength{\tabcolsep}{3.2pt}
        \begin{threeparttable}
            \begin{tabularx}{\columnwidth}{%
                            l | r r r r r r r r r%
                            }
            \toprule
                    Model & BrierFDE & BrierATE  & BrierCTE & ATE  & CTE \\
                    {} & K=6 & K=6 & K=6 & K=1 & K=1 \\
                \midrule
        MultiPath*\cite{chai2019multipath} & 5.37 & 2.42 & 0.68 & 17.07 & 0.86 \\
        MTP*\cite{cui2018multimodal}  & 3.43 & 1.64 & 0.88 & 5.08  & 0.59 \\
        LaneGCN\cite{lgn}  & 3.78 & 1.31 & 0.89 & 4.67  & 0.81 \\
        SceneTransformer*\cite{ngiam2021scene}  & 3.17 & 1.51 & 1.15 & 3.56  & 0.71 \\
        \ourmodel & $\mathbf{2.63}$ & $\mathbf{1.14}$ & $\mathbf{0.57}$ & $\mathbf{2.27}$  & $\mathbf{0.40}$ \\
            \bottomrule
            \end{tabularx}
        \end{threeparttable}
      }
      \vspace{-8pt}
      \caption{\scriptsize \textbf{Comparison against state of the art on HighwaySim}. All metrics are minimum error over $K$ modes.}
      \vspace{-25pt}
      \label{table:main_highwaysim}
  \end{table}

\vspace{0.2cm}
\noindent{\bf Goal-based decoder:}
To decode agent future trajectories, we first predict their goal (last waypoint).
We cast the goal prediction as a multi-class classification problem over lane graph nodes \cite{zeng2021lanercnn}, as this allows us to leverage the rich map embeddings computed by the heterogeneous scene encoder. To achieve a resolution beyond that of the lane-graph, we also regress a continuous offset for each candidate goal with respect to its lane-graph node anchor.
To predict these classification scores and regression offsets, we create a graph composed of $A$ connected components, one for each agent in the scene. 
Each connected component contains one actor and a copy of all the map nodes. 
Regarding the graph connectivity, we preserve the same map-to-map edges as in upstream components, and connect all possible map-to-actor and actor-to-map pairs.
The agent and map node features are initialized to the outputs of the heterogeneous scene encoder, and they are updated by a stack of HMP layers.

\vspace{0.2cm}
\noindent{\bf Greedy goal sampler:}
Predicting multi-modal trajectories is essential for the motion planner to be safe with respect to any future that might unfold. 
To predict $K$ modes, we propose a simple yet effective greedy
goal sampler. At every iteration, we (1) sample the goal with the highest probability, (2) remove every
node closer than $\gamma$ meters, (3) downweight every node closer than $\nu$ ($\nu > \gamma$)
meters by a factor of $\tau$. We repeat this process for $K$ iterations. The intuition is that if the probability distribution is uni-modal we should draw as many samples as possible from it, while if the distribution is multi-modal (e.g., agent at intersection), it is desirable to sample from different modes even if the probability mass around each mode is imbalanced.

\vspace{0.2cm}
\noindent{\bf Trajectory completion:}
Finally, we perform trajectory completion to all goals in parallel. 
We perform the trajectory prediction in agent-relative coordinates such that the network can leverage the prior that vehicles initially progress along their heading direction (i.e., the x-axis). 
We use a shared MLP across agent classes (vehicles, pedestrians and cyclists) that predicts the sequence of 2D waypoints as a flat vector in $\mathbb{R}^{2T_f}$, where $T_f$ is the prediction horizon. We find that sharing the MLP layers across agent classes is beneficial as there are some priors that all trajectories follow, which is beneficial for learning in class-imbalanced settings.

\vspace{0.2cm}
\noindent{\bf Training:}
We optimize a multi-task loss, which is  a linear combination of 3 terms: goal classification,
goal regression and trajectory completion.
We employ focal loss \cite{lin2017focal} for goal classification,  serving as a form of hard
example mining.
We supervise the offset regression only for the node that is closest to the goal, and use a Huber
loss in node frame.
We train our trajectory completion using teacher forcing, meaning that during training we feed the
ground-truth goal to the trajectory completion module. 
This is beneficial since our goal predictions may not always capture the right mode for the goal, particularly at early stages of training.
We employ a simple Huber loss on each waypoint coordinate in agent-centric frame.
Our final model was trained for 17 epochs, using 16 T4 GPUs for a total batch size of 64. We use Adam optimizer with a learning rate 5.e-4, with a step-wise scheduler with 0.25 decay and 15 step-size.

\section{Experimental Evaluation}
\label{sec:exps}

\begin{table}[t]
        \centering
    \begin{threeparttable}
    \setlength\tabcolsep{5pt} %
    \begin{tabularx}{\columnwidth}{ l | c c c c | s s s%
                    }
                    \toprule
            
            Model$^\dagger$ & $\mathcal{E}$ & $\mathcal{D}$ & $\mathcal{G}$ & $\bm{e}_{i \to j}$ &
            BrierFDE K=6  & FDE K=6 & FDE K=1 \\
    
            \midrule
    
            $M_1 $  & \xmark & \cmark & \cmark & \cmark & 1.53 & 1.18 & 3.52 \\
            $M_2 $ & \cmark & \xmark & \xmark & \cmark & 2.19 & 1.59 & 5.67 \\
            $M_3 $  & \cmark & \cmark & \xmark & \cmark & 1.65 & 1.21 & $\mathbf{2.77}$ \\
            $M_4 $  & \cmark & \cmark & \cmark & \xmark & 1.89 & 1.43 & 3.44 \\
            $\ourmodel$  & \cmark & \cmark & \cmark & \cmark & $\mathbf{1.45}$ & $\mathbf{1.08}$ & $\mathbf{2.77}$ \\

            \bottomrule
    \end{tabularx}
    \end{threeparttable}
    \vspace{-8pt}
    \caption{\scriptsize \textbf{Ablation study} on Argoverse 2 (val). We ablate the scene encoder $\mathcal{E}$, goal-based decoder $\mathcal{D}$, greedy sampler $\mathcal{G}$, and PairPose in HMP $\bm{e}_{i \to j}$.
    $^\dagger$ Models trained on 25$\%$ of the training set.}
    \vspace{-25pt}
    \label{table:ablation_argoverse}
    \end{table}

In this section, we first show that our method outperforms the
state-of-the-art  in both urban and highway benchmarks. 
We then show through a comprehensive ablation study that each component of the model contribute to our improvements. 
Finally, we show qualitative results.
Please see our supplementary materials \ref{sec:supp} for implementation details, a quantitative study of the advantages of viewpoint invariance when perturbing the SDV heading, an analysis of how runtime scales with number of actors, %
and experiments on the sample efficiency when using pair-wise relative positional encoding vs. geometric features in a global coordinate frame.

\begin{figure*}[t]
    \centering
    {\arrayrulecolor{gray}
    \setlength{\tabcolsep}{0pt}
    \renewcommand{\arraystretch}{0.5}
    \begin{tabular} {c | c | c | c}
        \includegraphics[width=0.245\textwidth, trim={8cm, 2cm, 3cm, 1cm}, clip]{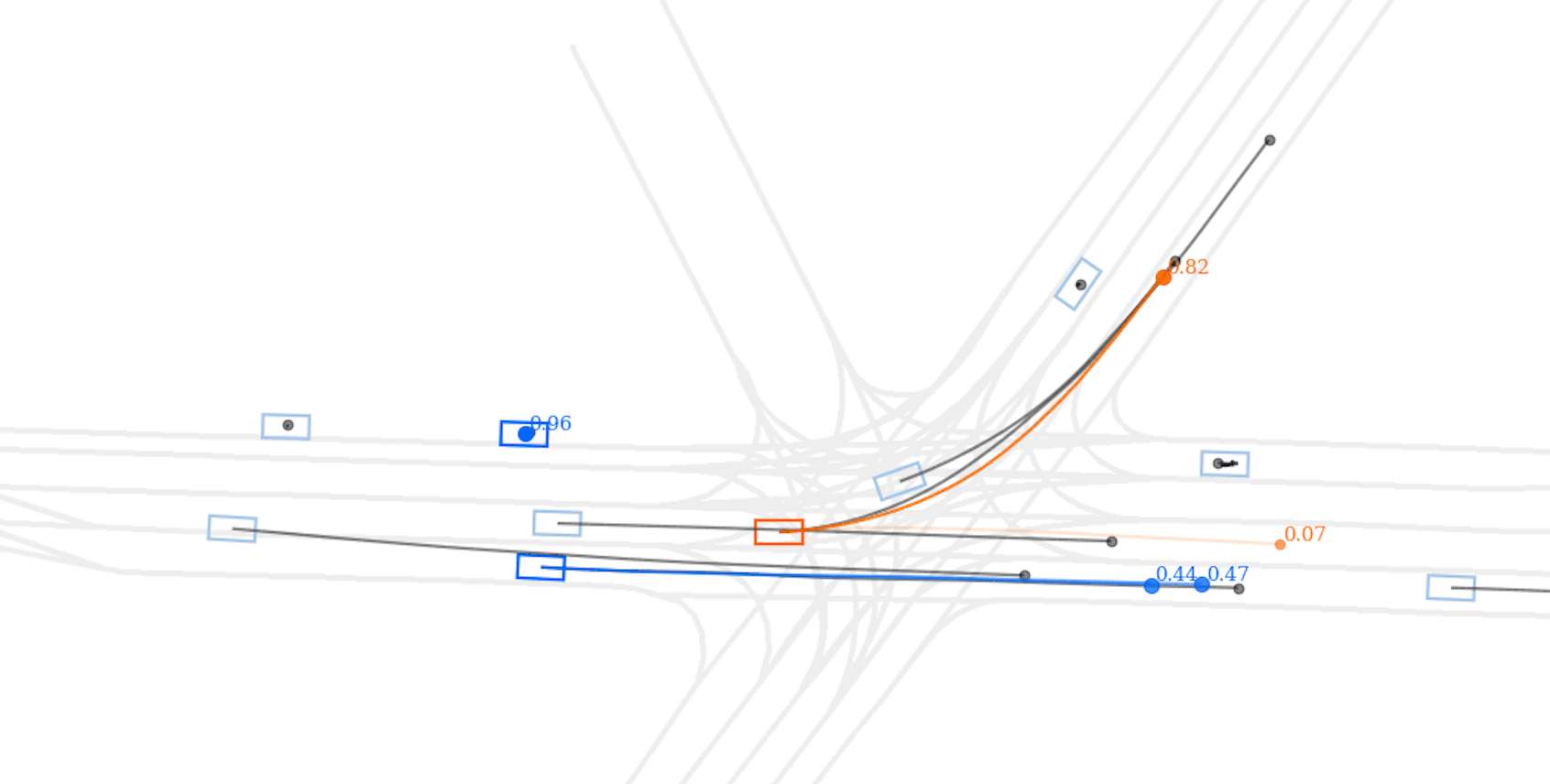} &
        \includegraphics[width=0.245\textwidth, trim={8cm, 2cm, 3cm, 1cm}, clip]{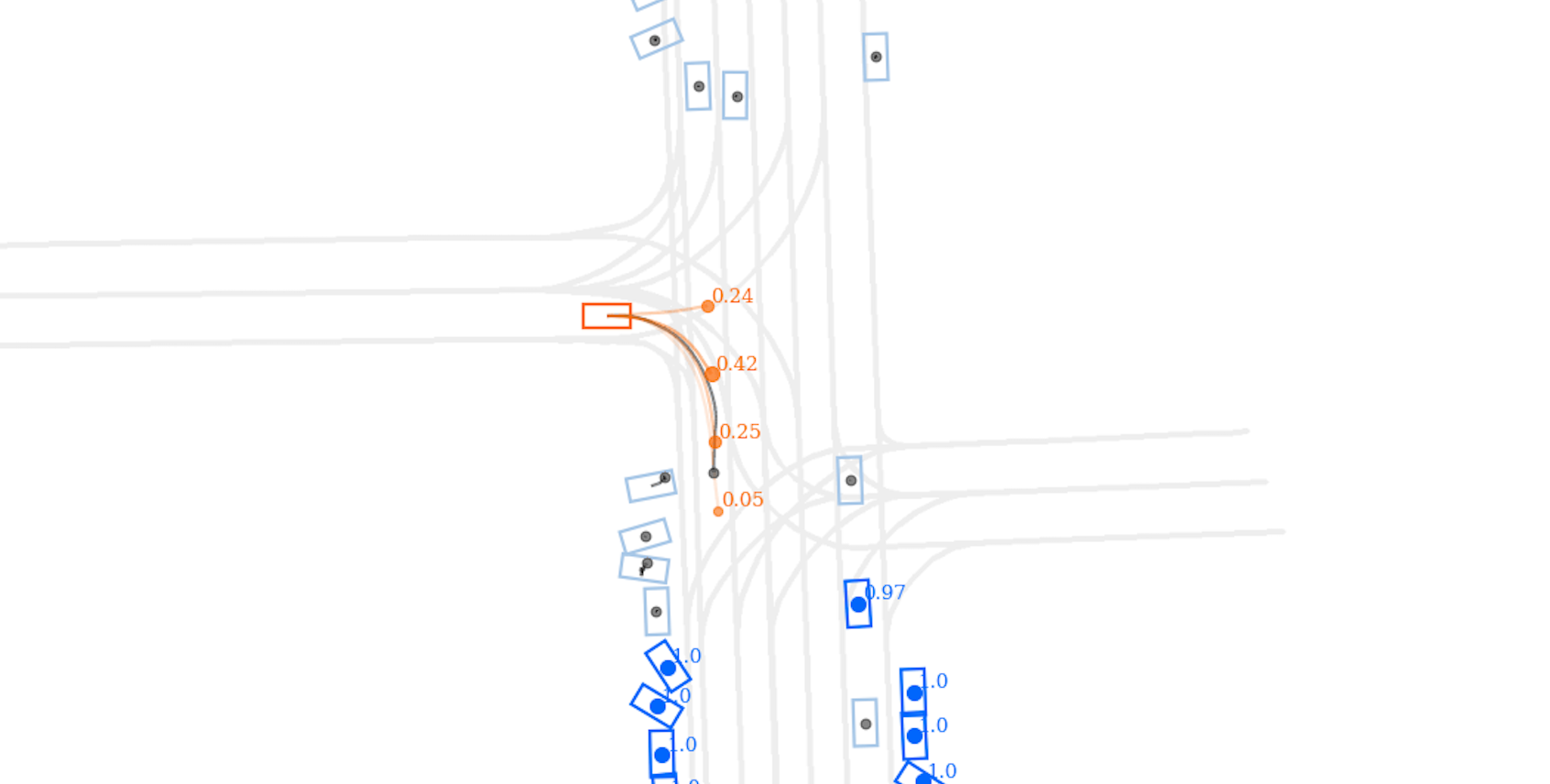} &
        \includegraphics[width=0.245\textwidth, trim={5cm, 2cm, 6cm, 1cm}, clip]{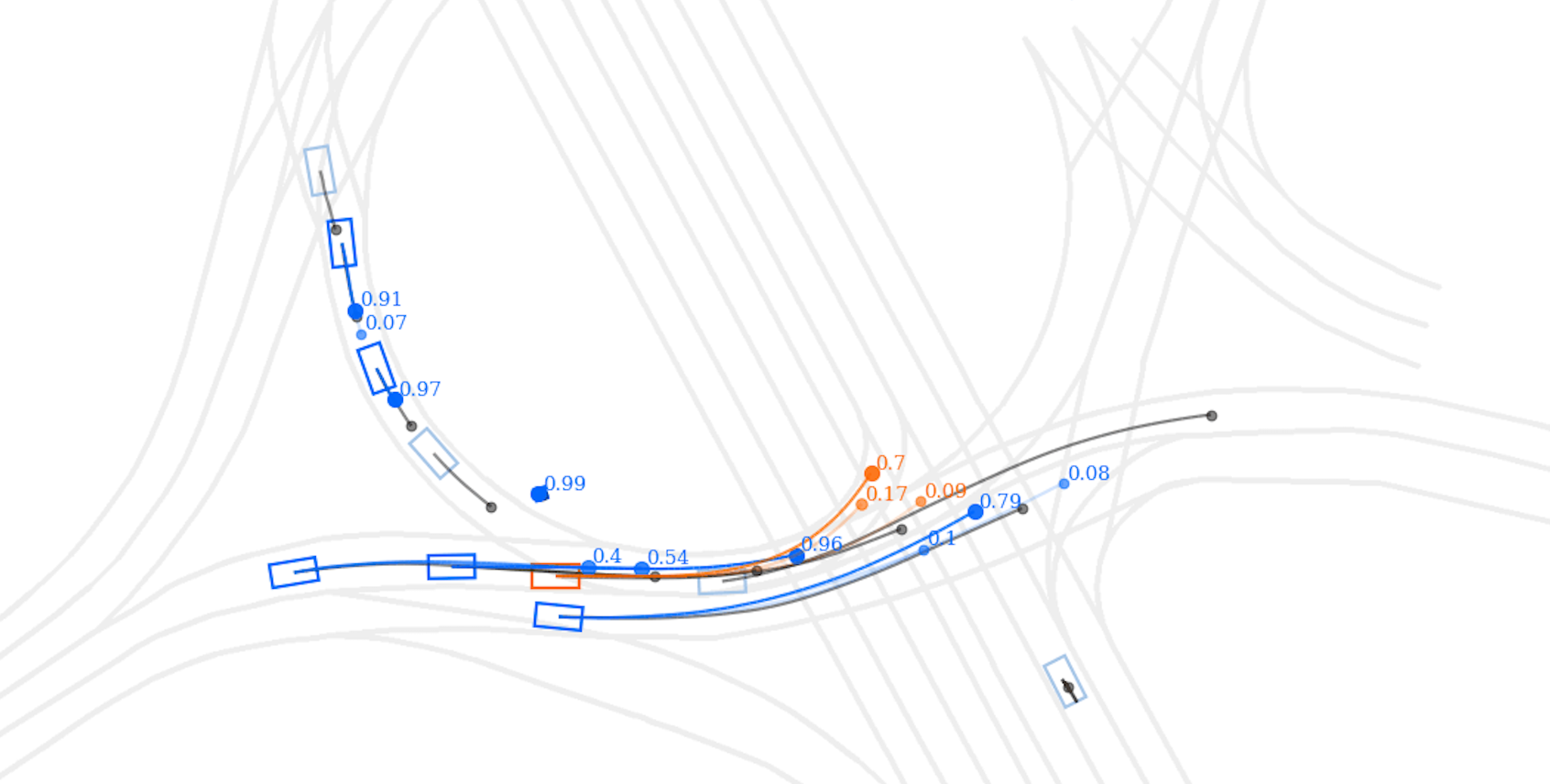} &
        \includegraphics[width=0.245\textwidth, trim={3cm, 2cm, 8cm, 1cm}, clip]{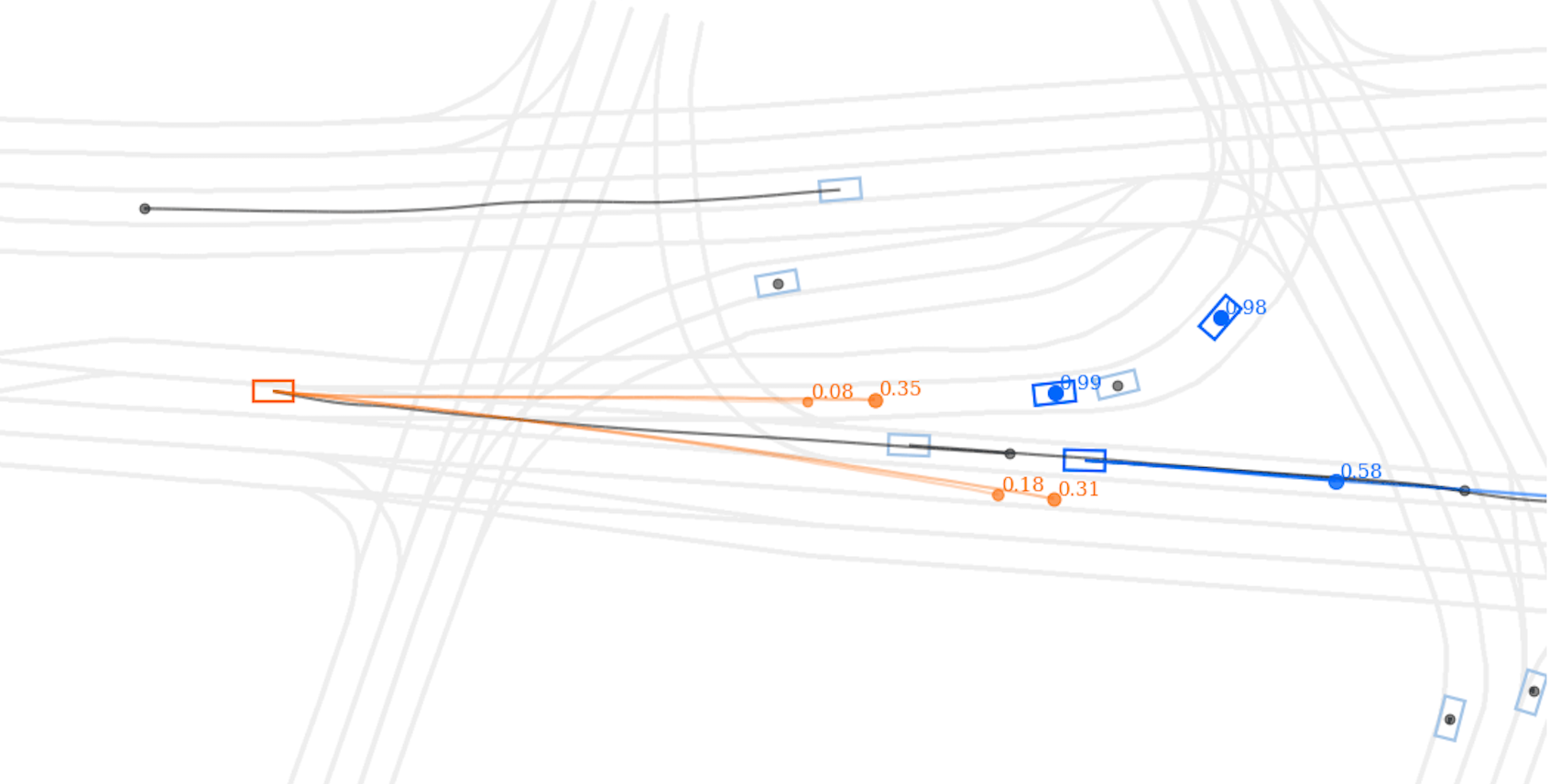} \\
        \midrule
        \includegraphics[width=0.245\textwidth, trim={8cm, 2cm, 3cm, 1cm}, clip]{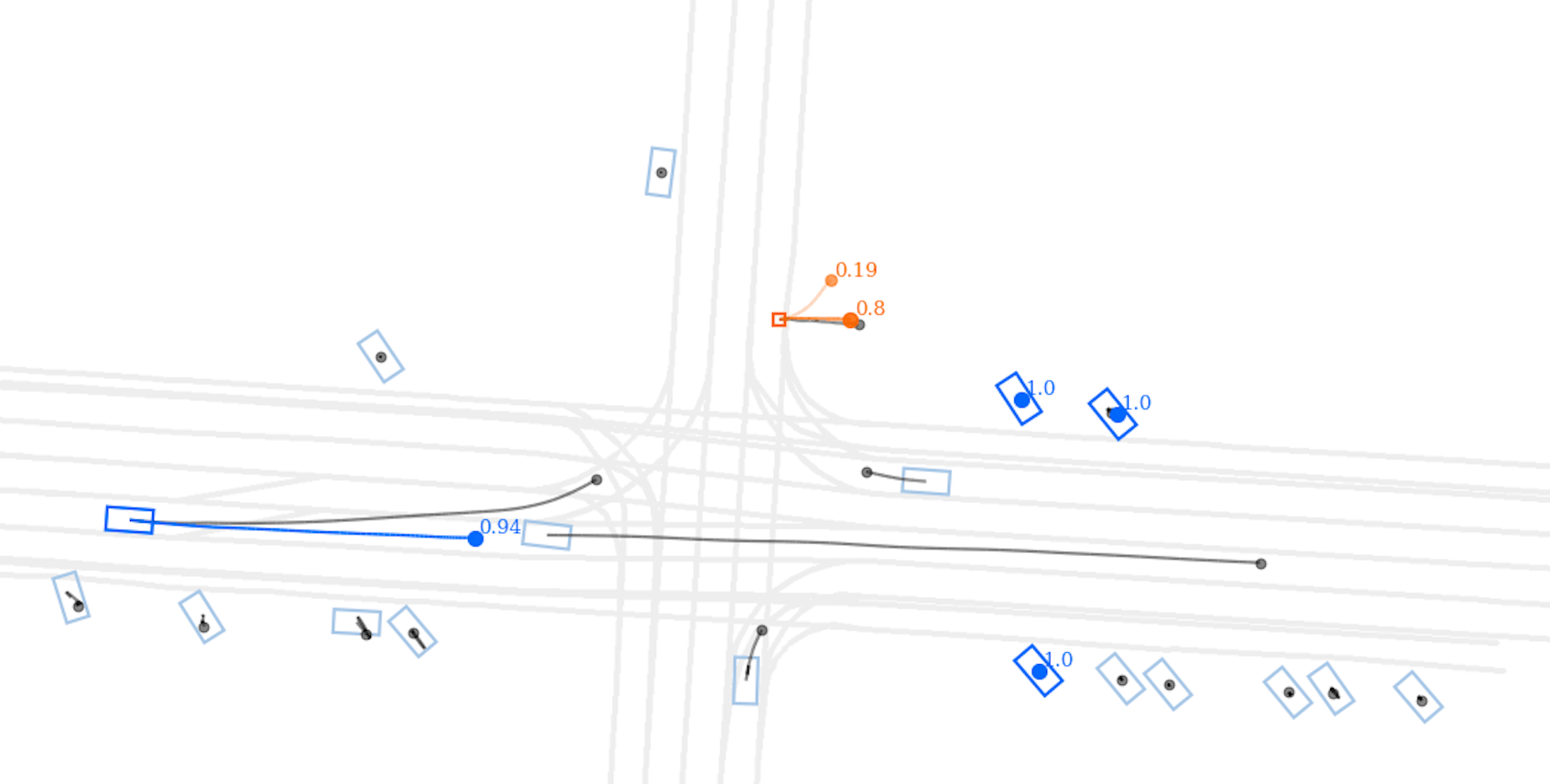} &
        \includegraphics[width=0.245\textwidth, trim={6cm, 2cm, 5cm, 1cm}, clip]{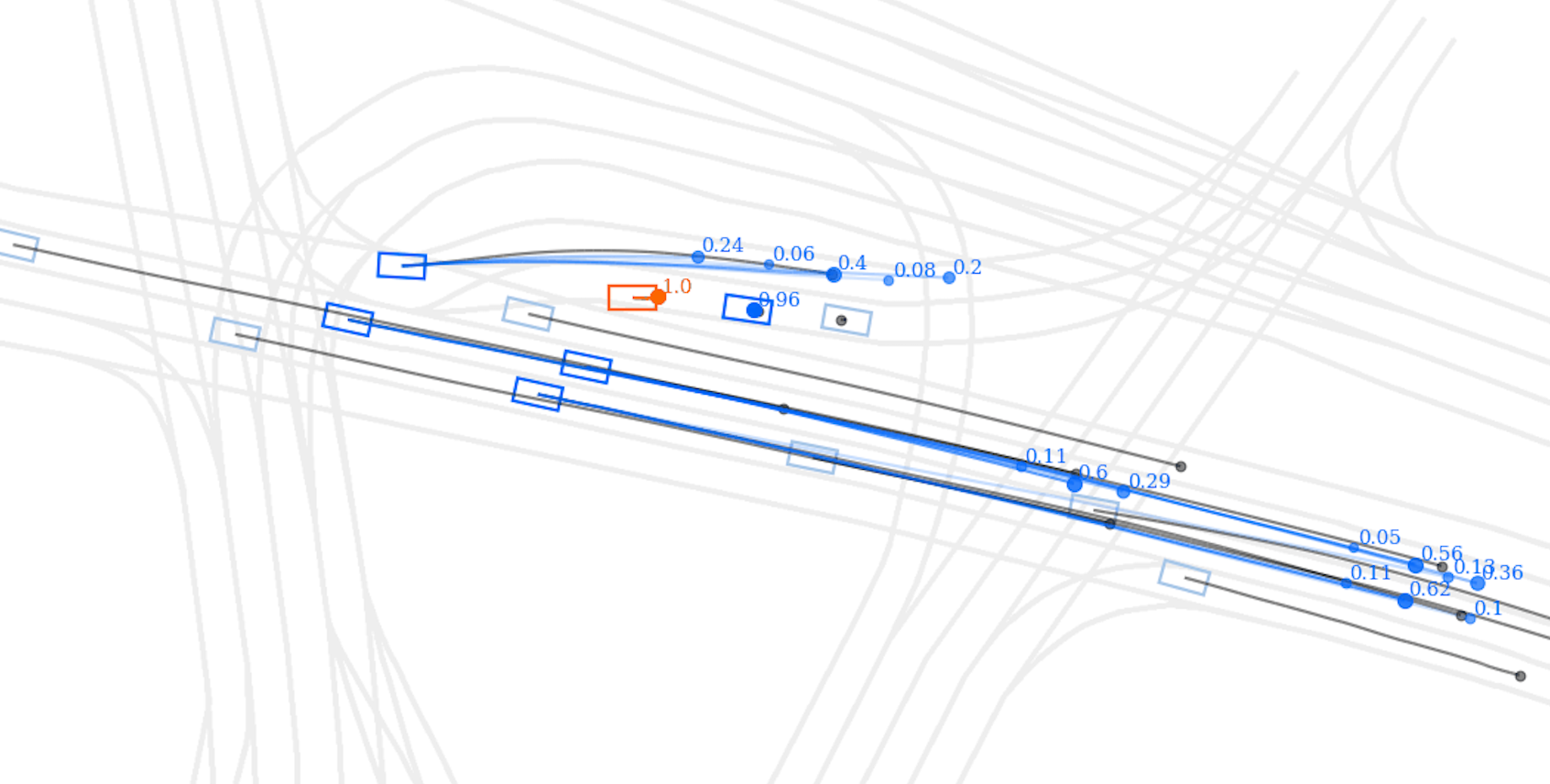} &
        \includegraphics[width=0.245\textwidth, trim={3.5cm, 2cm, 7.5cm, 1cm}, clip]{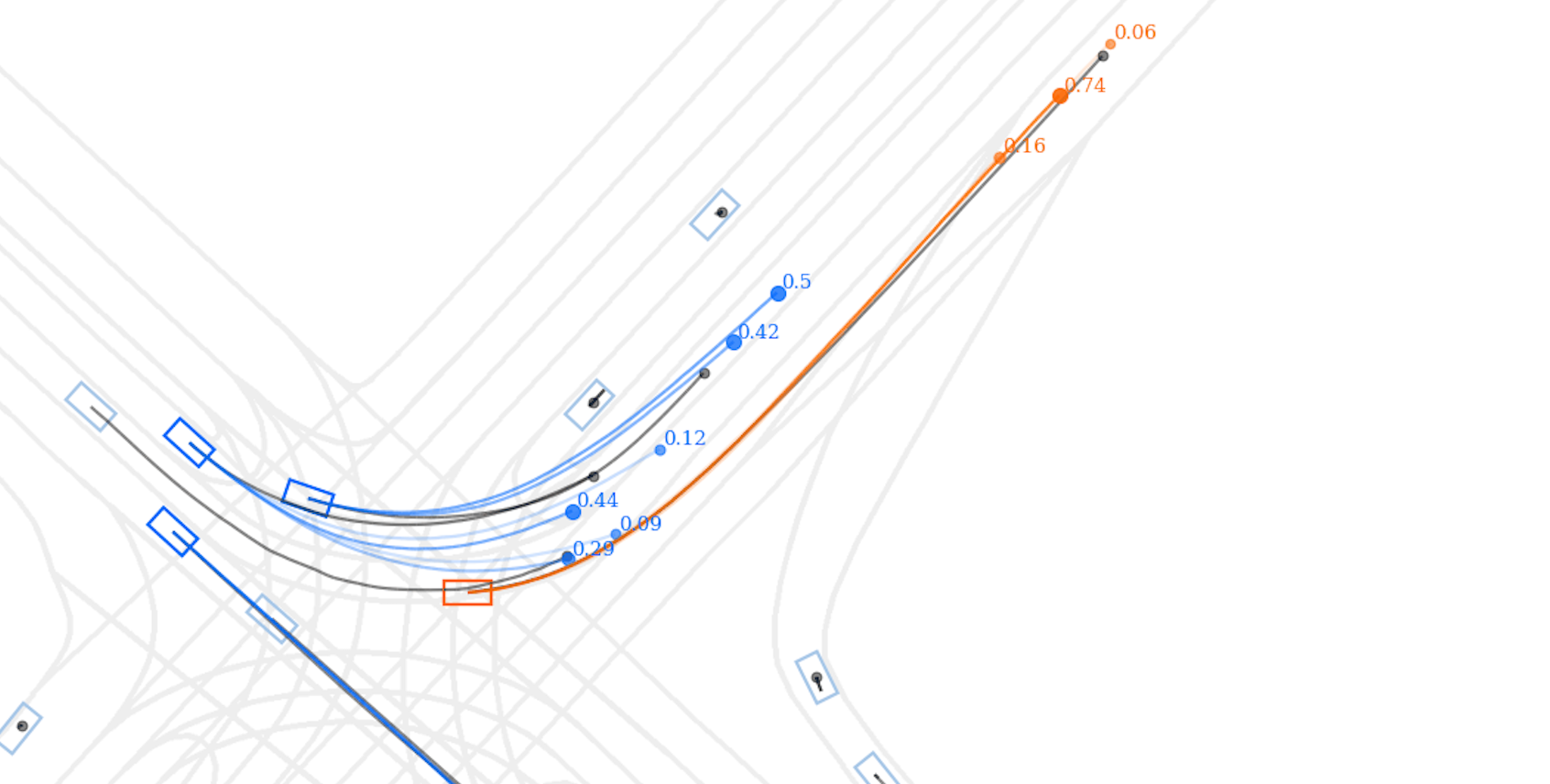} &
        \includegraphics[width=0.245\textwidth, trim={3cm, 2cm, 8cm, 1cm}, clip]{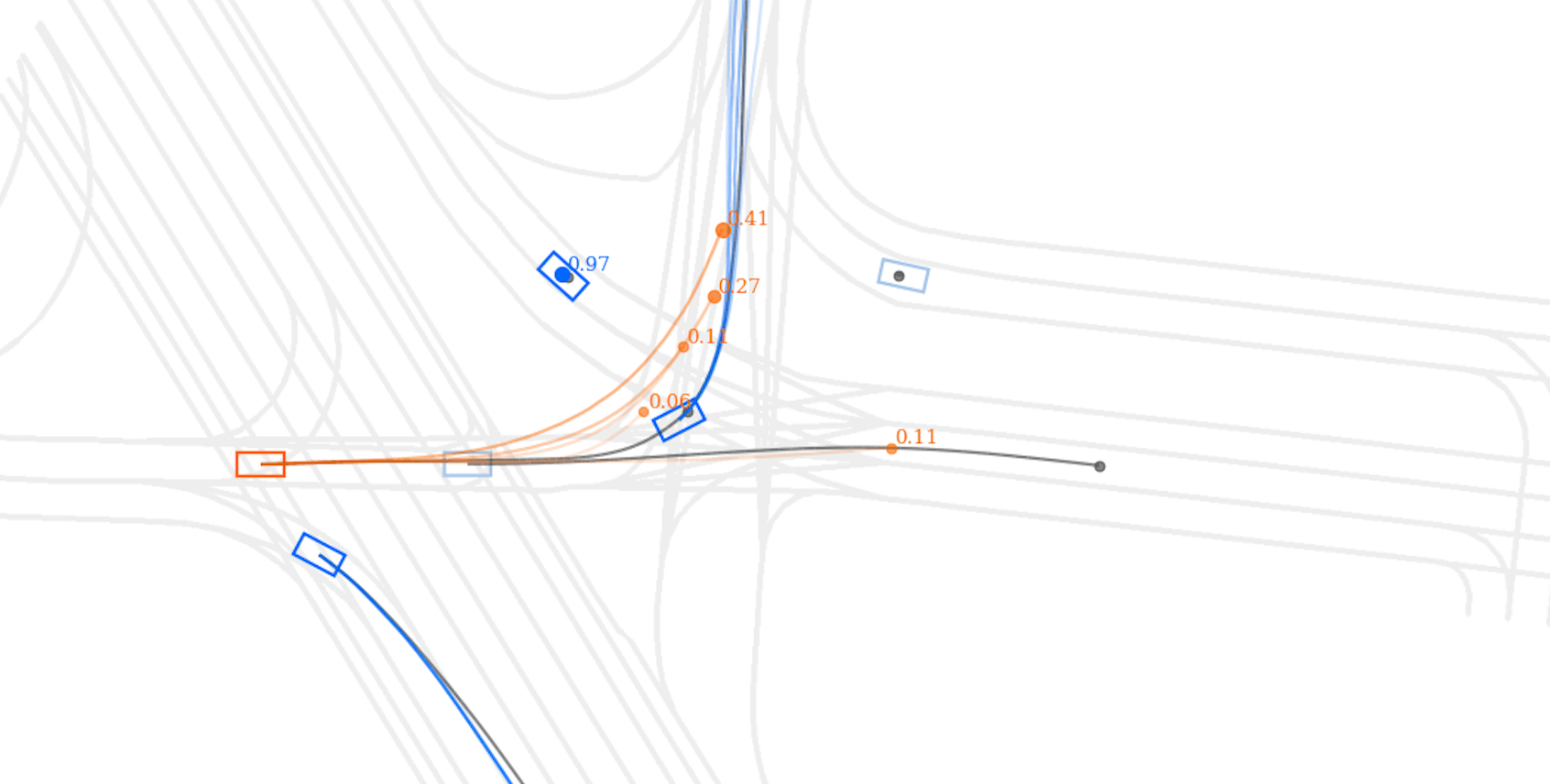} \\
    \end{tabular}
    }
    \caption{\textbf{Qualitative results in Argoverse 2}. 6 seconds long multi-modal motion forecasts for {\color{Orange} focal} and {\color{NavyBlue} scored} agents. Ground-truth plotted in gray.}
    \vspace{-10pt}
    \label{fig:qualitative}
\end{figure*}

\vspace{0.2cm}
\noindent{\bf Datasets:} We use two complementary datasets to evaluate how well our method works in both urban and highway domains, which present different challenges.
 Argoverse 2 \cite{wilson2021argoverse} includes 250,000 city traffic scenarios where trajectories must be predicted for the future 6s, given 5s of history at 10Hz and high-definition maps. The main difficulty is characterizing the multi-modality of the future trajectory distribution due to (i) complex road topologies and (ii) multi-agent interactions, since the scenarios were mined for such cases. For evaluation, agents are divided into focal (one per scenario, hard examples), scored (interactive agents with focal), and unscored (background agents that are not prediction targets).
Our second benchmark is HighwaySim, a novel simulated dataset composed of 70,000 frames of object tracks, including vehicles and motorcycles. We generated this dataset with a state-of-the-art simulator using reactive actors and a variety of highway map topologies such as curved roads, forks and merges. Models are tasked with predicting a 6s trajectory at 2Hz, using 0.5s of history at 10Hz for all agents in the scenario. 
The much higher speeds at which vehicles travel stress the need for a large receptive field, which in turn demands an efficient architecture. Since the SDV is also traveling at high speeds, predicting precise trajectories in terms of lateral deviations from lane centerlines is critical to avoid unsafe harsh brakes.%

\vspace{0.2cm}
\noindent{\bf Comparison against state-of-the-art:}
Table \ref{table:main_argoverse} shows our results in Argoverse 2. 
Our method outperforms all published methods in the leaderboard (test set), which evaluates the prediction quality on difficult-to-forecast agents (focal) in highly interactive scenarios.
Our model performs particularly well on the ranking metric BrierMinFDE, which evaluates not only how close the best mode is to the ground-truth trajectory, but also how much confidence is placed on it. This indicates that our predicted mode probabilities are better calibrated \cite{guo2017calibration} than the baselines'.
We refer the reader to the evaluation website\footnote{\url{https://bit.ly/argo2_eval_metrics}} for further details about the metrics.
We also show large gains in the validation set over various seminal works that are not available on the leaderboard. 
In this experiment, we evaluate multiple agents per scenario, which we care about for the downstream application of self-driving, where the focal and scored
agents are evaluated. We note that the baselines process the scene in the frame of the focal agent as they are not viewpoint invariant.

Table \ref{table:main_highwaysim} showcases the results on HighwaySim.
Due to the greater difference between longitudinal and lateral motion in the highway, it is important to break down the prediction error into Along Track Error (ATE) and Cross Track Error (CTE), which
evaluate the longitudinal and lateral deviations with respect to the ground-truth trajectory.  CTE is much more impactful for driving as it can cause the autonomy system to
confuse which lanes are occupied and thus cause the SDV to suddenly brake putting itself and nearby drivers at risk. 
We show that our method largely outperforms the baselines on all metrics
In particular, we see very large gains on BrierCTE, indicating that our method has a much better map understanding.
Due to the high speeds on the highway, small lateral deviations from the centerlines compound over time, causing large cross-track errors at 6s for the baselines.
For context, a truck only has around 30 cm buffer on each side of the lane taking into account its mirrors, so those errors can be critical.
Similar to results in Table \ref{table:main_argoverse}, the BrierMinFDE metric shows our better calibrated mode probabilities.

\vspace{0.2cm}
\noindent{\bf Importance of the heterogeneous scene encoder:} $M_1$ in Table \ref{table:ablation_argoverse} entirely bypasses this component, essentially feeding agent and map
features to the decoder directly out of the agent history and lane-graph encoders. We see $6\%$ higher BrierMinFDE@K=6.

\vspace{0.2cm}
\noindent{\bf Importance of goal-based decoder:} $M_2$ uses a simple MLP instead of our heterogeneous GNN to predict the trajectory mixture. This is the component that has the biggest impact, increasing the BrierMinFDE@K=6 by $51\%$ when replaced.

\vspace{0.2cm}
\noindent{\bf Importance of greedy goal sampler:} $M_3$ replaces the proposed greedy goal sampler with top-k, an approach used by \cite{zeng2021lanercnn}. We see $14\%$ higher BrierMinFDE@K=6.

\vspace{0.2cm}
\noindent{\bf Importance of pair-wise relative heterogeneous GNN:} $M_4$ replaces this component with a simpler GNN without edge attributes. Geometric information is now encoded on the nodes attributes in a global coordinate frame as proposed in \cite{lgn,gilles2021thomas,ngiam2021scene}. We see $30\%$ higher BrierMinFDE@K=6.

\vspace{0.2cm}
\noindent{\bf Qualitative results:}
We visualize \ourmodel{}'s predictions in a diverse set of scenarios from Argoverse 2 in Fig. \ref{fig:qualitative}.
The first 3 columns show scenarios in which our model accurately predicts a mode very close to the ground-truth for the focal agent, which we emphasize are mined for hard cases. From the visualizations, it is clear our model achieves a good understanding of the map as well as the agent-agent interactions, predicting realistic multi-modal distributions.
The last column showcases failure modes of our model such as predicting an agent will lane change when in fact is keeping its lane, and cutting corners too much in a left turn.

\section{Conclusion}
In this work, we have proposed \ourmodel{}, a motion forecasting model that predicts more accurate multi-agent multi-modal trajectories both in urban and highway roads.
As shown by our experiments, our model achieves this through several contributions:
(i) a viewpoint-invariant architecture that makes learning more efficient thanks to the proposed pair-wise relative positional encoding,
(ii) a versatile graph neural network that understands interactions in heterogeneous spatial graphs with agent and map nodes related by multi-edge adjacency matrices (e.g., lane-lane, agent-lane, agent-agent), and
(iii) a probabilistic goal-based decoder that leverages the lane-graph to propose realistic goals paired up with a simple greedy sampler that encourages diversity.

\pagebreak

\bibliographystyle{IEEEtran}
\bibliography{bibliography}

\clearpage

\section{Appendix}
\label{sec:supp}
\setcounter{section}{0}
\renewcommand\thesection{\Alph{section}}

These supplementary materials present additional experiments in Section \ref{sec:additional_exps} that support the claims in the main paper: improved robustness to rare SDV states, improved sample efficiency, improved runtime as well as providing additional insights via qualitative comparisons on both urban and highway scenarios.
Additionally, implementation details are provided in Section \ref{sec:impl_details} for improved reproducibility.

\section{Additional experiments}
\label{sec:additional_exps}

\begin{figure*}[t]%
    \centering
    \vspace{-10pt}
    \subfloat{{\includegraphics[width=0.35\textwidth]{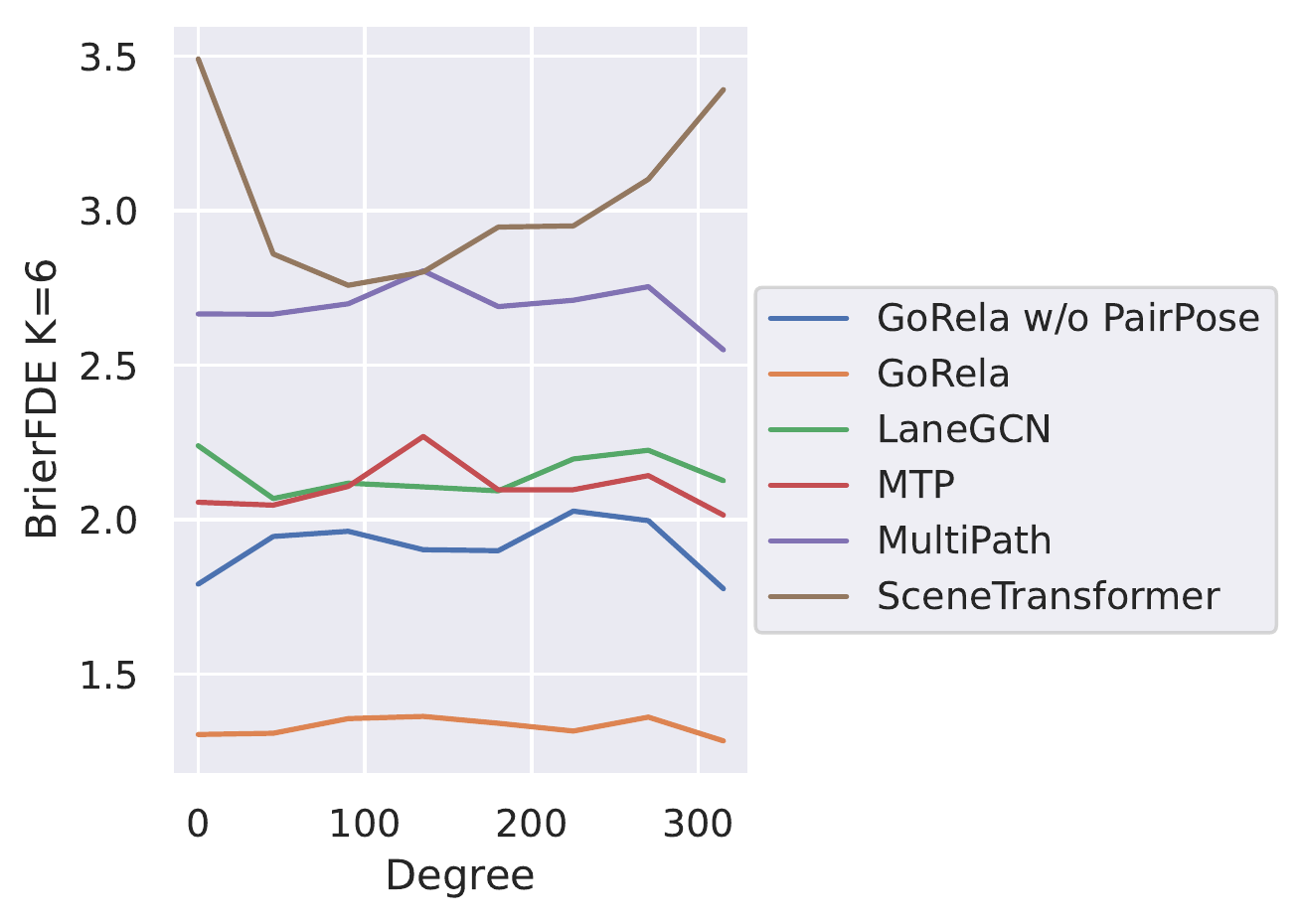} }}%
    \qquad
    \subfloat{{\includegraphics[width=0.55\textwidth]{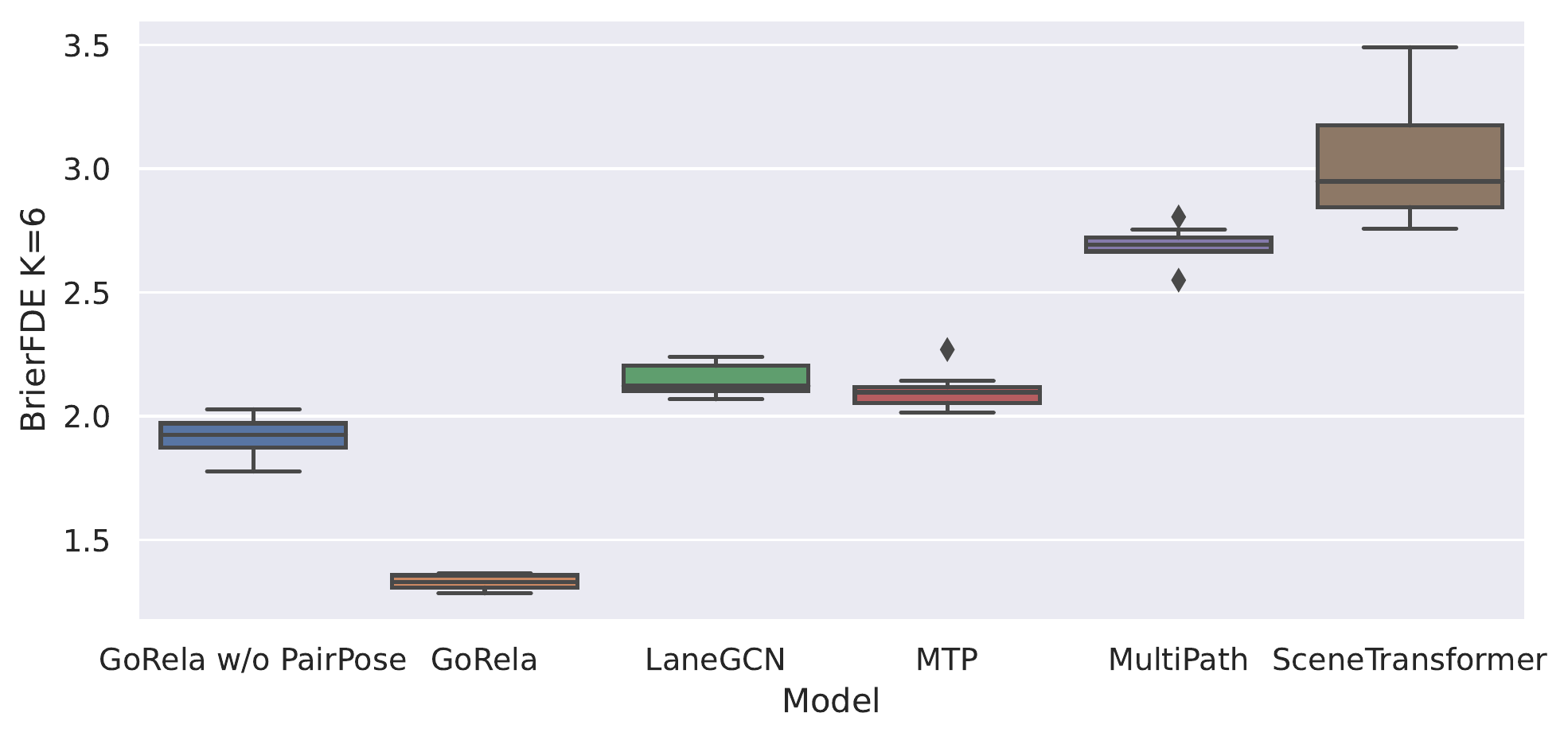} }}%
    \caption{\textbf{Quantitative analysis on viewpoint sensitivity} by randomly rotating scenes around the SDV. 
    Left: error bucketed by rotation bins.
    Right: boxplot of error (variance measures the variability across different rotation bins.)
    }%
    \vspace{-10pt}
    \label{fig:sdv_rotation}%
\end{figure*}

\begin{figure}[t]
    \centering
    {
    \begin{tabular} {c | c}
        \rotatebox[origin=c]{90}{MTP} & \raisebox{-0.5\height}{\includegraphics[width=0.9\columnwidth]{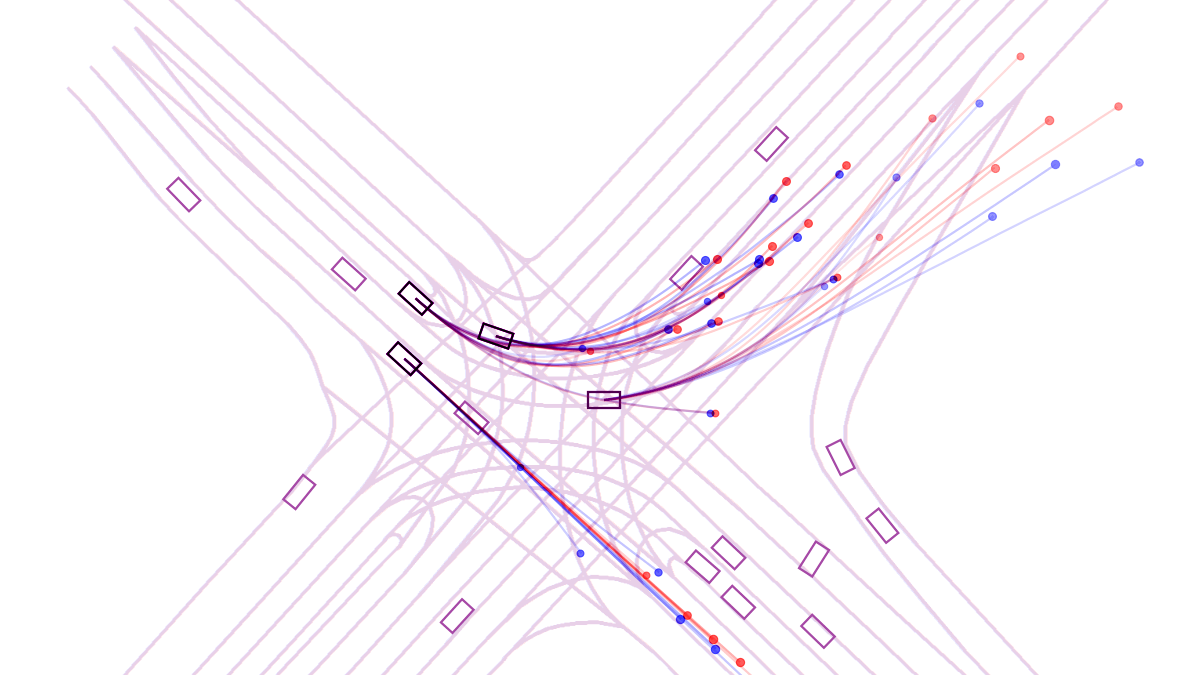}} \\
        \rotatebox[origin=c]{90}{LaneGCN} & \raisebox{-0.5\height}{\includegraphics[width=0.9\columnwidth]{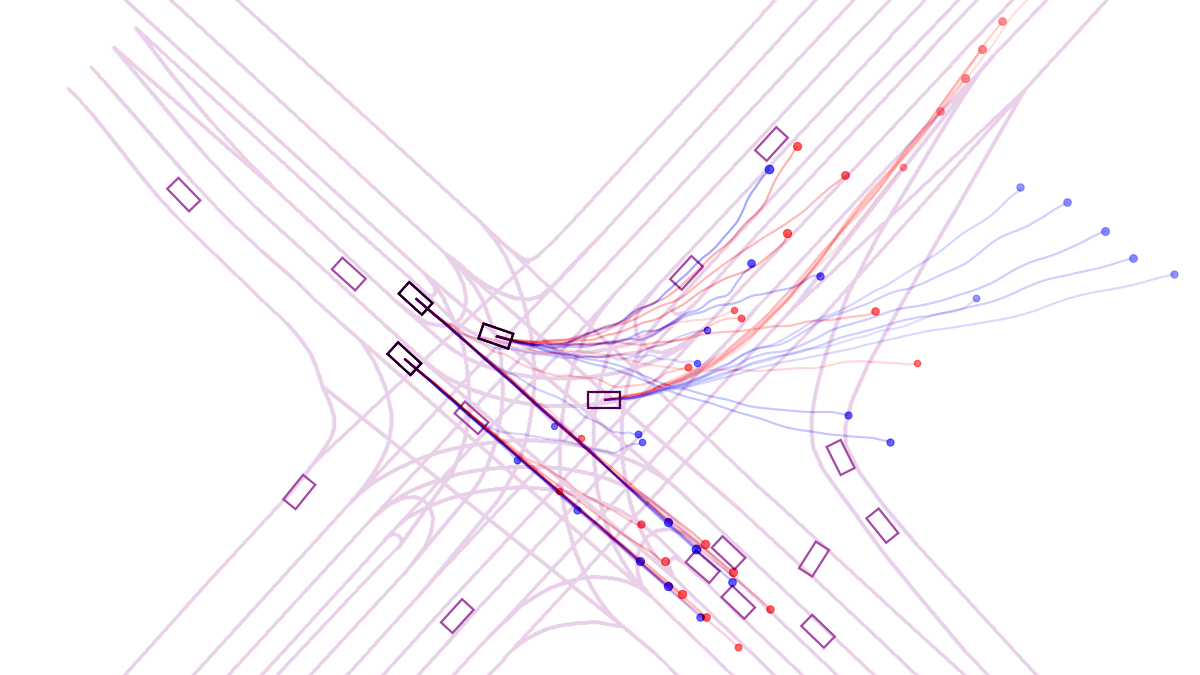}} \\
        \rotatebox[origin=c]{90}{SceneTransformer} & \raisebox{-0.5\height}{\includegraphics[width=0.9\columnwidth]{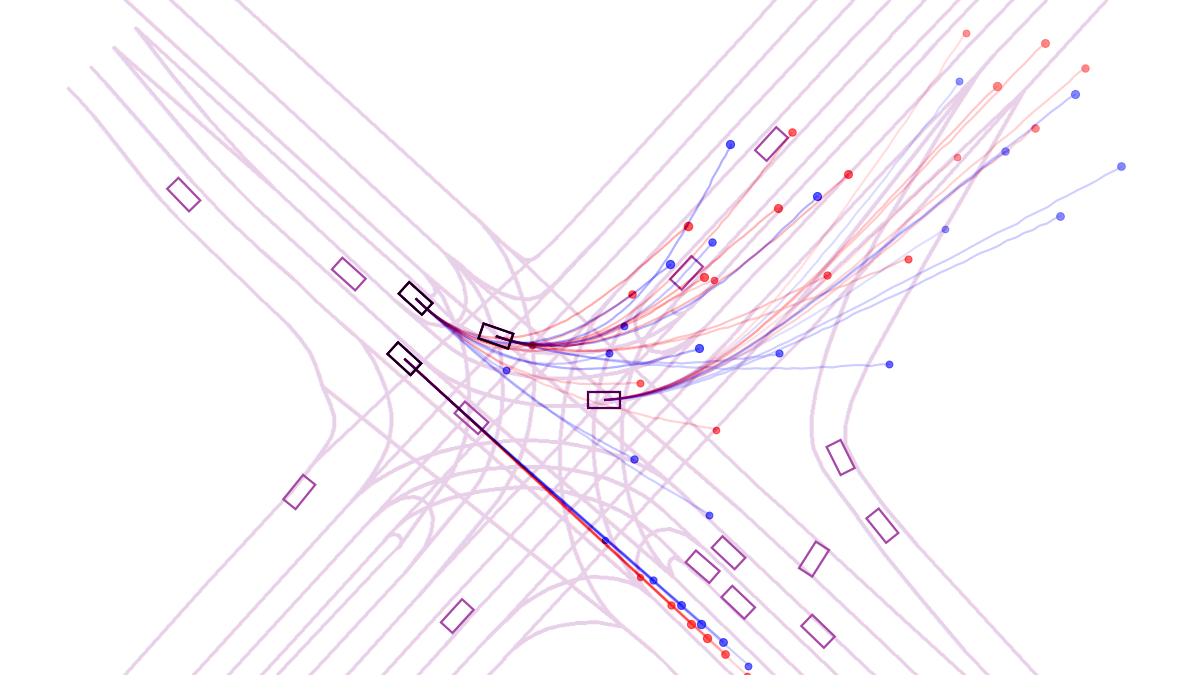}} \\
        \rotatebox[origin=c]{90}{GoRela} & \raisebox{-0.5\height}{\includegraphics[width=0.9\columnwidth]{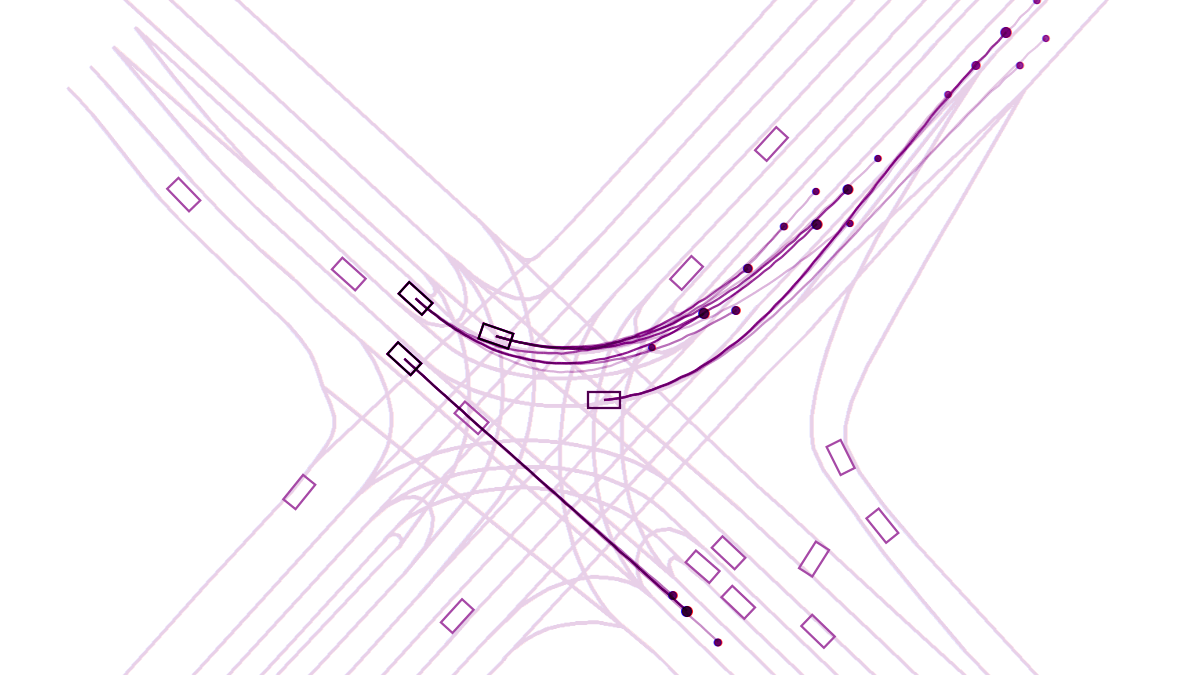}} \\
    \end{tabular}
    }
    
    \caption{\textbf{Qualitative results on viewpoint sensitivity}.
    Overlaid predictions when performing inference with the  {\color{Red} original scenario} vs.  {\color{RoyalBlue} rotated scenario by $90^{\circ}$.}
    }
    \vspace{-10pt}
    \label{fig:sdv_rotation_qualitative}
\end{figure}

\vspace{0.2cm}
\noindent{\bf Viewpoint sensitivity analysis:}
Self-driving is a safety-critical application. As such, it is important to evaluate the robustness of autonomy components with respect to rarely seen vehicle states. 
In this experiment, we simulate the self-driving vehicle (SDV) in rare states by shifting its  heading from its original pose recorded in the log (i.e., rotating the scene around the ego vehicle centroid).
We divide the scenarios in the validation set of Argoverse 2 into 8 rotation buckets $\{[0^{\circ}, 45^{\circ}), [45^{\circ}, 90^{\circ}), \dots, [315^{\circ}, 360^{\circ})\}$
In particular, Fig. \ref{fig:sdv_rotation}-Left shows the bucketed BrierMinFDE@K=6.
We can observe that \ourmodel{}'s performance is almost constant across buckets.
We believe the small offsets are due to different scenarios being on every bucket.
On the other hand, the baselines (MultiPath\cite{chai2019multipath}, LaneGCN\cite{lgn}, MTP\cite{cui2018multimodal}, Multipath \cite{chai2019multipath}, SceneTransformer\cite{ngiam2021scene}) have a much higher variability over different rotation buckets. 
To make sure this effect comes from the rotation invariance of our method and not our high-level architecture, we include the ablation of \ourmodel{} without PairPose, which preserves the same architecture but doesn't use relative positional encoding on the edges of the graphs.
We include a box plot in Fig. \ref{fig:sdv_rotation}-Right, which directly displays the variance across buckets.
Finally, we showcase this qualitatively in Fig. \ref{fig:sdv_rotation_qualitative}. It is obvious that the baselines' predictions when performing inference on the original vs. the rotated scenarios are very inconsistent. In contrast, \ourmodel{}'s predictions are exactly the same thanks to its viewpoint invariance.

\begin{figure}[t]
    \centering
    \includegraphics[width=0.9\columnwidth]{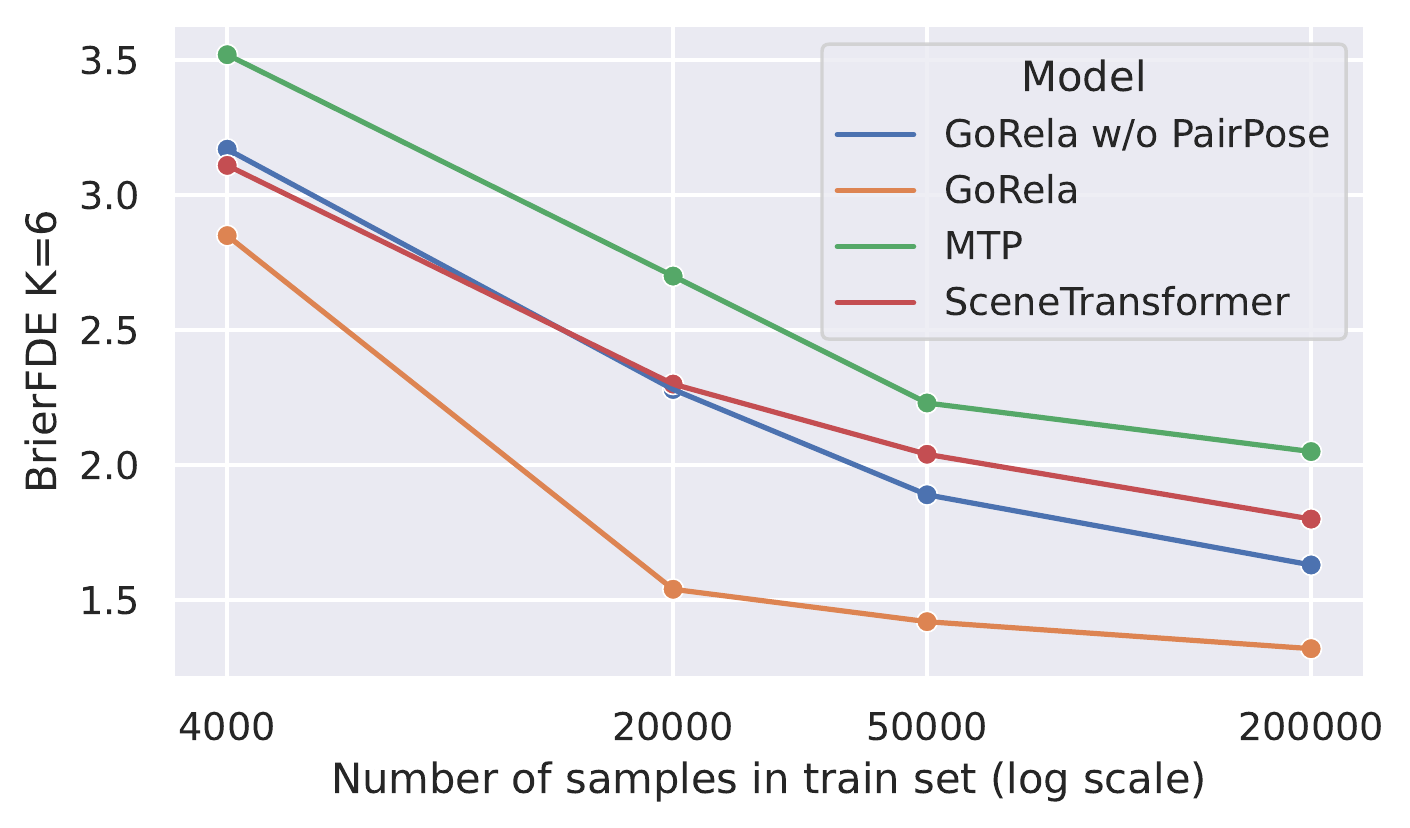}
    \caption{\textbf{Sample efficiency analysis}. We evaluate the performance of \ourmodel{} and the baselines when training with different data scales.
    }
    \vspace{-8pt}
    \label{fig:sample_efficiency}
\end{figure}

\vspace{0.2cm}
\noindent{\bf Sample efficiency curves:}
Fig. \ref{fig:sample_efficiency} showcases the sample efficiency of the different methods.
In this experiment, we train \ourmodel{} and some interesting baselines  for different training set sizes (we do not train all of them as these experiments are computationally costly).
We can observe that \ourmodel{} converges much faster than the baselines, attaining better performance than the baselines even when using 90$\%$ less data (only 20,000 examples).
In other words, the baselines are a lot more data hungry, and it seems that they would greatly
benefit from using more than 200,000 examples.
Sample efficiency in motion forecasting is very important since labeling these datasets for supervised learning is very costly.
We also ablate the importance of PairPose by removing it from our model, which makes it very clear that our viewpoint-invariant approach with pair-wise relative positional encodings is really the key to this improvement in sample efficiency.
Intuitively, this result aligns with our expectations, since the rotation and translation invariance of our method shrinks the input space quite a bit, making scenarios in the validation set look more similar to those in the training set.

\begin{figure}[t]
    \centering
    \vspace{-10pt}
    \includegraphics[width=0.7\columnwidth]{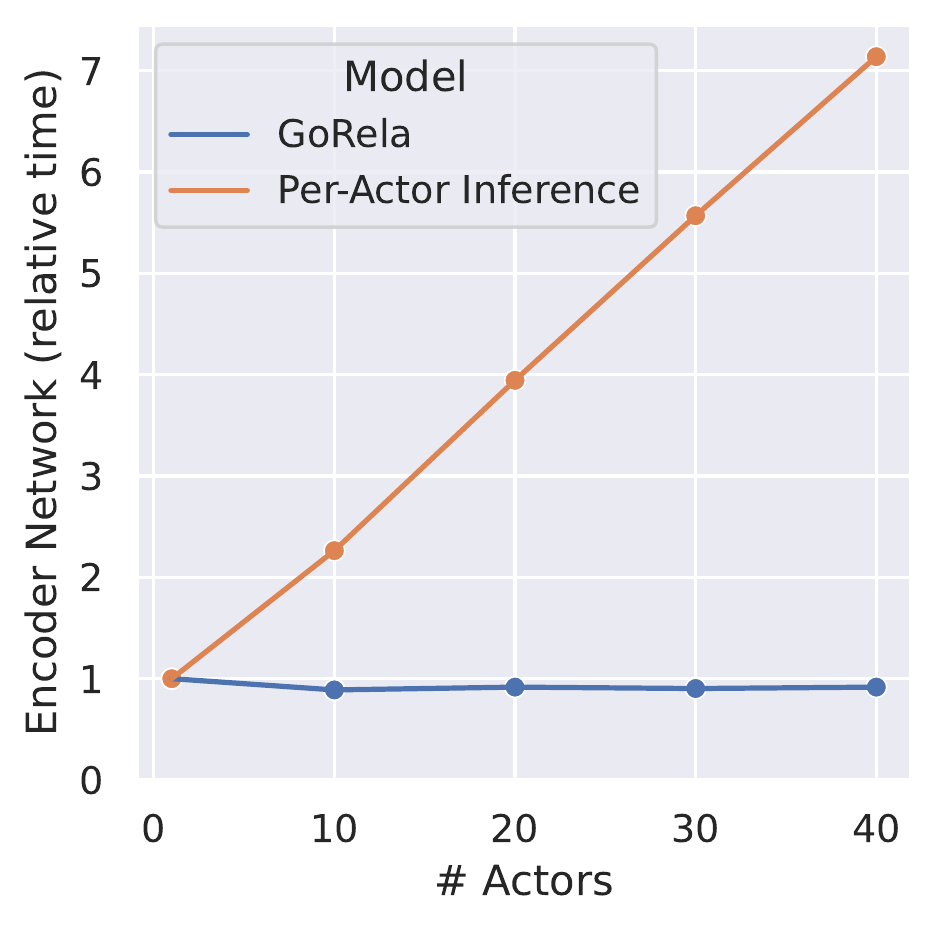}
    \caption{\textbf{Runtime scaling with actors analysis}. We measure how the encoder runtime scales with actors where we share a scene encoding across all actors (\ourmodel{}) and where the scene-context is encoded per actor (batched across actors). The runtime is measured relative to the runtime with a single actor. The number of lane graph nodes (250) is held constant.
    }
    \vspace{-8pt}
    \label{fig:runtime_actor_enc}
\end{figure}
\begin{figure}[t]
    \centering
    \vspace{-10pt}
    \includegraphics[width=0.7\columnwidth]{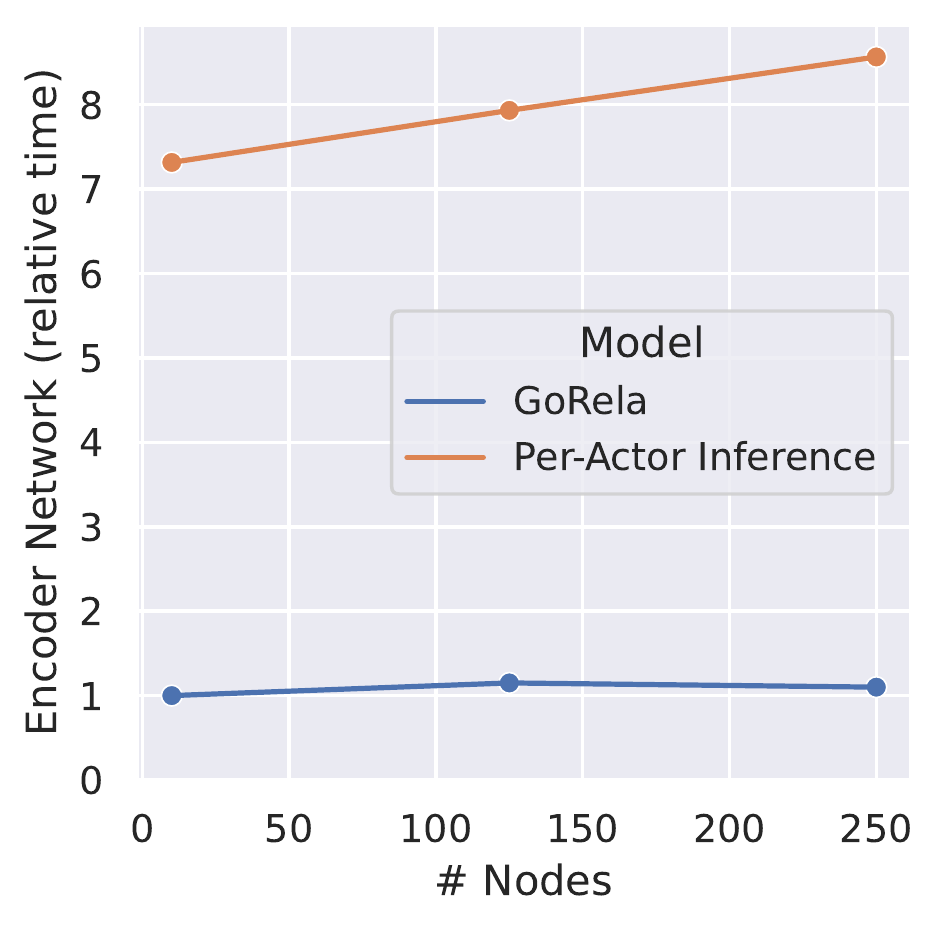}
    \caption{\textbf{Runtime scaling with lane graph nodes analysis}. We measure how the encoder runtime scales with the number of lane graph nodes, where we share a scene encoding across all actors (\ourmodel{}) and where the scene-context is encoded per actor (batched across actors). The runtime is measured relative to the runtime with 10 nodes using shared scene encoding. The number of actors (40) is held constant.
    }
    \vspace{-8pt}
    \label{fig:runtime_node_enc}
\end{figure}

\vspace{0.2cm}
\noindent{\bf Runtime analysis:}
In this experiment, we measure the benefit of a shared scene encoding for all agents as achieved by \ourmodel{} against processing the scene separately for each agent (batched in GPU).
We measure how the encoder runtime scales with number of agents in Fig. \ref{fig:runtime_actor_enc} as well as map nodes in the graph in Fig. \ref{fig:runtime_node_enc}. 
The encoder here includes all up to the heterogeneous scene encoder (inclusive). 
We can see that our model has nearly constant runtime regardless of the number of actors due to our shared scene encoding. This is because the number of agents and their inbound and outbound connections in the graph is relatively low compared to the map nodes.
On the other hand, the per-actor scene encoding increases the runtime linearly to 7x with 40x more actors. 
We can see that both models scale relatively slowly with the number of lane graph nodes, with per-actor inference starting with a much higher fixed cost.
Overall, these results demonstrate the runtime benefits of using a shared scene encoding.
We do not compare the decoder runtime between the two approaches as it would be the same given that \ourmodel{}'s decoder acts on a graph composed of one connected component per agent.

\vspace{0.2cm}
\noindent{\bf Graph layer ablation:}
Table \ref{table:gnn_ablation_argoverse} shows the results of the ablation of heterogeneous message passing (HMP) as our graph convolutional layer. This layer is used in the heterogeneous scene encoder and per-actor goal decoder. We based our experiment on the public HEAT \cite{mo2022multi} and the GATv2 \cite{brody2021attentive} implementations in \cite{Fey/Lenssen/2019}. 
We show that our model outperforms these state-of-the-art layers in both multimodal ($K=6$) and unimodal error.

\begin{table}[h]
    \centering
\begin{threeparttable}
\setlength\tabcolsep{5pt} %
\begin{tabularx}{\columnwidth}{ l | s s s%
                }
                \toprule
        
        Graph layer $^\dagger$ &
        BrierFDE K=6  & FDE K=6 & FDE K=1 \\

        \midrule

        HEAT \cite{mo2022multi} & 1.52 & 1.17 & 2.98 \\
        GATv2 \cite{brody2021attentive} & 1.48 & 1.13 & 2.86 \\
        HMP (ours) & $\mathbf{1.45}$ & $\mathbf{1.08}$ & $\mathbf{2.77}$ \\

        \bottomrule
\end{tabularx}
\end{threeparttable}
\caption{\textbf{Ablation study} of graph convolutional layers on Argoverse 2 (val). We ablate HMP with other state-of-the-art graph convolution layers.
$^\dagger$ Models trained on 25$\%$ of the training set. 
}
\vspace{-15pt}
\label{table:gnn_ablation_argoverse}
\end{table}

\begin{figure*}[t]
    \centering
    {
    \begin{tabular} {c | c | c | c}
        MTP & LaneGCN & SceneTransformer & Gorela \\
        \includegraphics[width=0.245\textwidth, trim={16cm, 16cm, 16cm, 12cm}, clip]{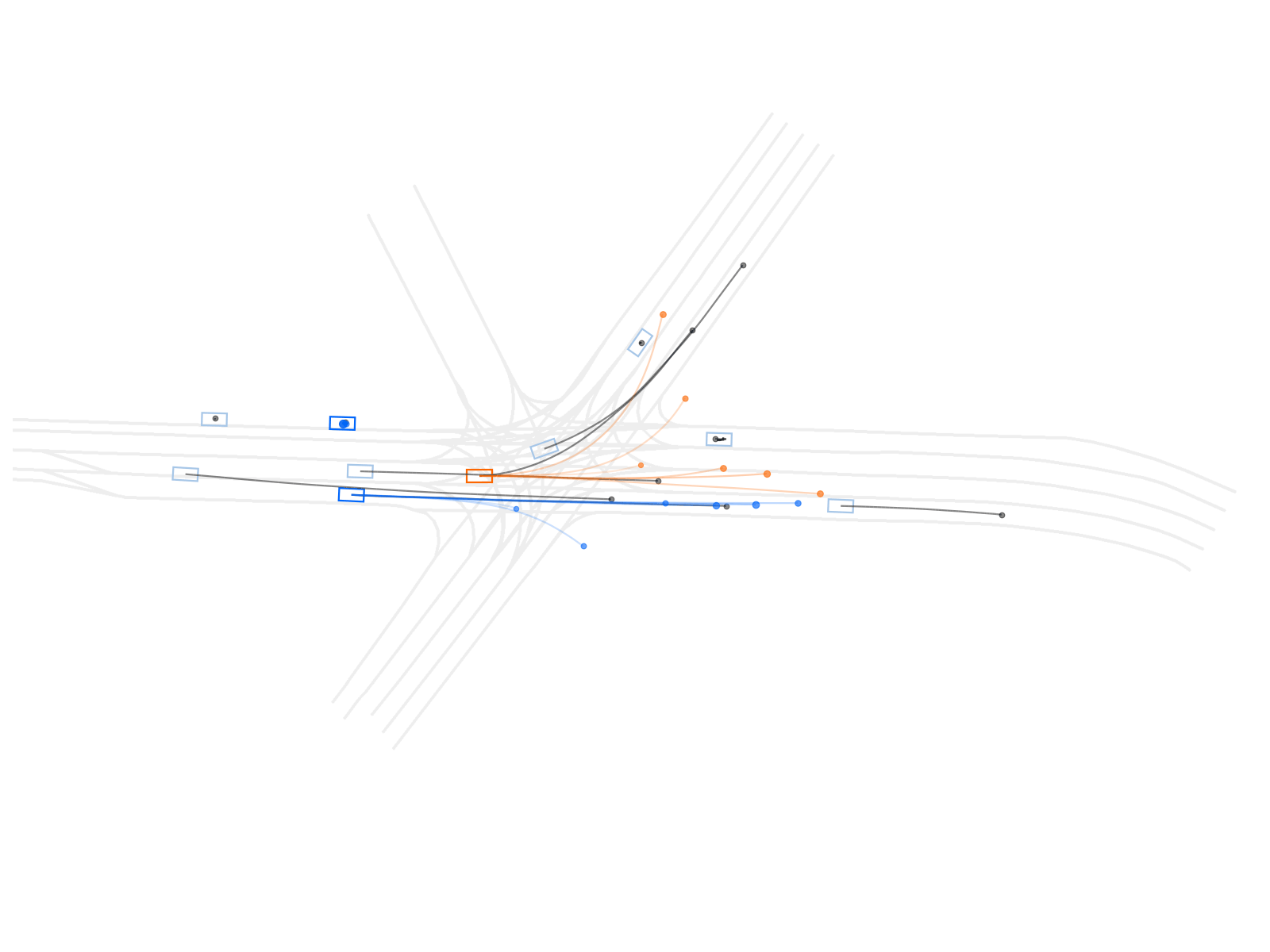} &
        \includegraphics[width=0.245\textwidth, trim={16cm, 16cm, 16cm, 12cm}, clip]{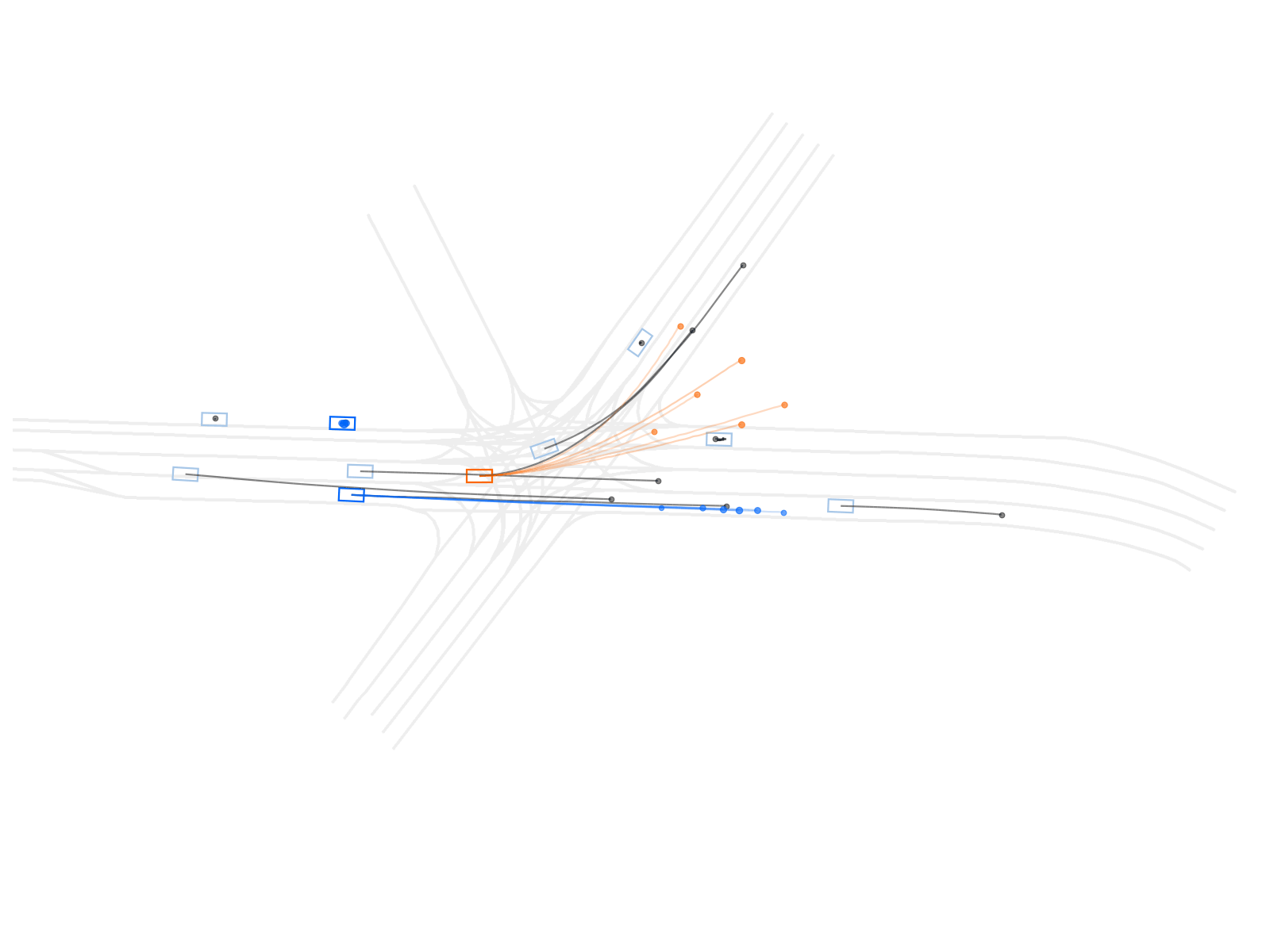} &
        \includegraphics[width=0.245\textwidth, trim={16cm, 16cm, 16cm, 12cm}, clip]{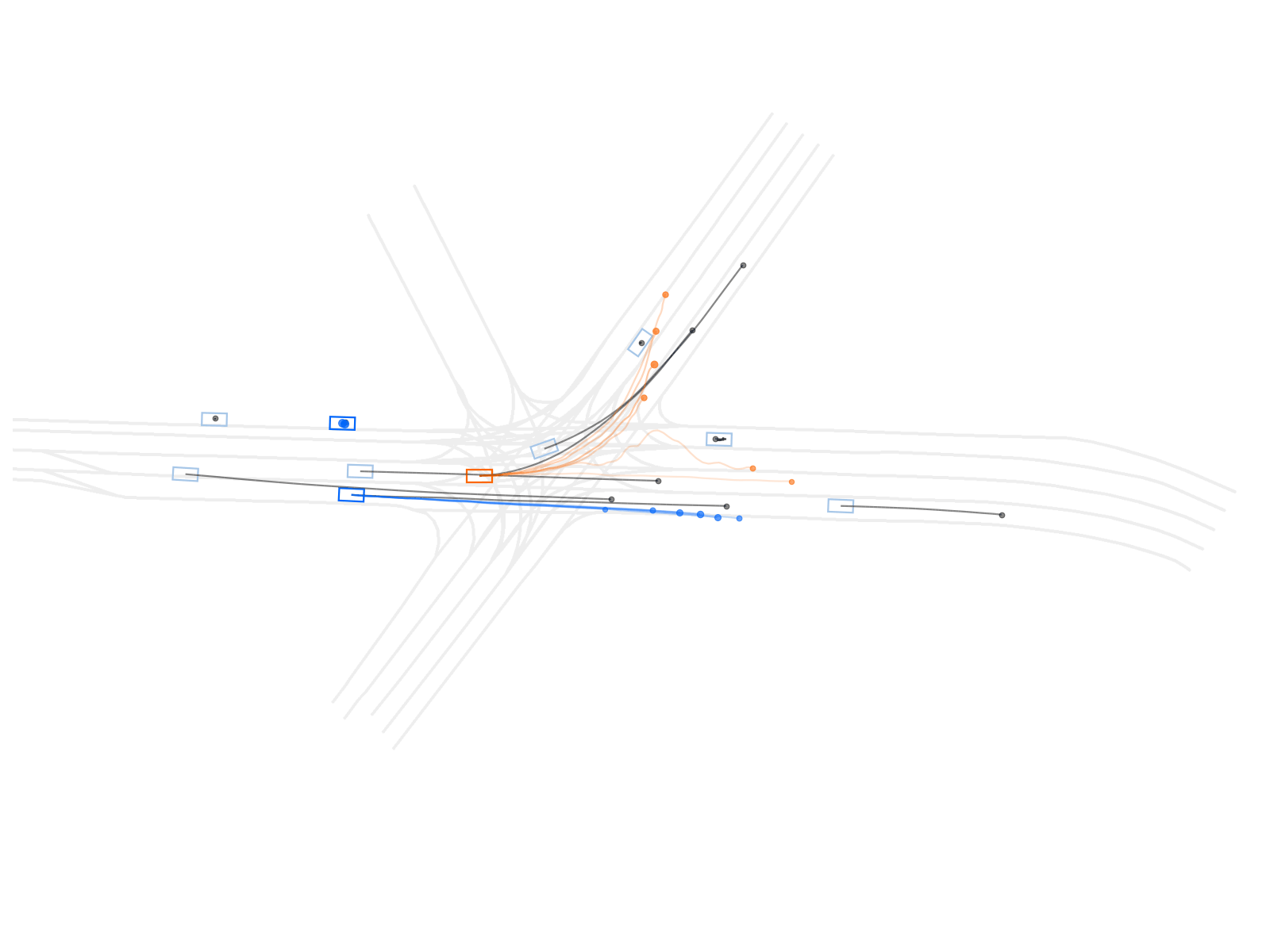} &
        \includegraphics[width=0.245\textwidth, trim={16cm, 16cm, 16cm, 12cm}, clip]{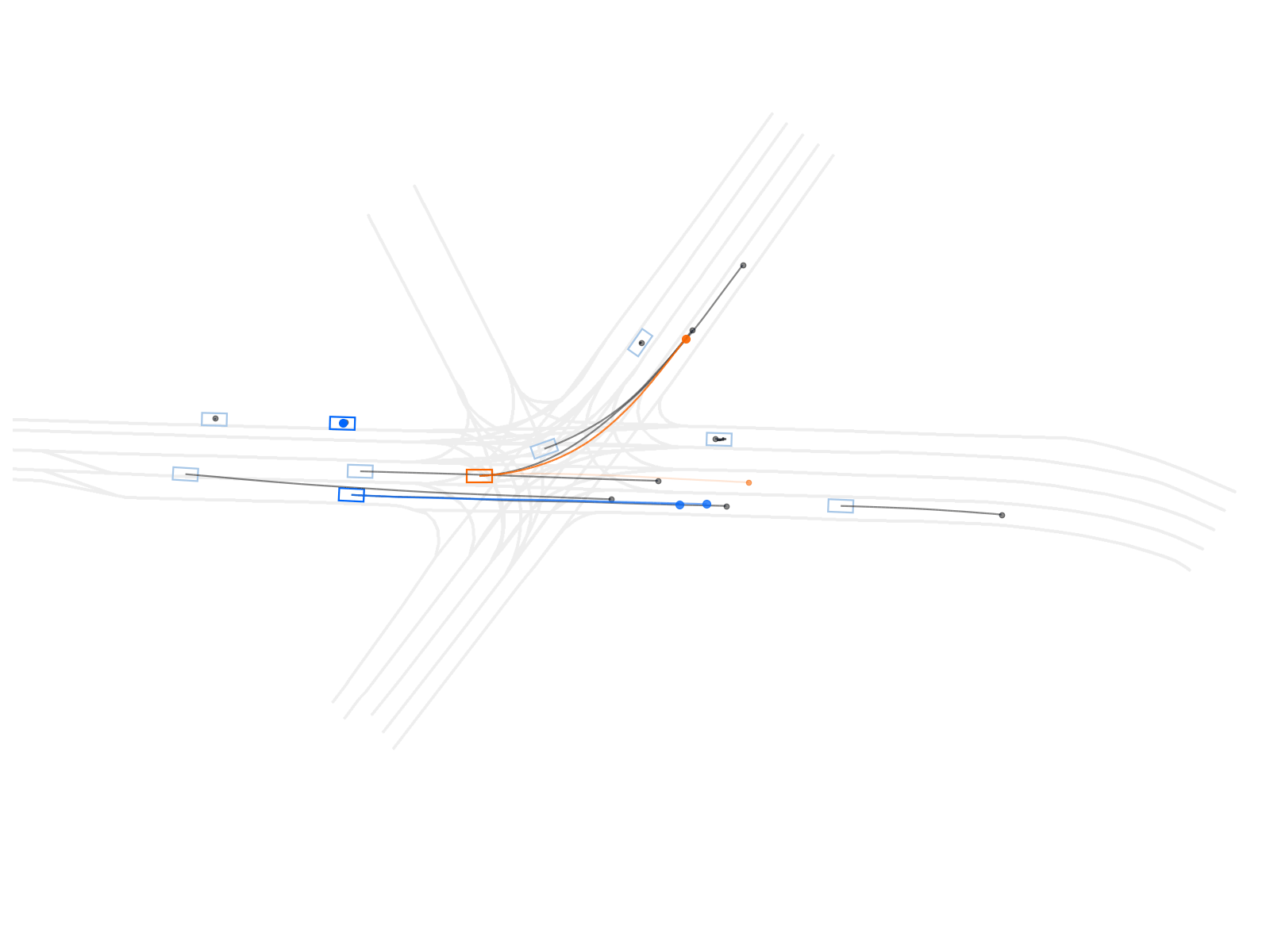} \\
        \midrule
        \includegraphics[width=0.245\textwidth, trim={16cm, 16cm, 16cm, 12cm}, clip]{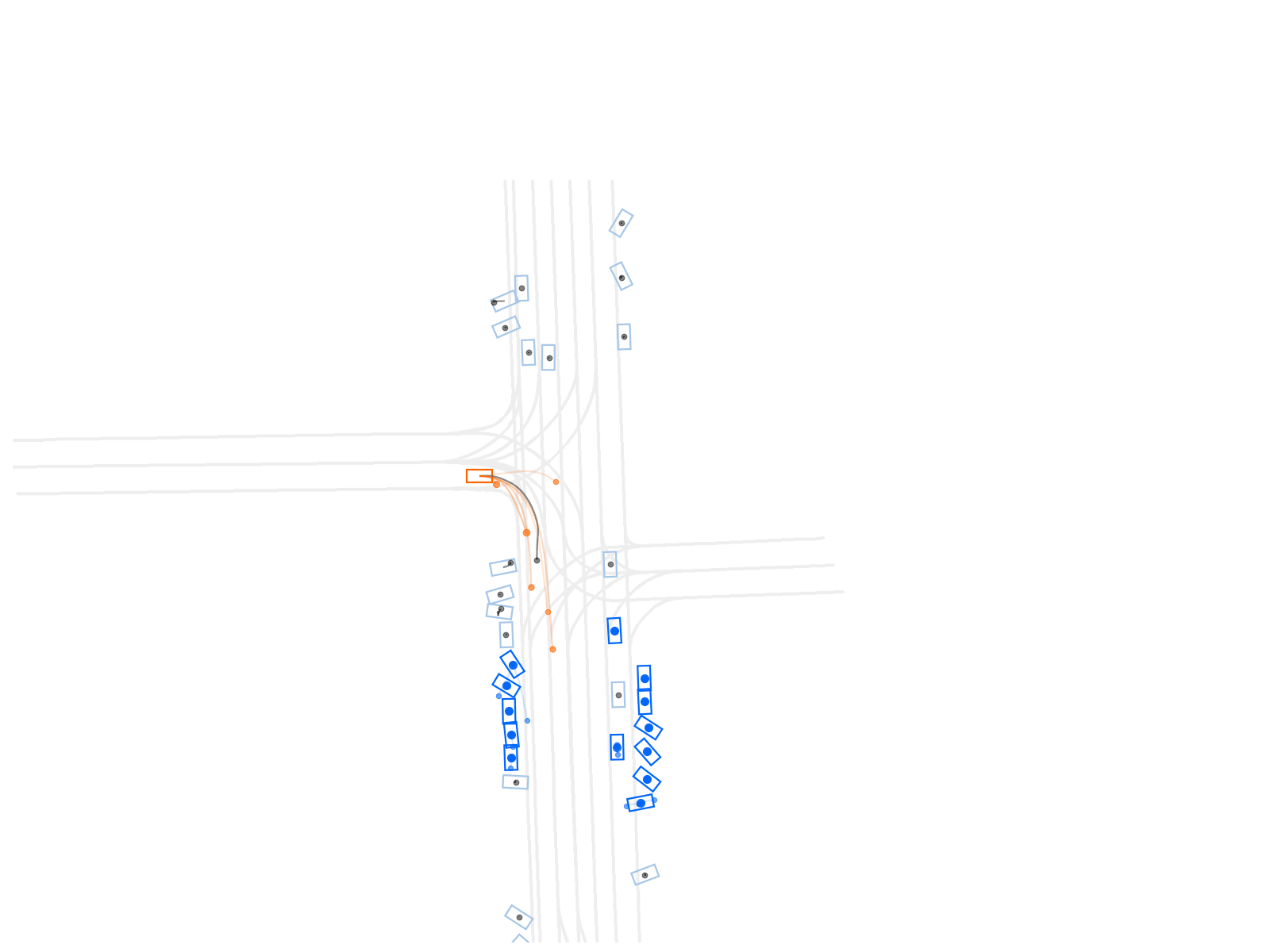} &
        \includegraphics[width=0.245\textwidth, trim={16cm, 16cm, 16cm, 12cm}, clip]{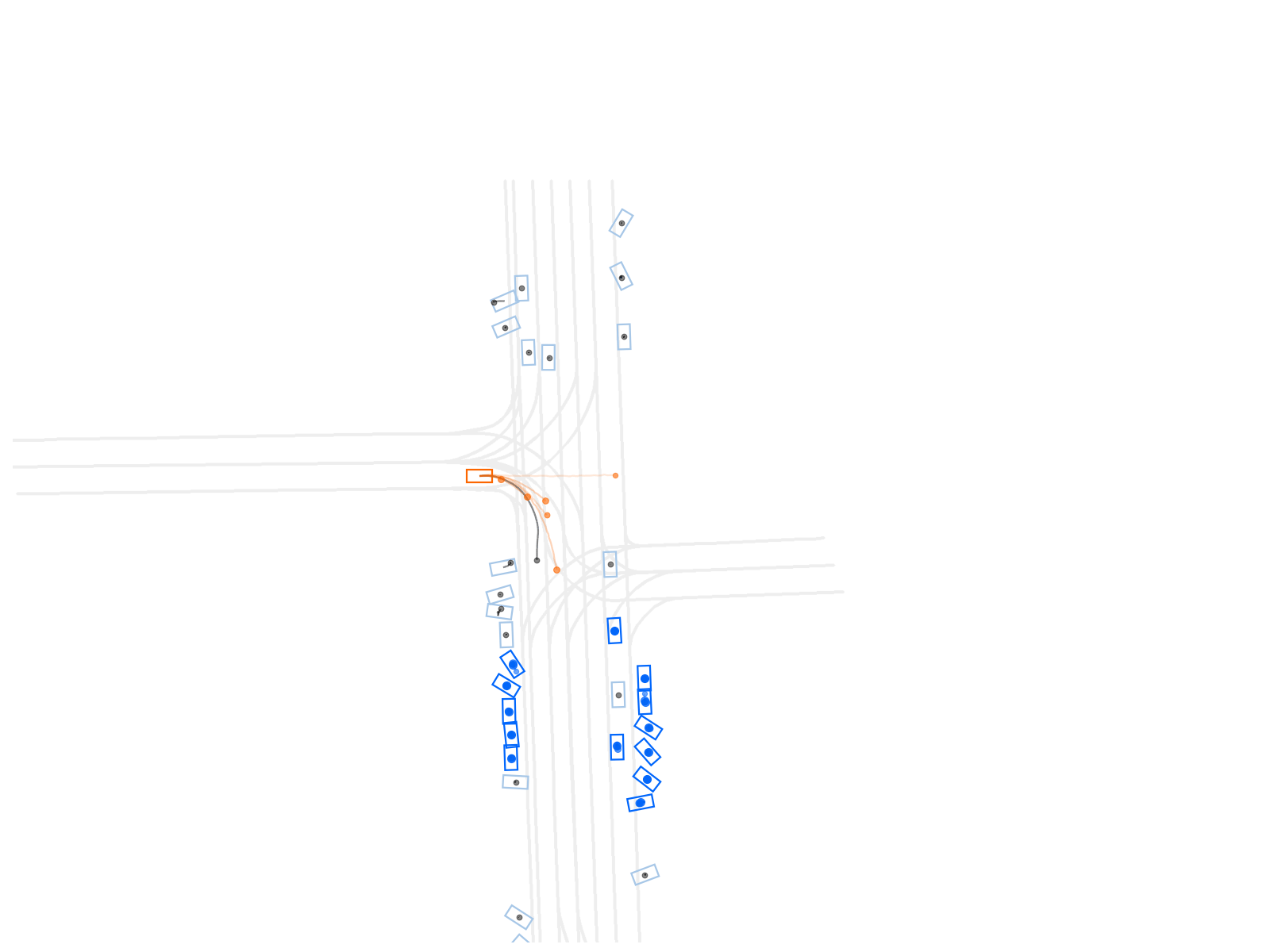} &
        \includegraphics[width=0.245\textwidth, trim={16cm, 16cm, 16cm, 12cm}, clip]{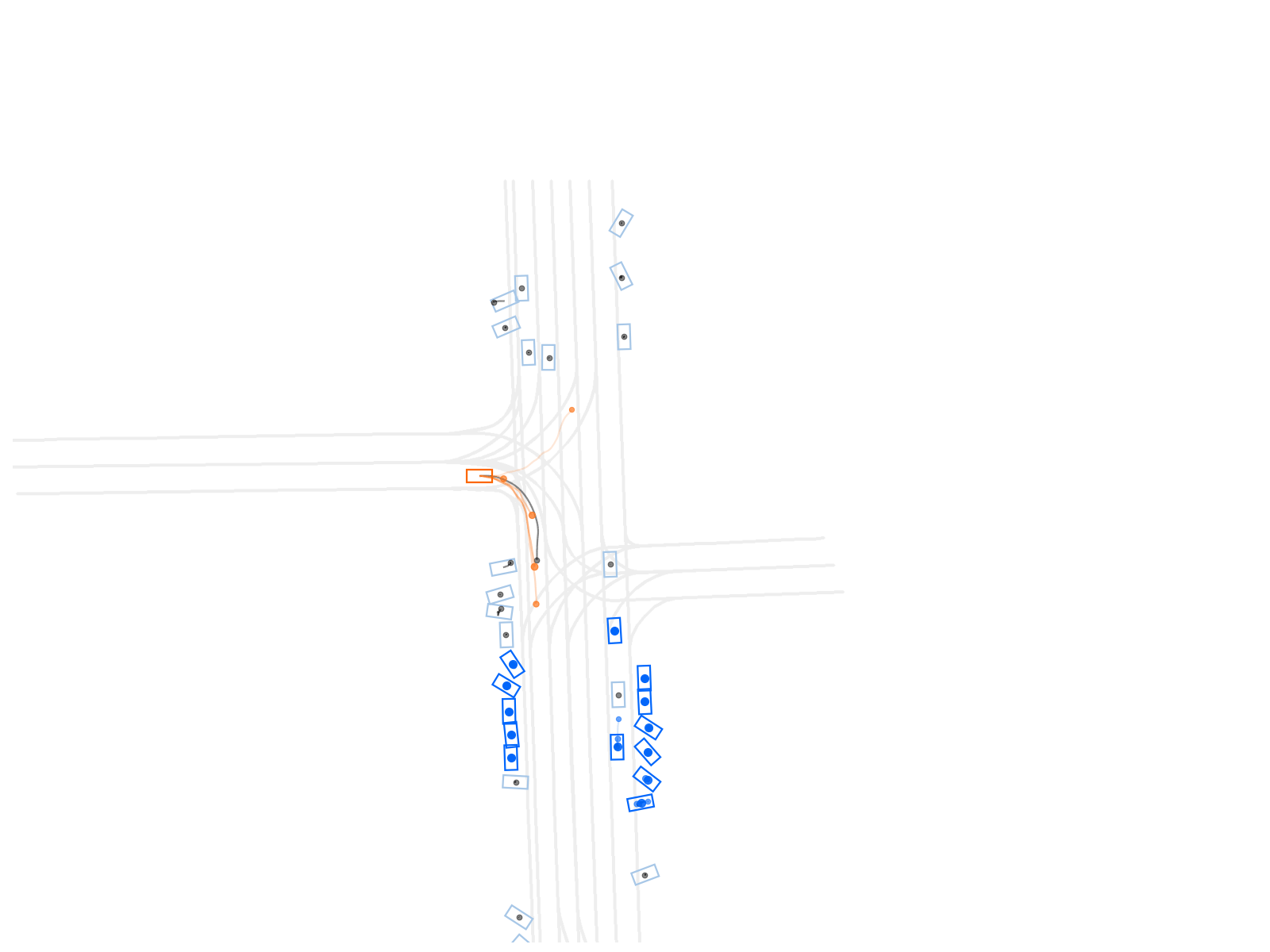} &
        \includegraphics[width=0.245\textwidth, trim={16cm, 16cm, 16cm, 12cm}, clip]{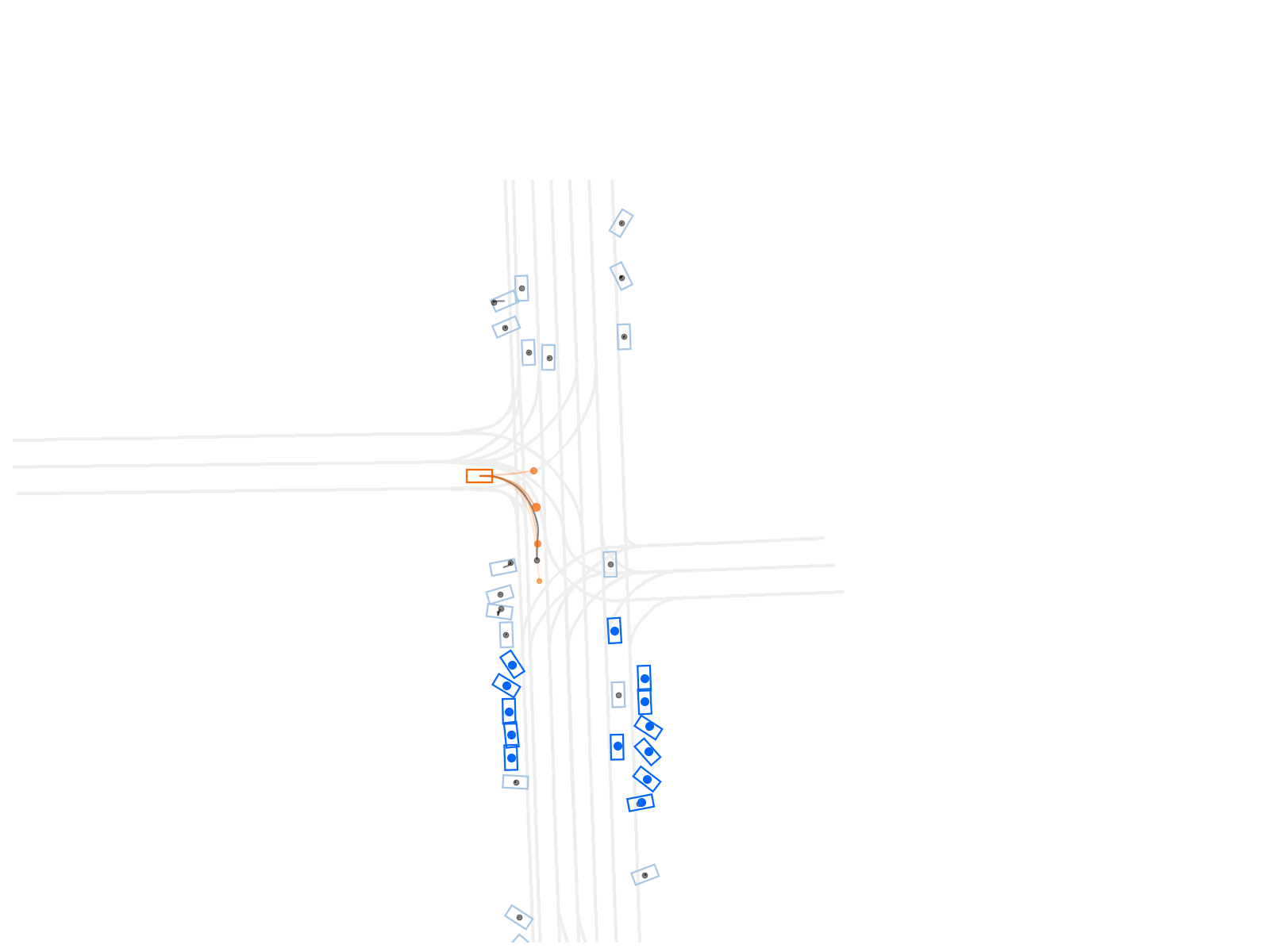} \\
        \midrule
        \includegraphics[width=0.245\textwidth, trim={16cm, 16cm, 16cm, 12cm}, clip]{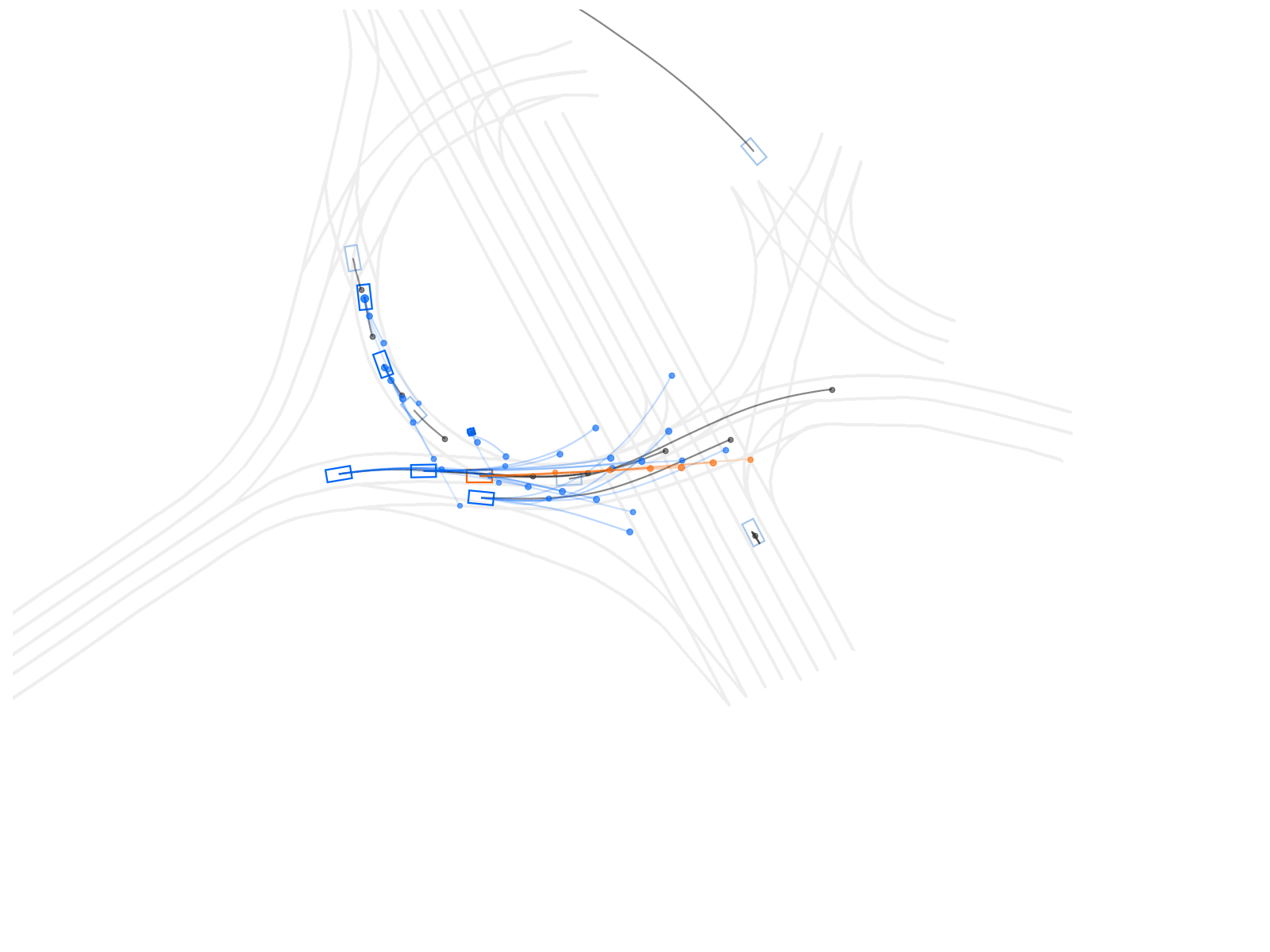} &
        \includegraphics[width=0.245\textwidth, trim={16cm, 16cm, 16cm, 12cm}, clip]{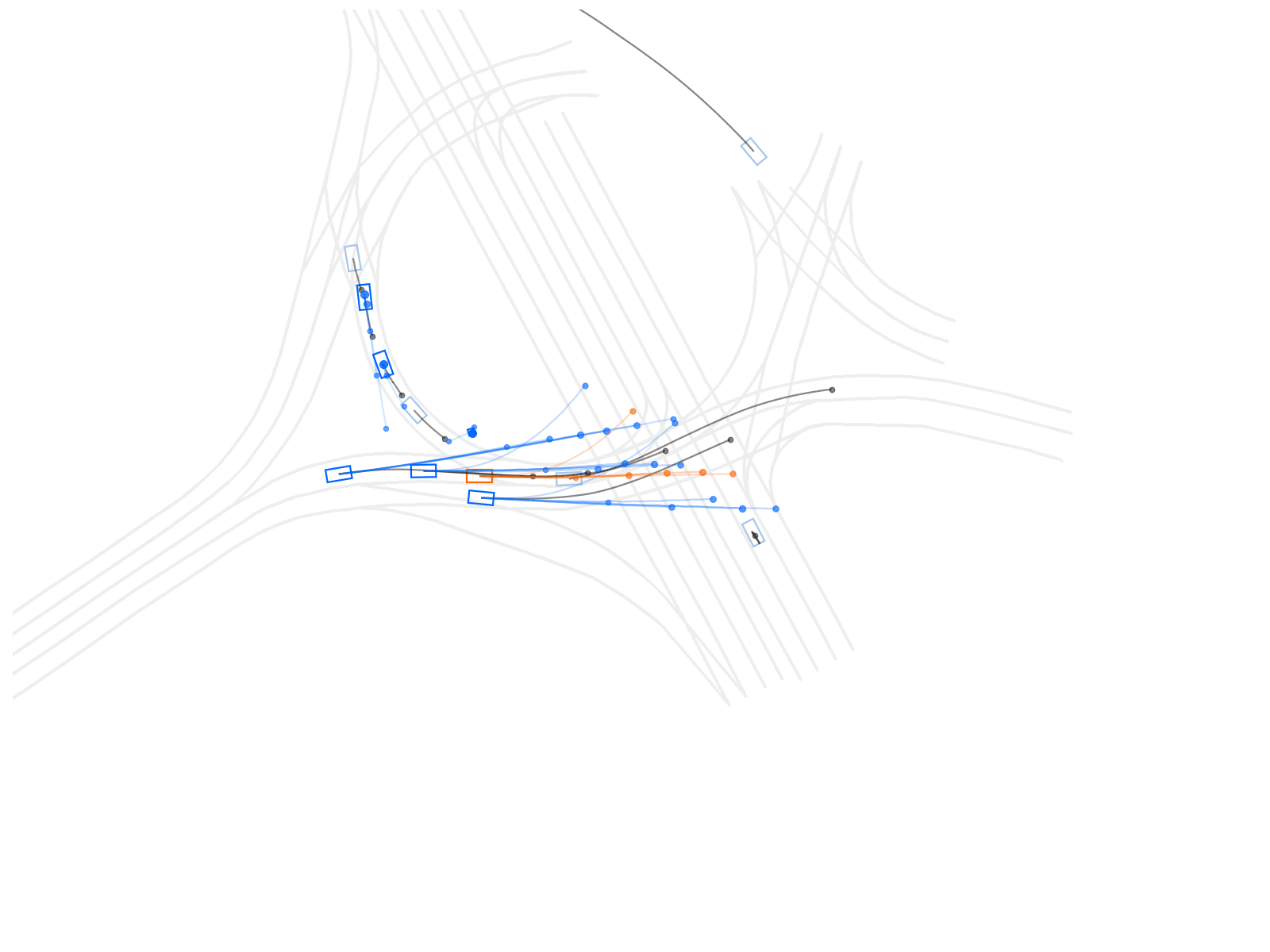} &
        \includegraphics[width=0.245\textwidth, trim={16cm, 16cm, 16cm, 12cm}, clip]{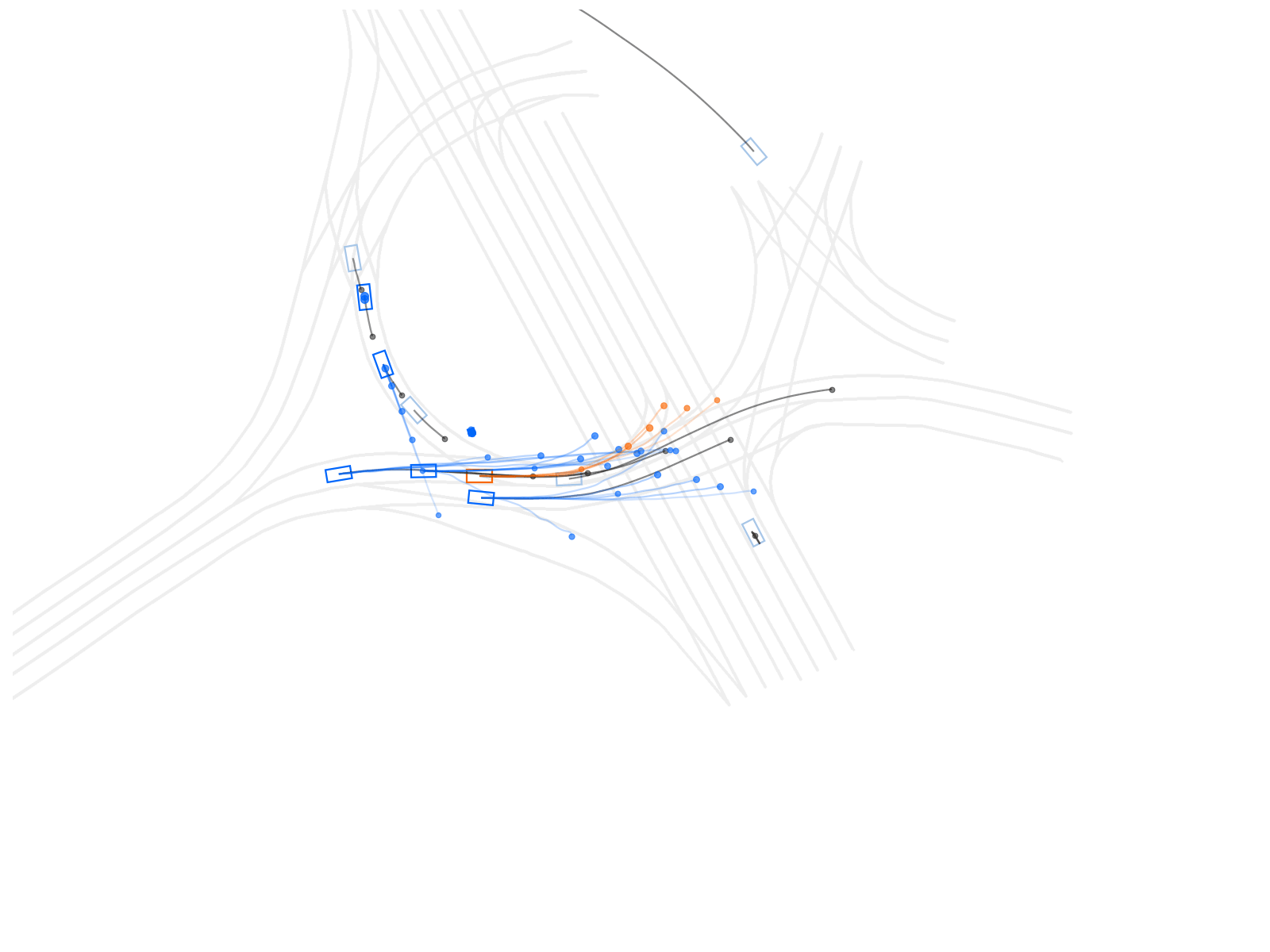} &
        \includegraphics[width=0.245\textwidth, trim={16cm, 16cm, 16cm, 12cm}, clip]{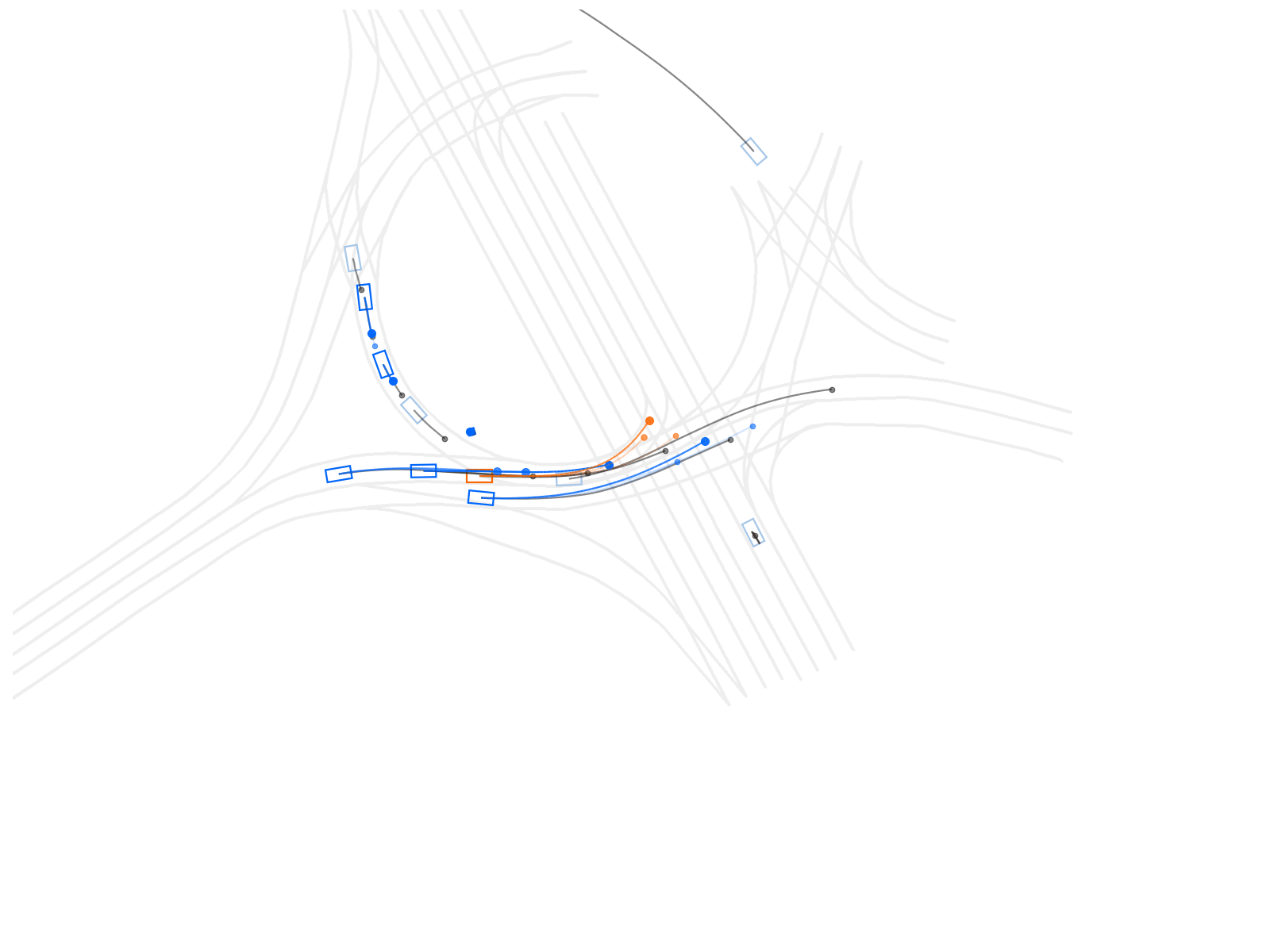} \\
        \midrule
        \includegraphics[width=0.245\textwidth, trim={20cm, 16cm, 12cm, 12cm}, clip]{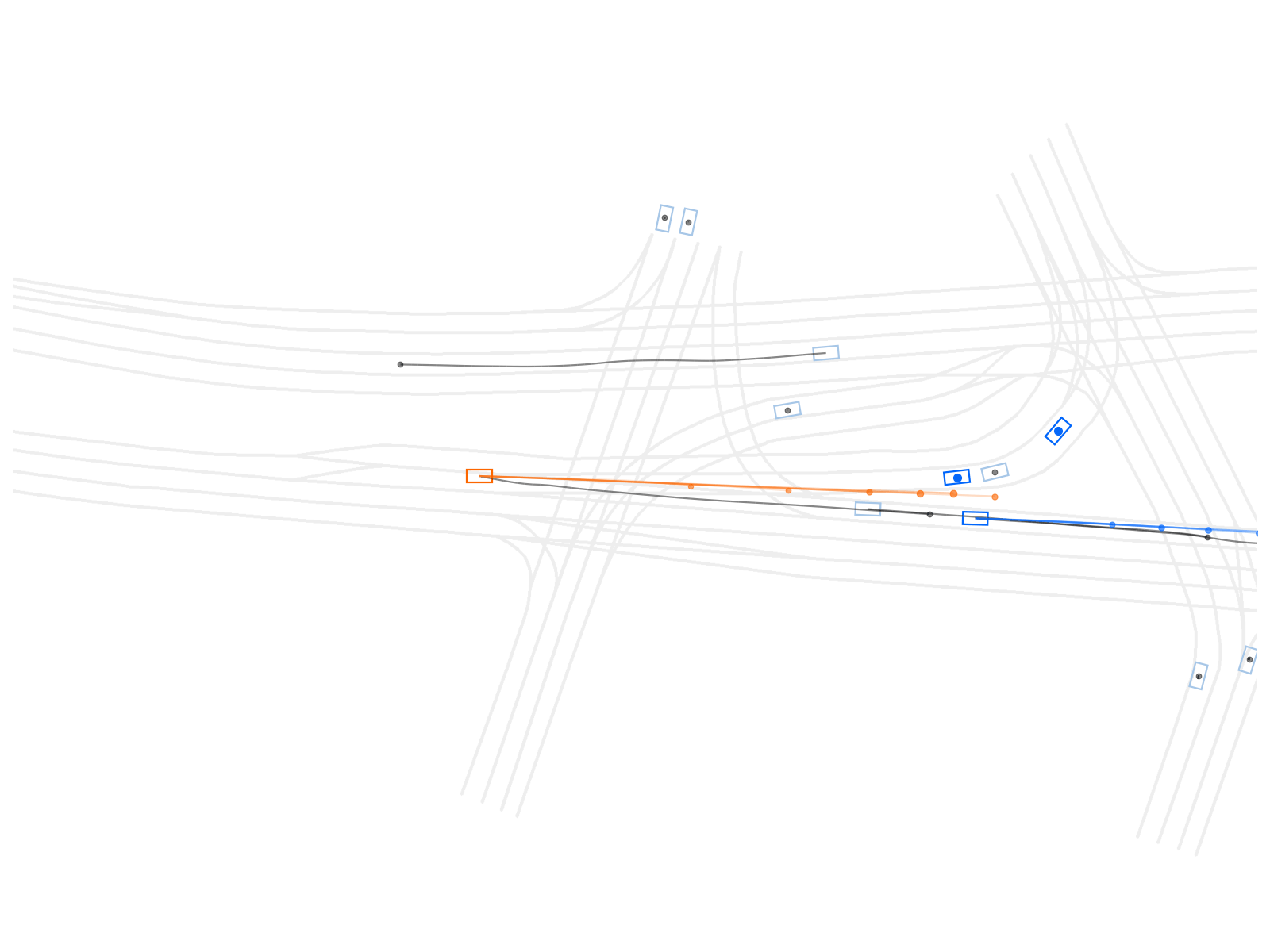} &
        \includegraphics[width=0.245\textwidth, trim={20cm, 16cm, 12cm, 12cm}, clip]{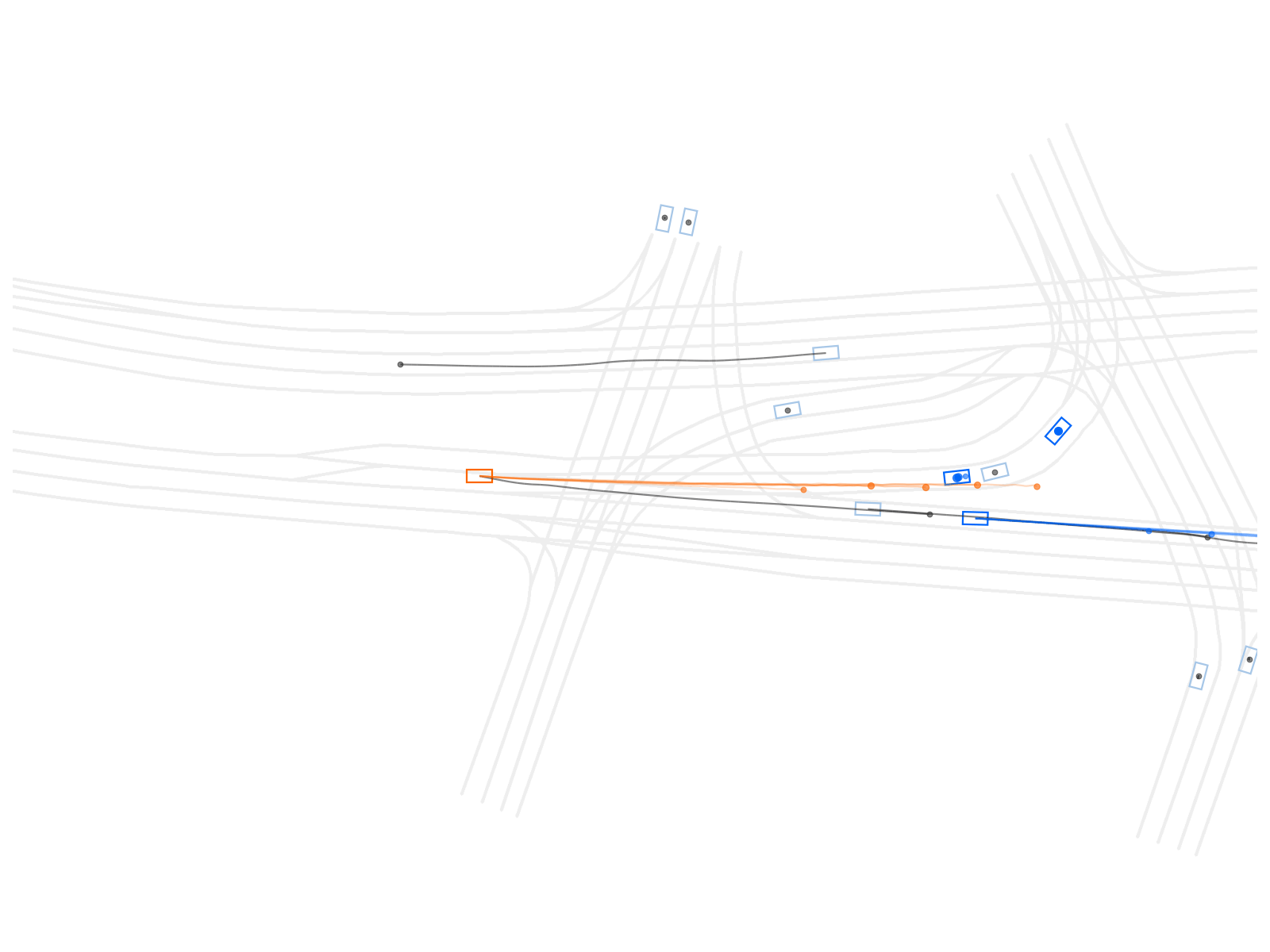} &
        \includegraphics[width=0.245\textwidth, trim={20cm, 16cm, 12cm, 12cm}, clip]{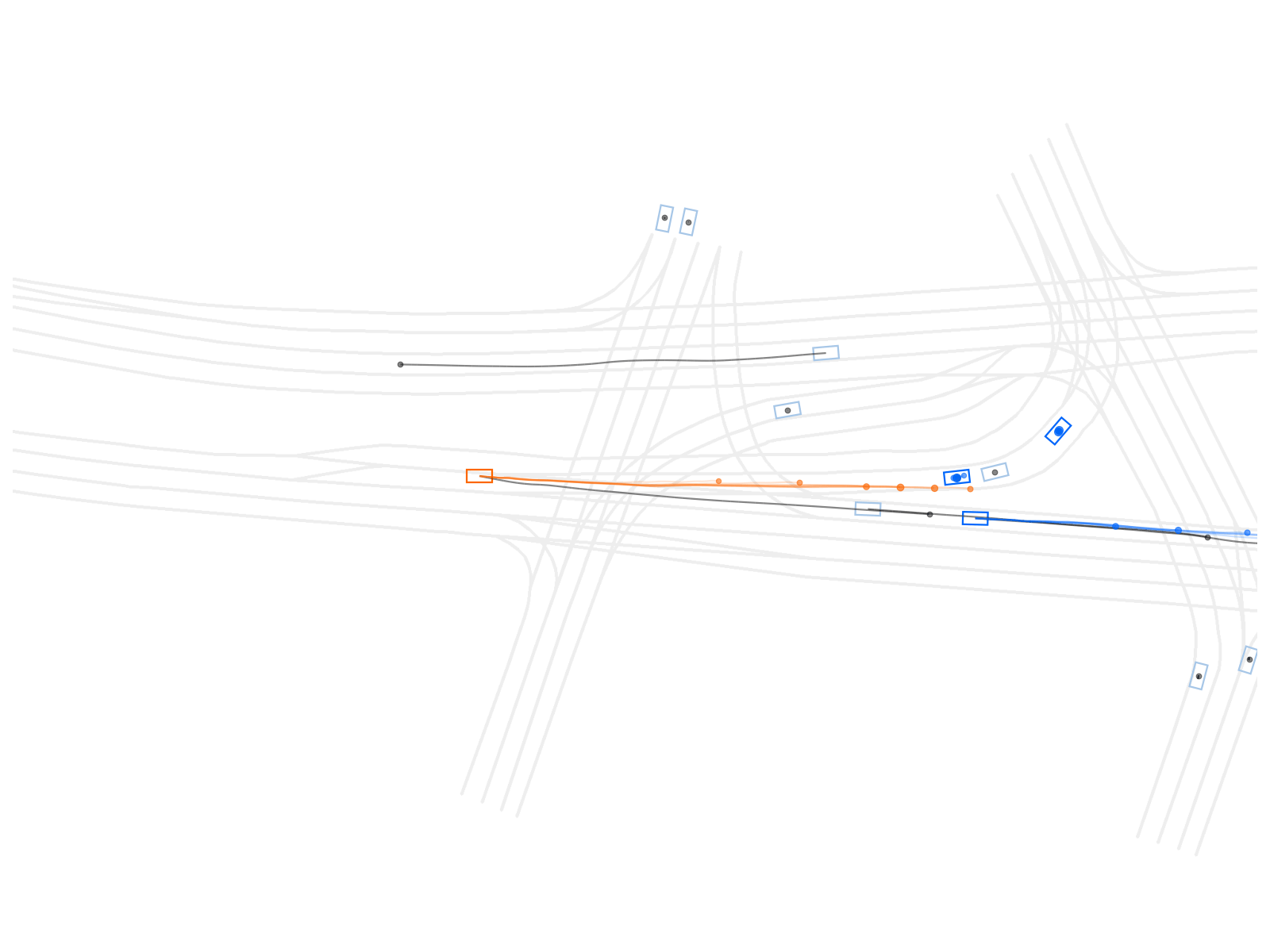} &
        \includegraphics[width=0.245\textwidth, trim={20cm, 16cm, 12cm, 12cm}, clip]{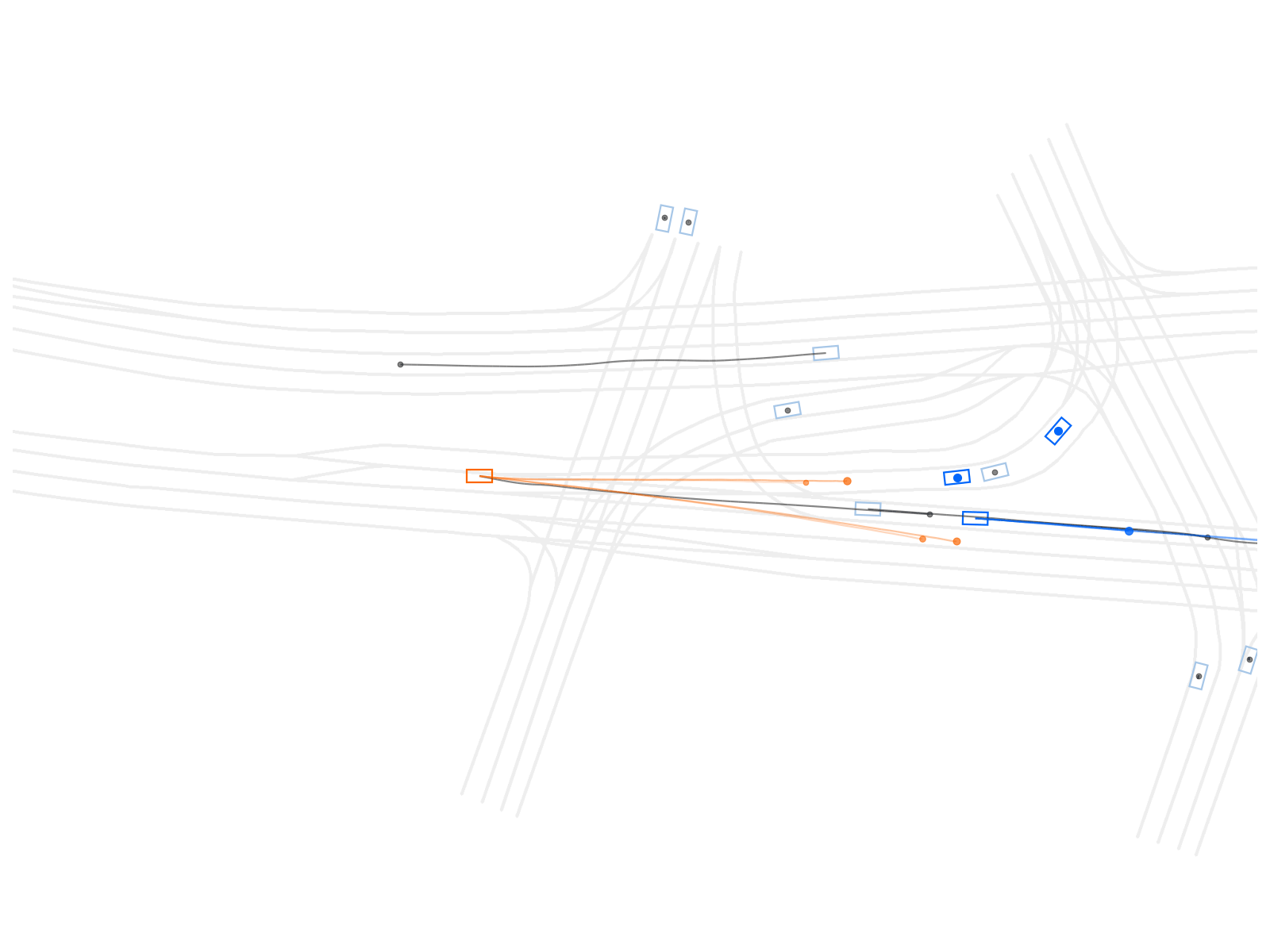} \\
        \midrule
        \includegraphics[width=0.245\textwidth, trim={16cm, 16cm, 16cm, 12cm}, clip]{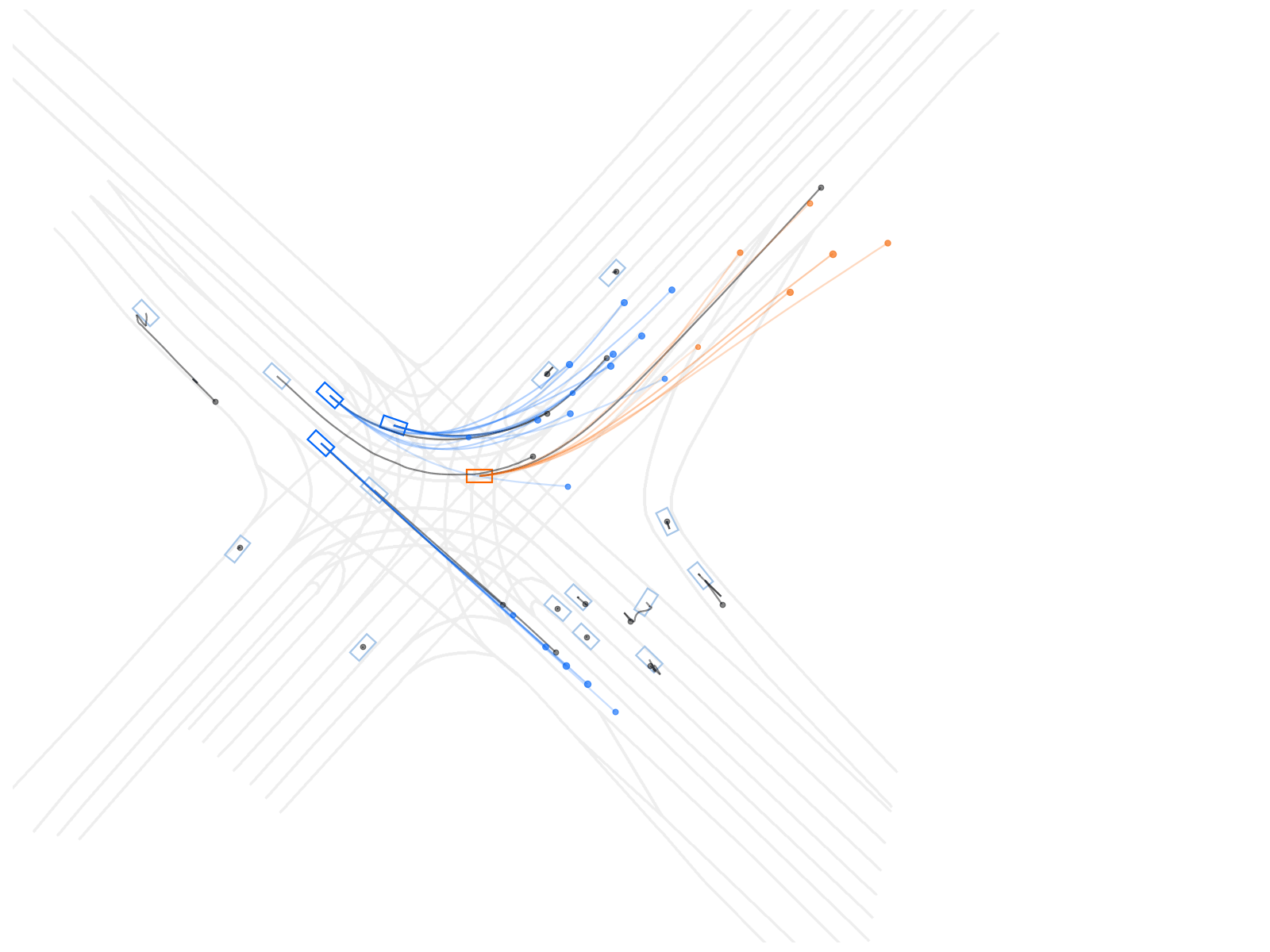} &
        \includegraphics[width=0.245\textwidth, trim={16cm, 16cm, 16cm, 12cm}, clip]{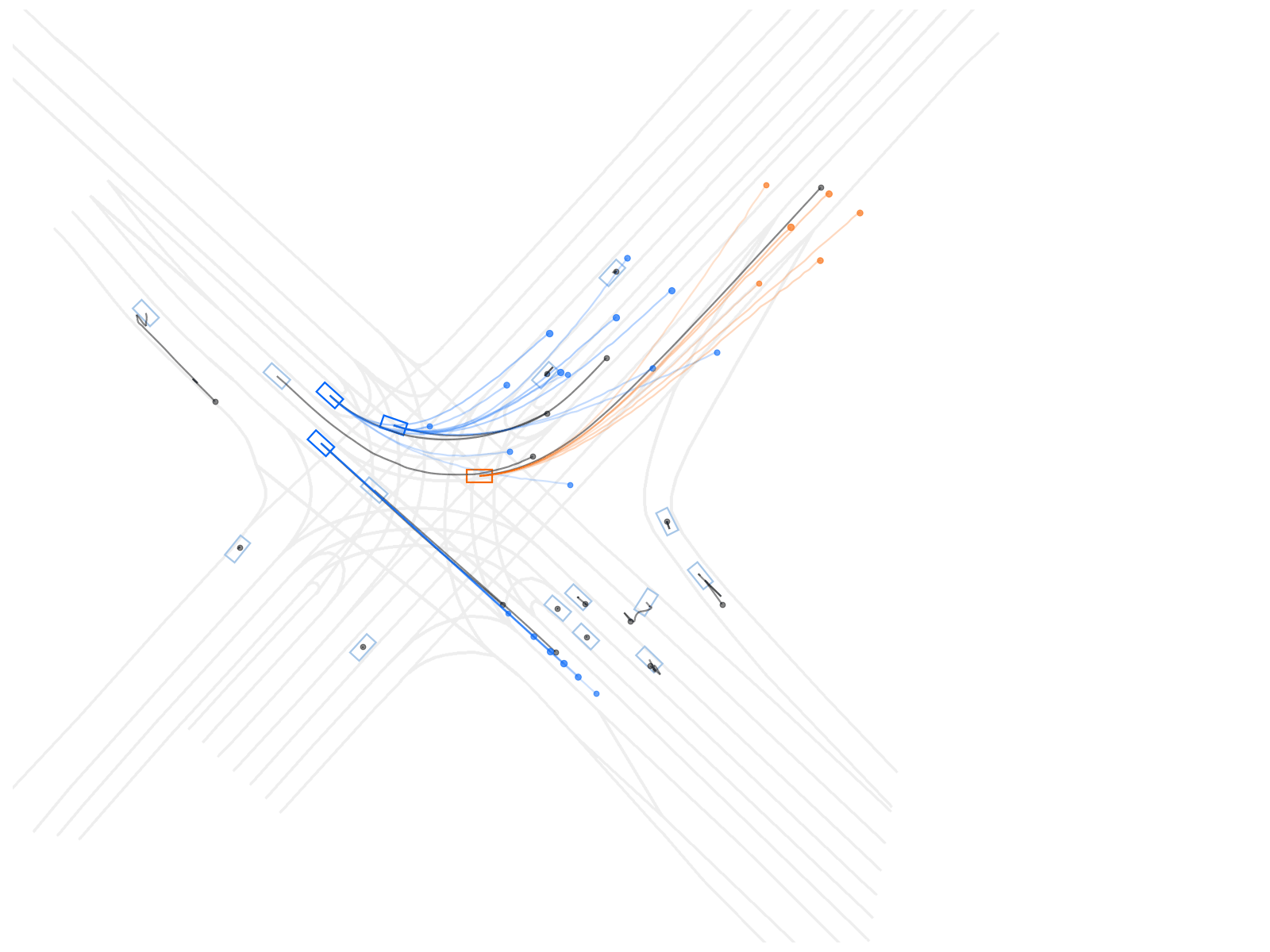} &
        \includegraphics[width=0.245\textwidth, trim={16cm, 16cm, 16cm, 12cm}, clip]{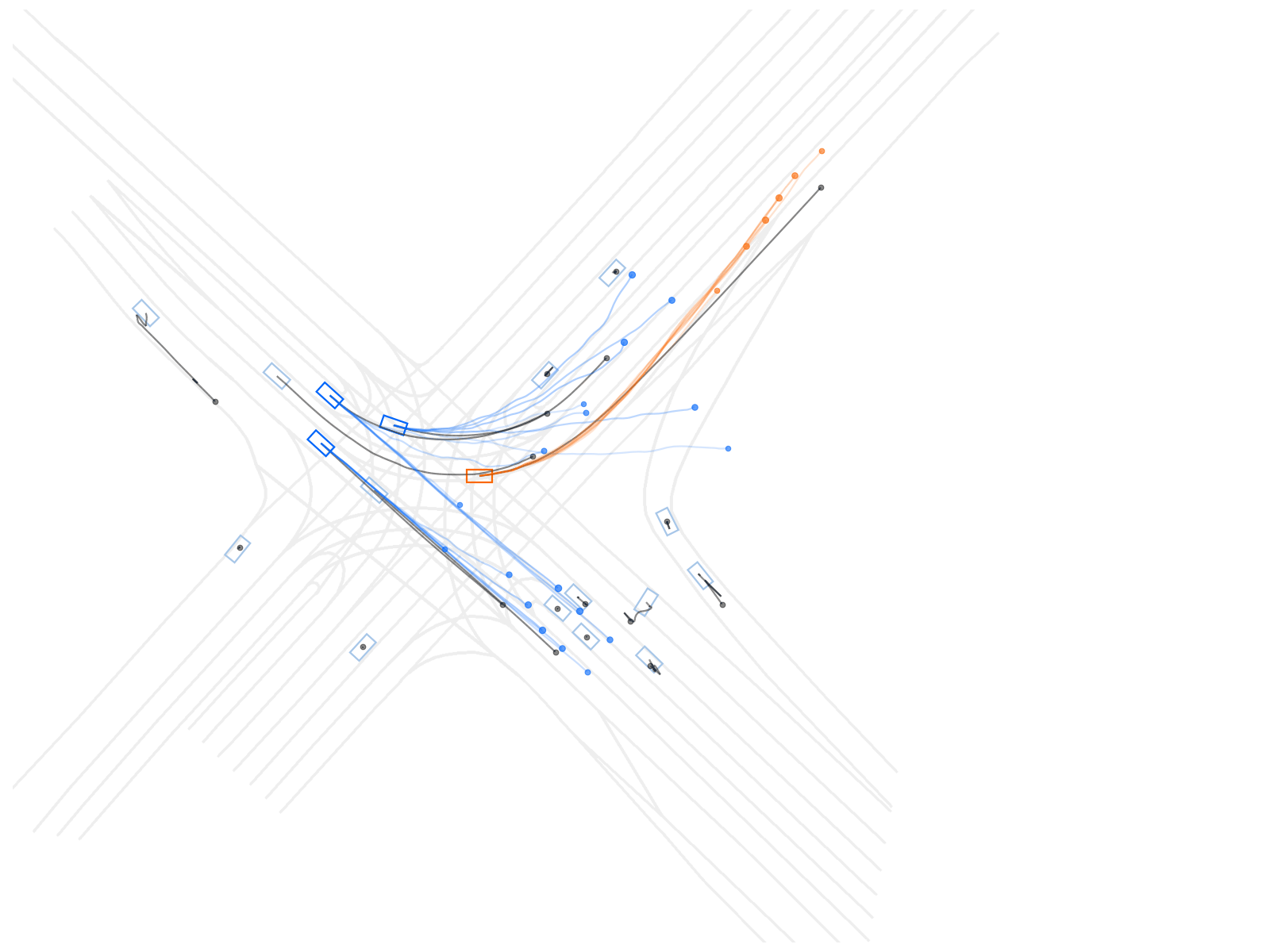} &
        \includegraphics[width=0.245\textwidth, trim={16cm, 16cm, 16cm, 12cm}, clip]{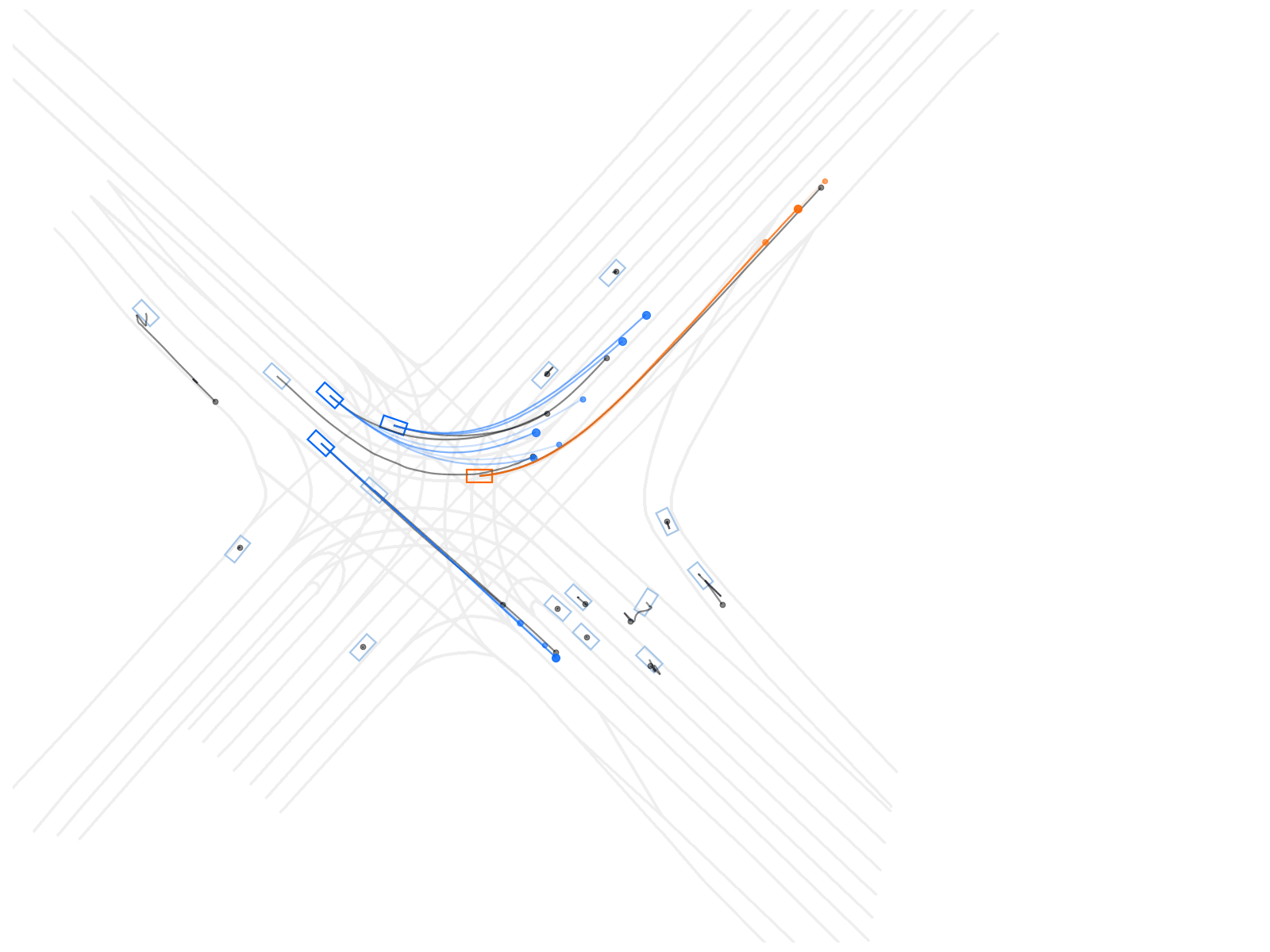} \\
        \midrule
        \includegraphics[width=0.245\textwidth, trim={16cm, 16cm, 16cm, 12cm}, clip]{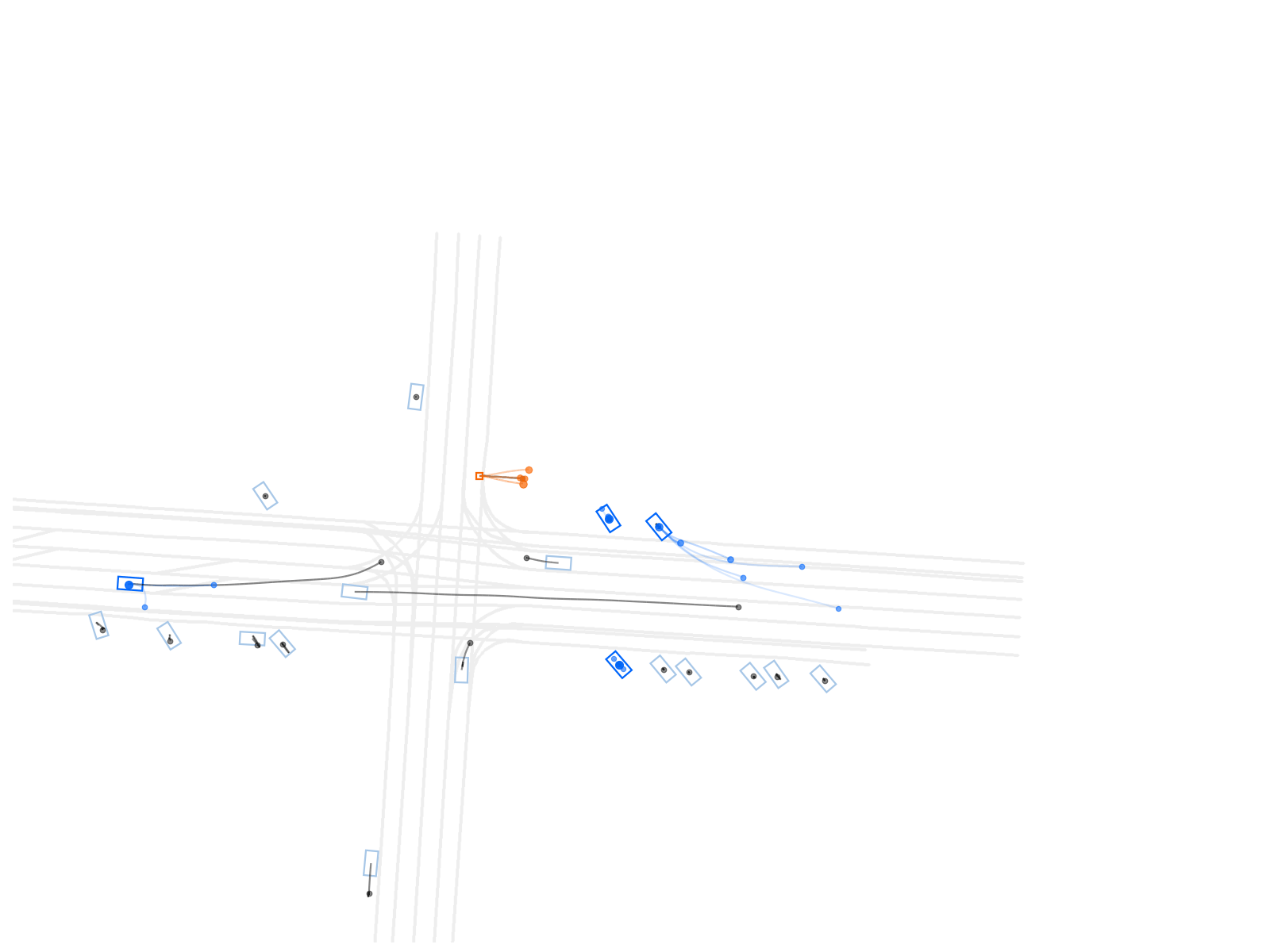} &
        \includegraphics[width=0.245\textwidth, trim={16cm, 16cm, 16cm, 12cm}, clip]{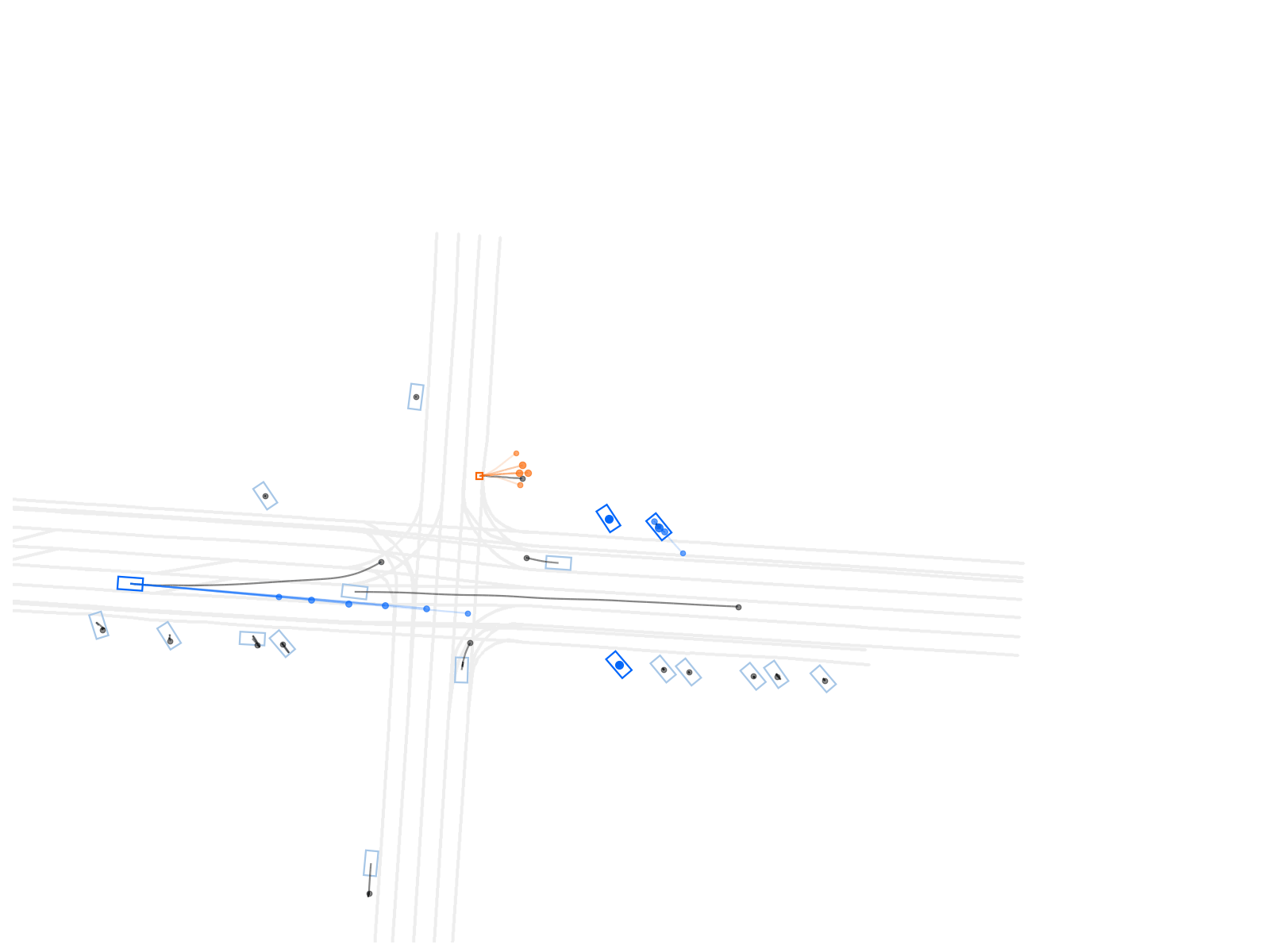} &
        \includegraphics[width=0.245\textwidth, trim={16cm, 16cm, 16cm, 12cm}, clip]{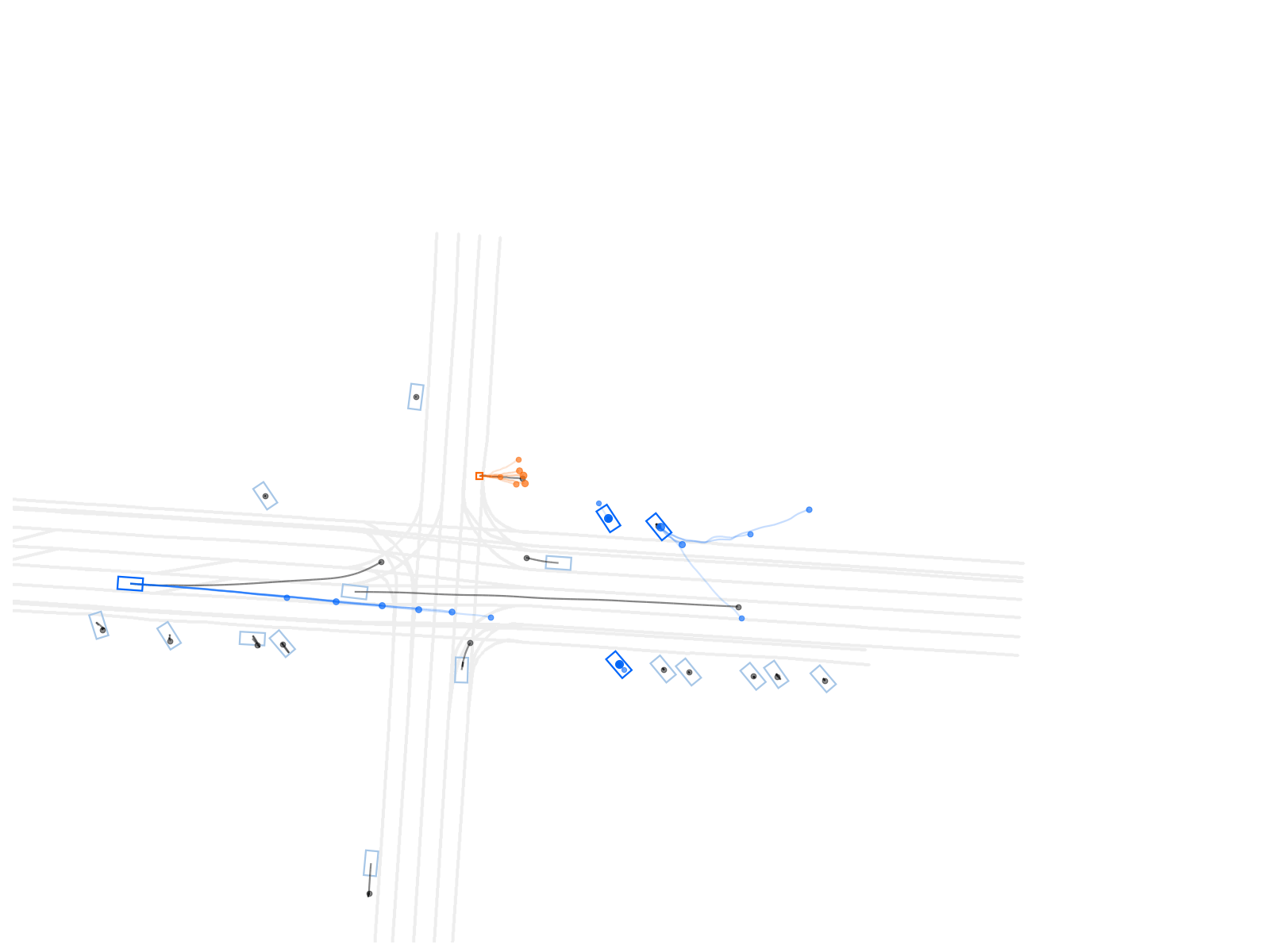} &
        \includegraphics[width=0.245\textwidth, trim={16cm, 16cm, 16cm, 12cm}, clip]{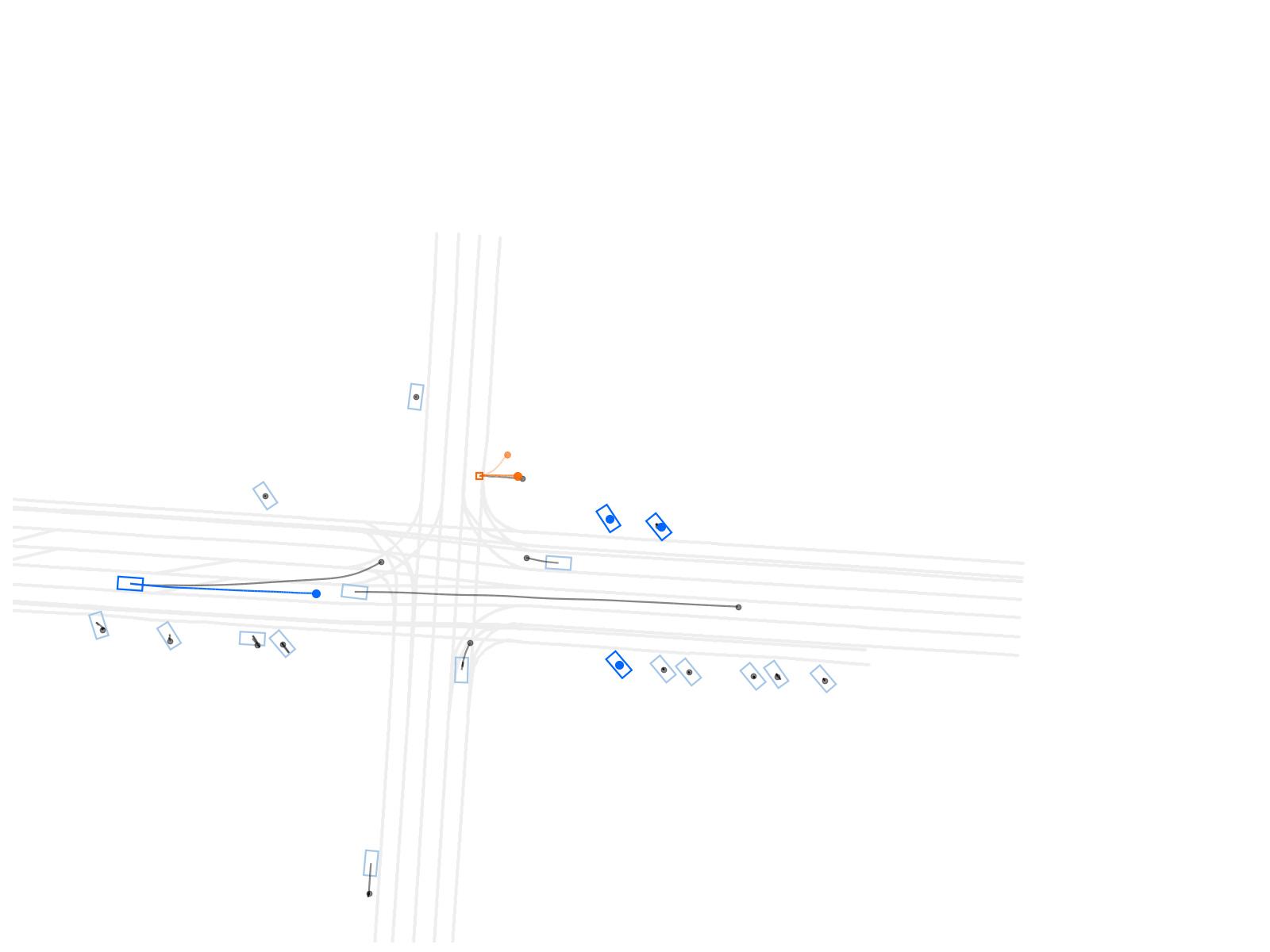} \\
        \midrule
        \includegraphics[width=0.245\textwidth, trim={16cm, 16cm, 16cm, 12cm}, clip]{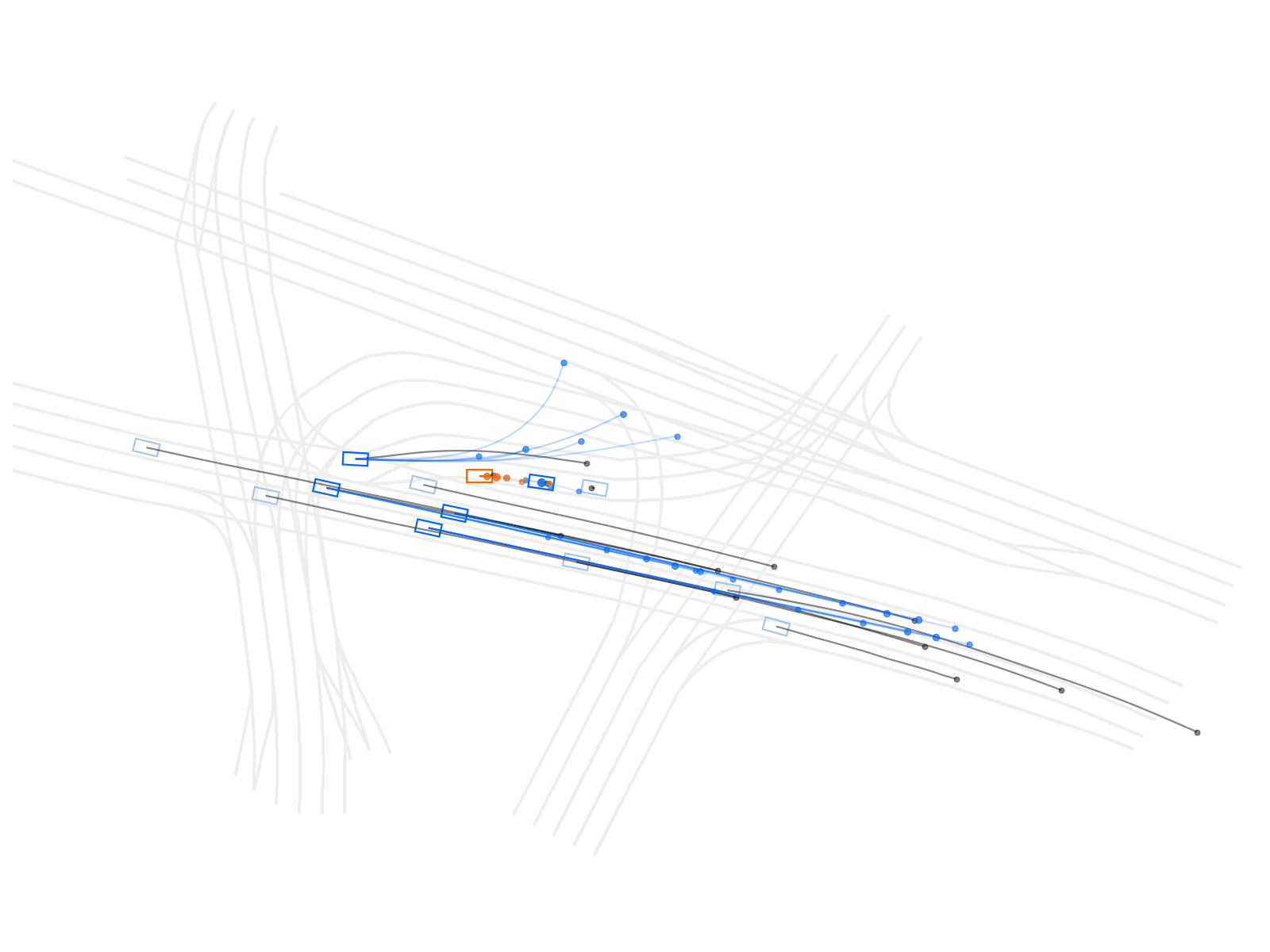} &
        \includegraphics[width=0.245\textwidth, trim={16cm, 16cm, 16cm, 12cm}, clip]{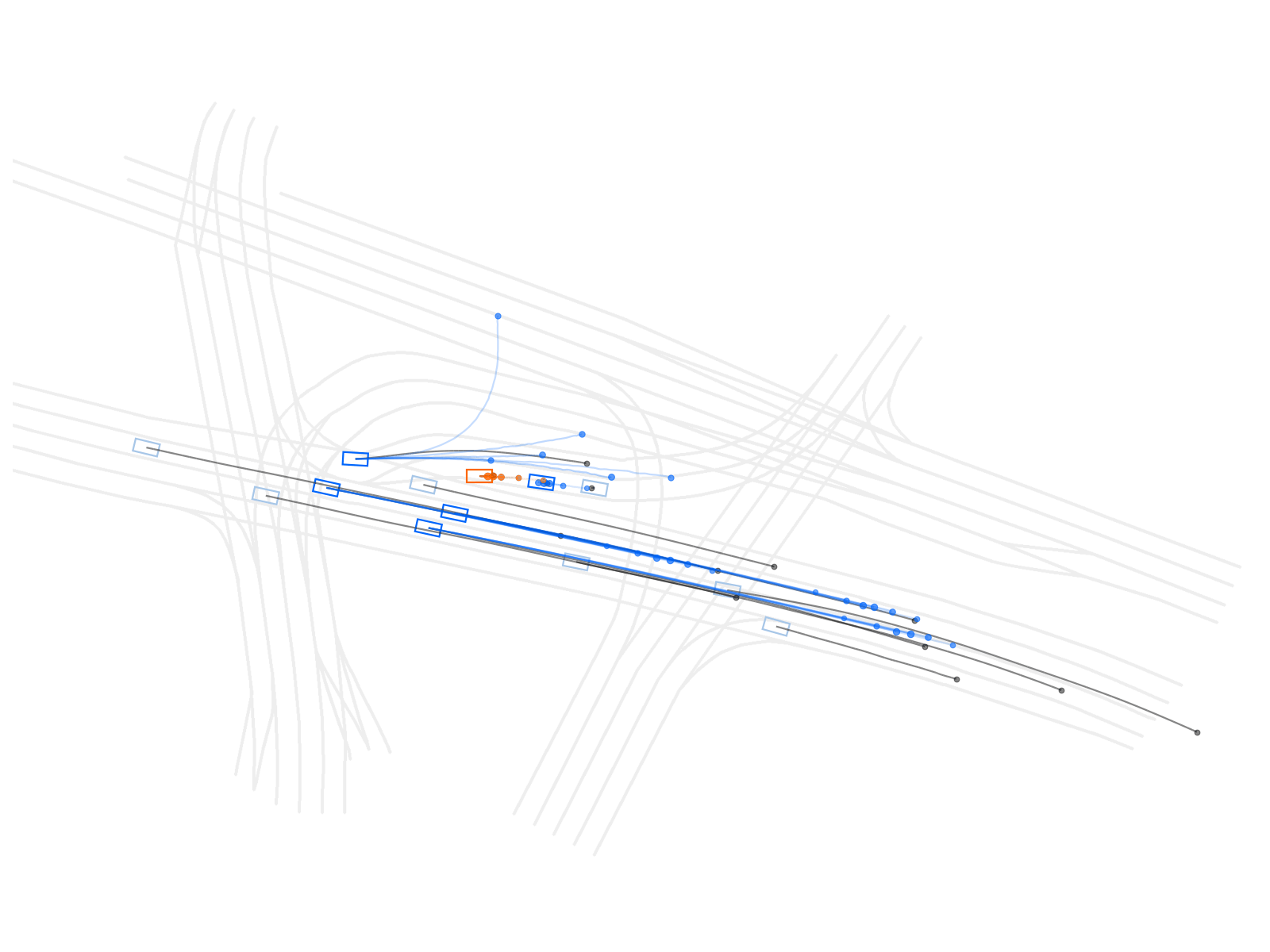} &
        \includegraphics[width=0.245\textwidth, trim={16cm, 16cm, 16cm, 12cm}, clip]{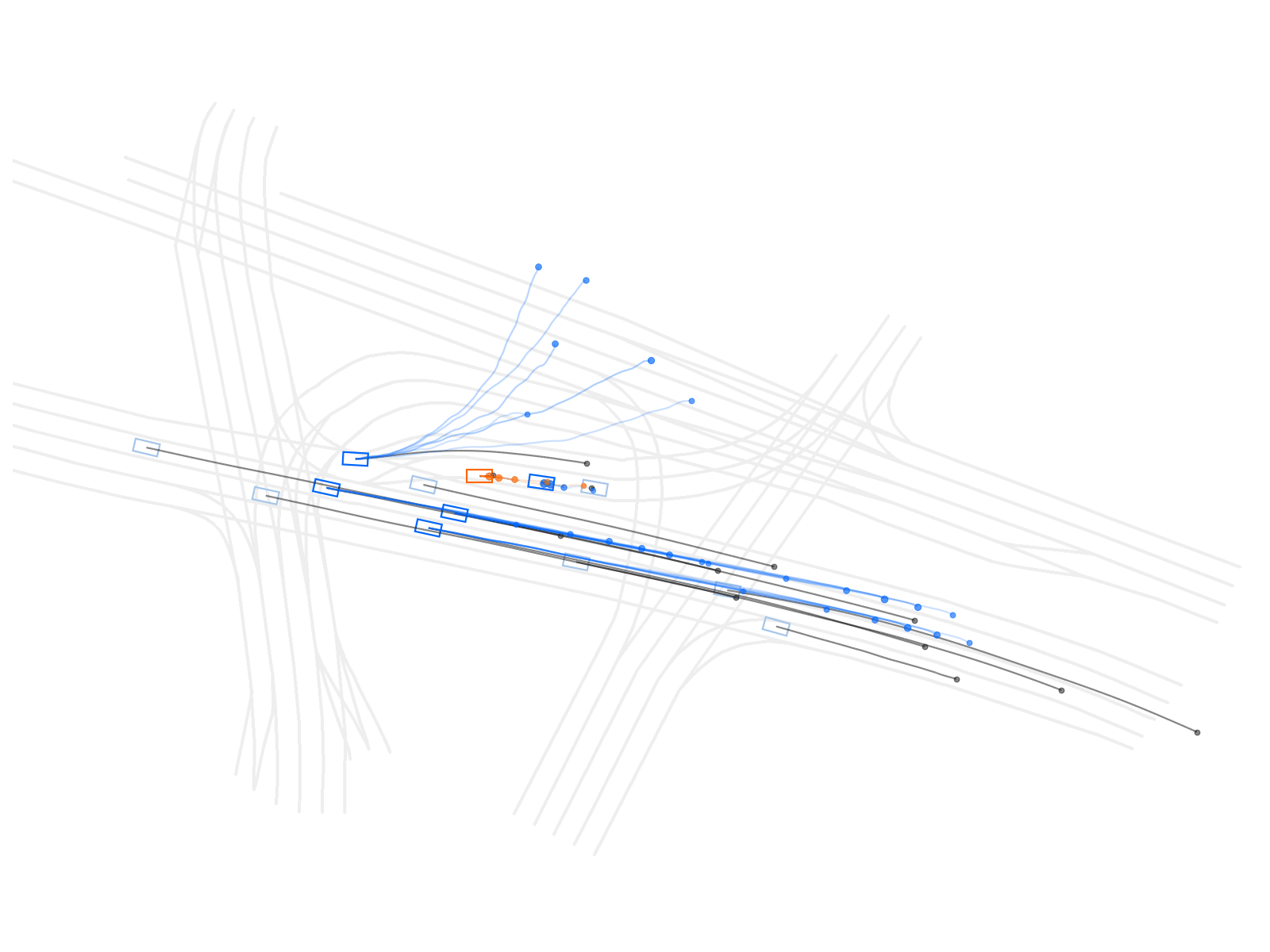} &
        \includegraphics[width=0.245\textwidth, trim={16cm, 16cm, 16cm, 12cm}, clip]{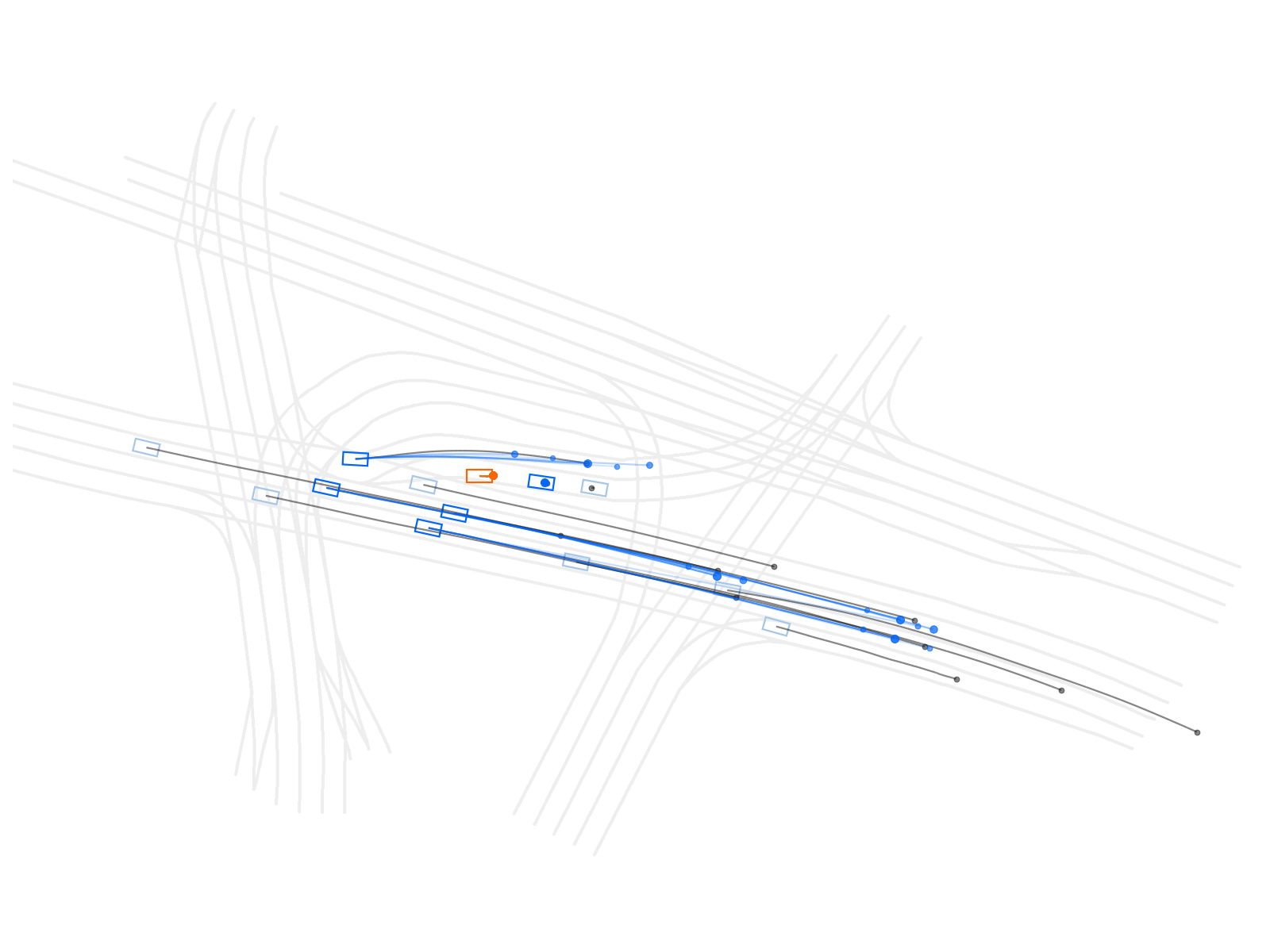} \\
        \midrule
        \includegraphics[width=0.245\textwidth, trim={16cm, 16cm, 16cm, 12cm}, clip]{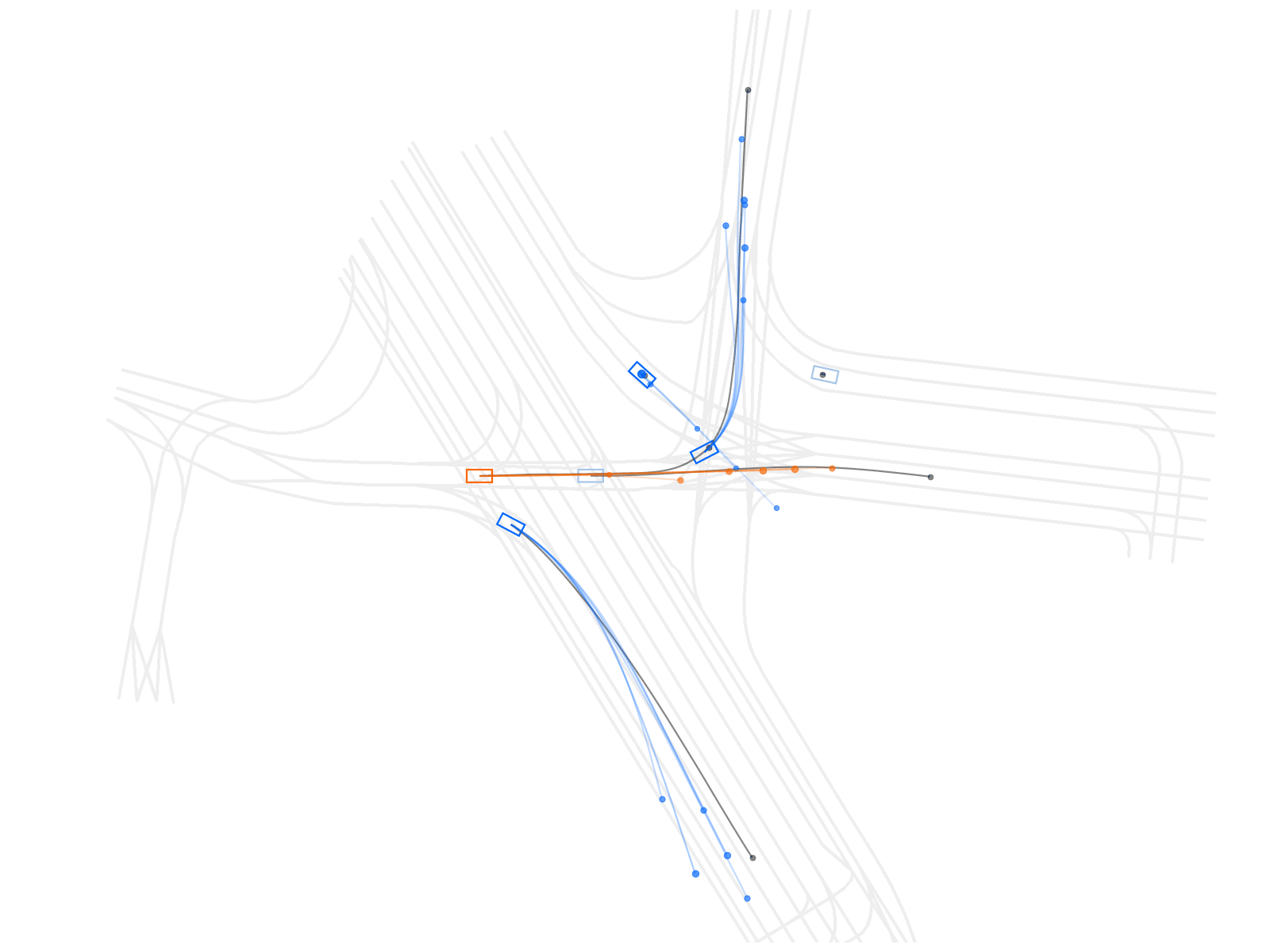} &
        \includegraphics[width=0.245\textwidth, trim={16cm, 16cm, 16cm, 12cm}, clip]{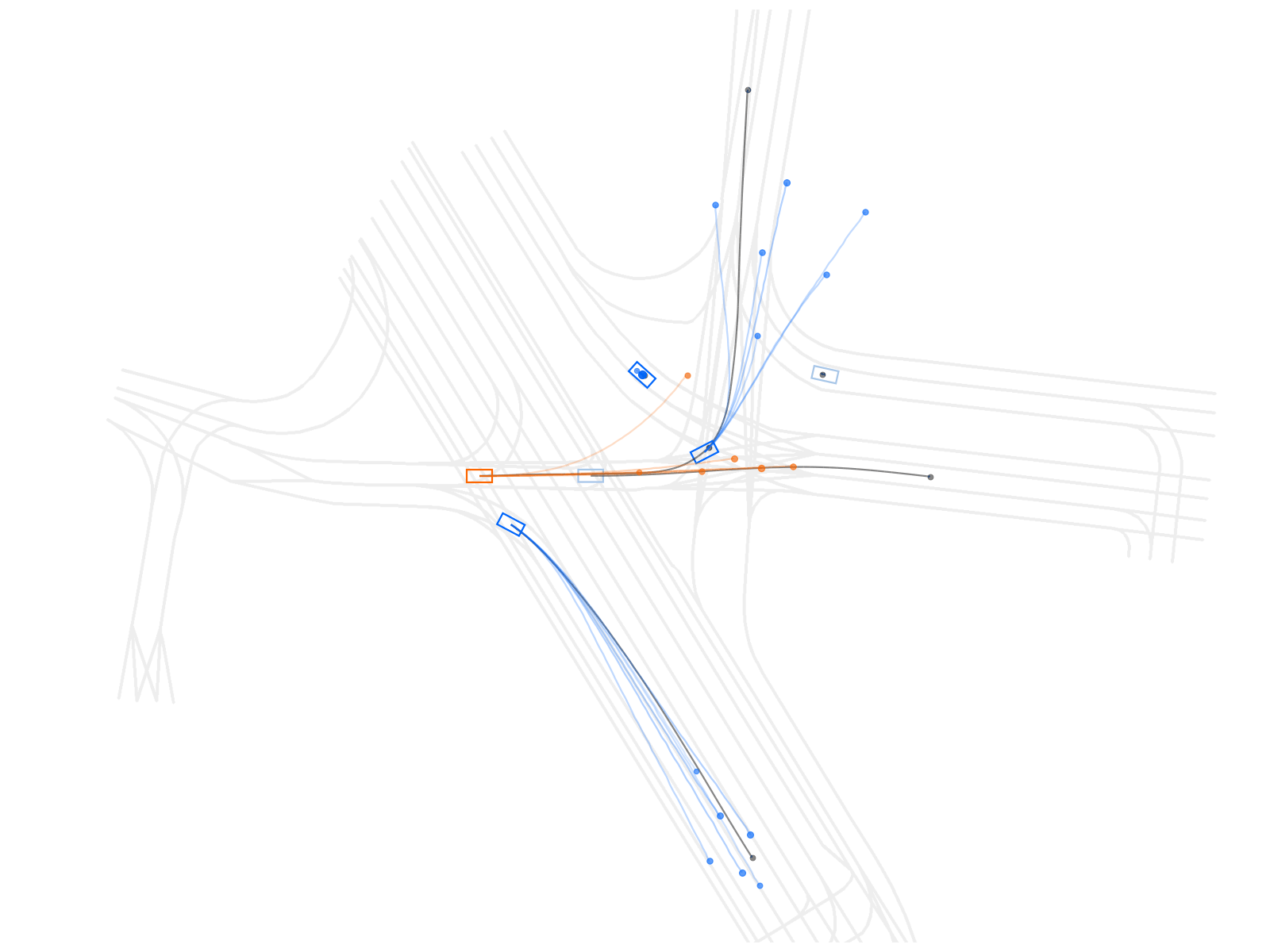} &
        \includegraphics[width=0.245\textwidth, trim={16cm, 16cm, 16cm, 12cm}, clip]{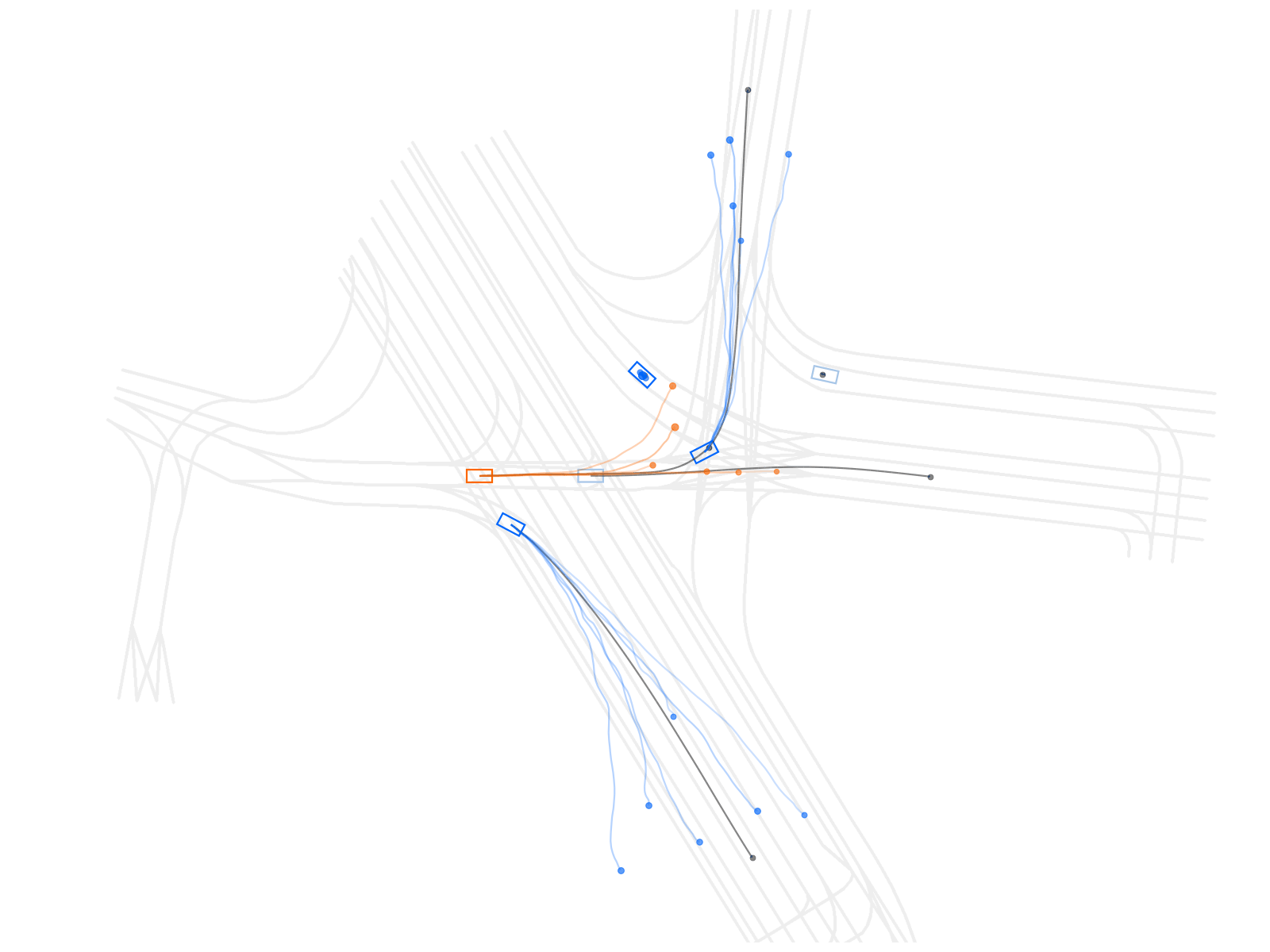} &
        \includegraphics[width=0.245\textwidth, trim={16cm, 16cm, 16cm, 12cm}, clip]{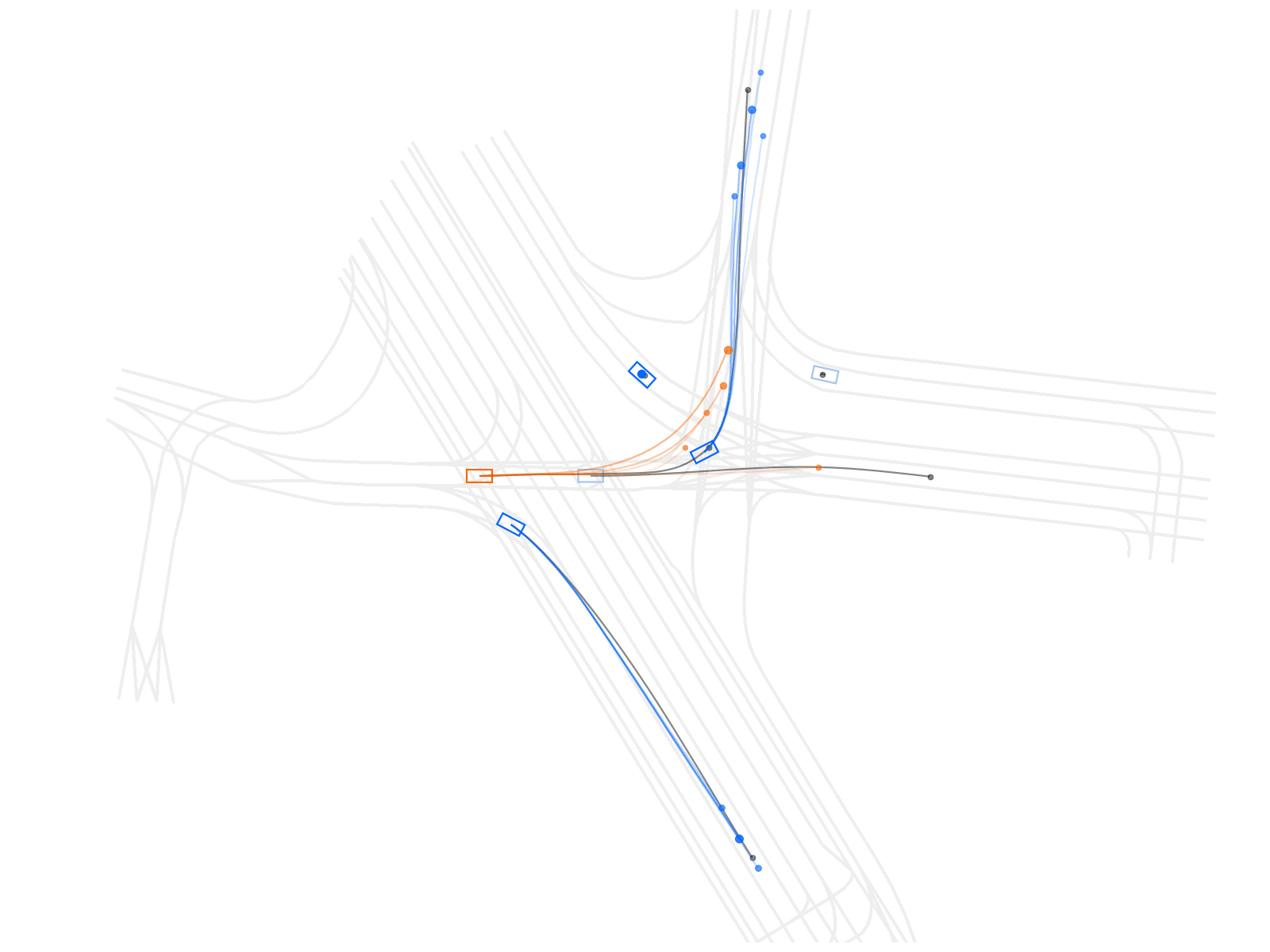} \\
    \end{tabular}
    }
    \caption{\textbf{Qualitative comparison in Argoverse 2}. 6 seconds long multi-modal motion forecasts for {\color{Orange} focal} and {\color{NavyBlue} scored} agents. Ground-truth plotted in gray. Every row showcases a different scenario.}
    \vspace{-10pt}
    \label{fig:qualitative_comparison_argo}
\end{figure*}
\vspace{0.2cm}
\noindent{\bf Qualitative comparisons in Argoverse 2:} Fig. \ref{fig:qualitative_comparison_argo} shows a qualitative comparison between \ourmodel{} and the baselines for 8 different scenarios.
We can clearly see that our model achieves a much better map understanding, showcased by predictions that follow the lanes well. In contrast, the baselines predict some unrealistic modes for most scenarios that are not compliant with the traffic rules and do not align well with typical driving behavior.
As shown by the quantitative experiments in the main paper, our model's best mode attains lower error than the baselines. 
The qualitative results show that our model is able to achieve this while predicting lower diversity modes. 
Given the analysis carried out by \cite{casas2020importance}, which highlights the importance of precision for downstream motion planning, we hypothesize this would be an important characteristic for safe and comfortable driving.

\begin{figure*}[t]
    \centering
    {
    \begin{tabular} {c | c}
        \rotatebox[origin=c]{90}{MTP} & \raisebox{-0.5\height}{\includegraphics[width=0.9\linewidth, trim={40cm, 21cm, 0cm, 24cm}, clip]{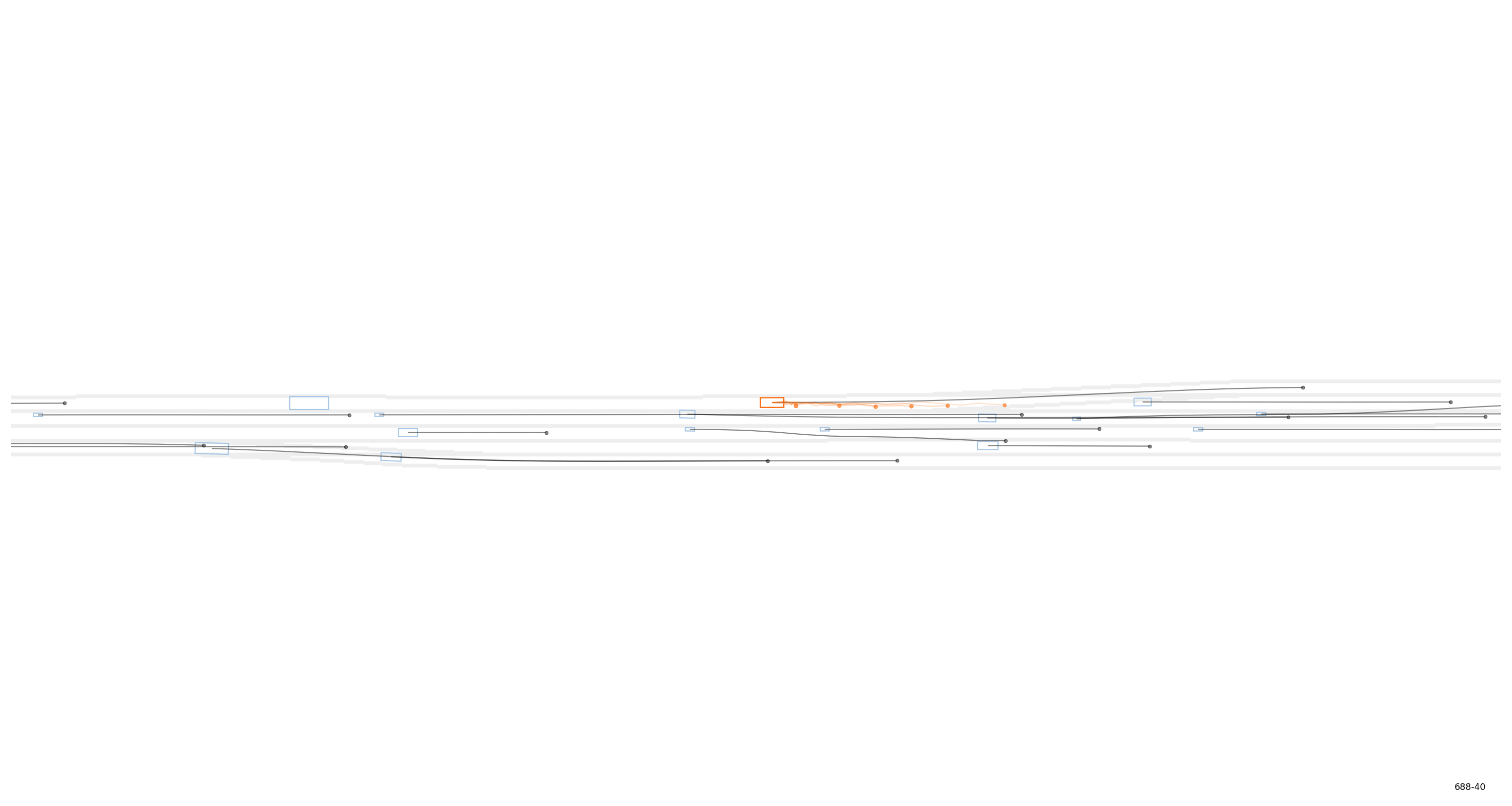}} \\
        \rotatebox[origin=c]{90}{LaneGCN} & \raisebox{-0.5\height}{\includegraphics[width=0.9\linewidth, trim={40cm, 21cm, 0cm, 24cm}, clip]{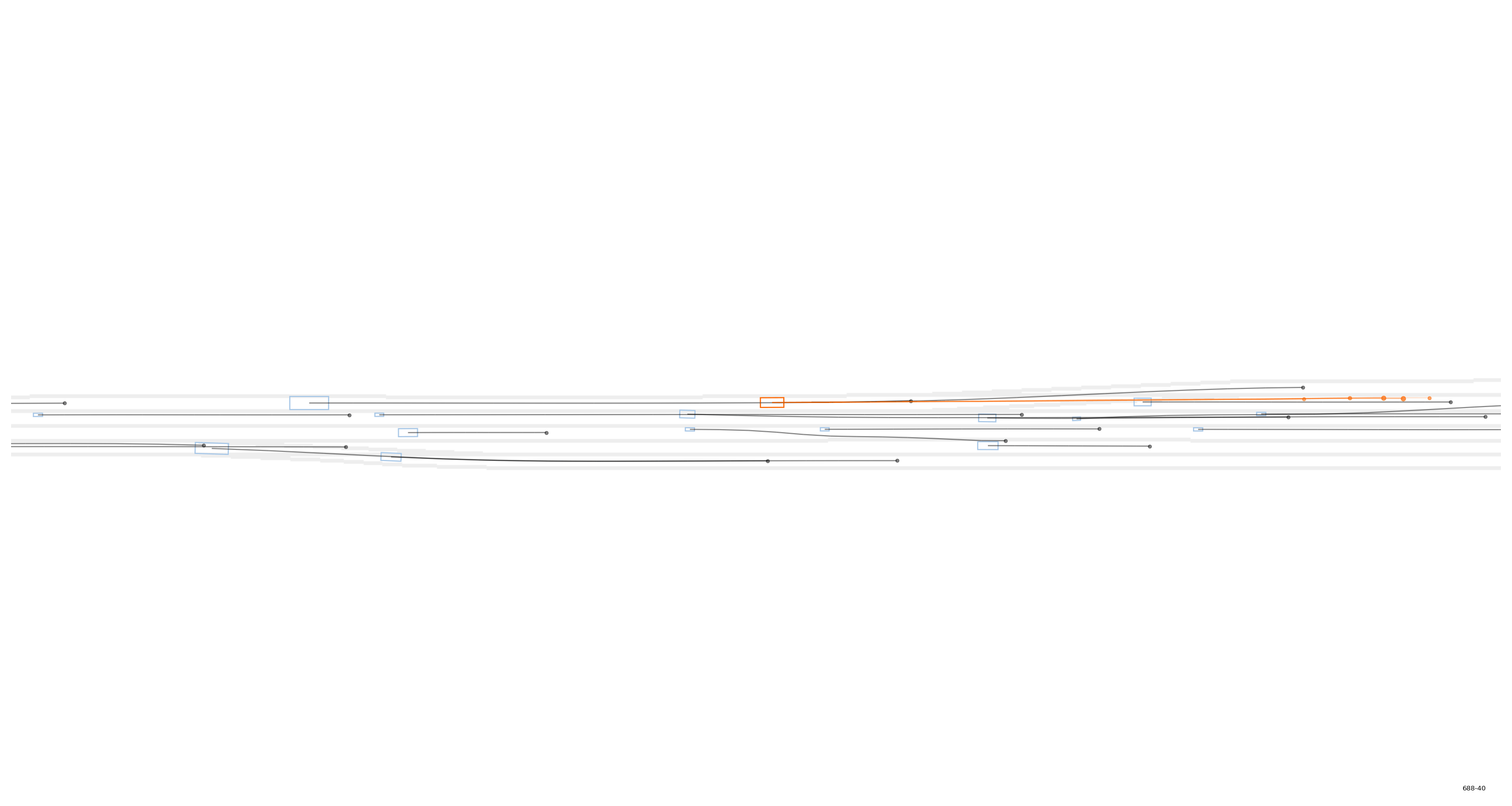}} \\
        \rotatebox[origin=c]{90}{SceneTransformer} & \raisebox{-0.5\height}{\includegraphics[width=0.9\linewidth, trim={40cm, 21cm, 0cm, 24cm}, clip]{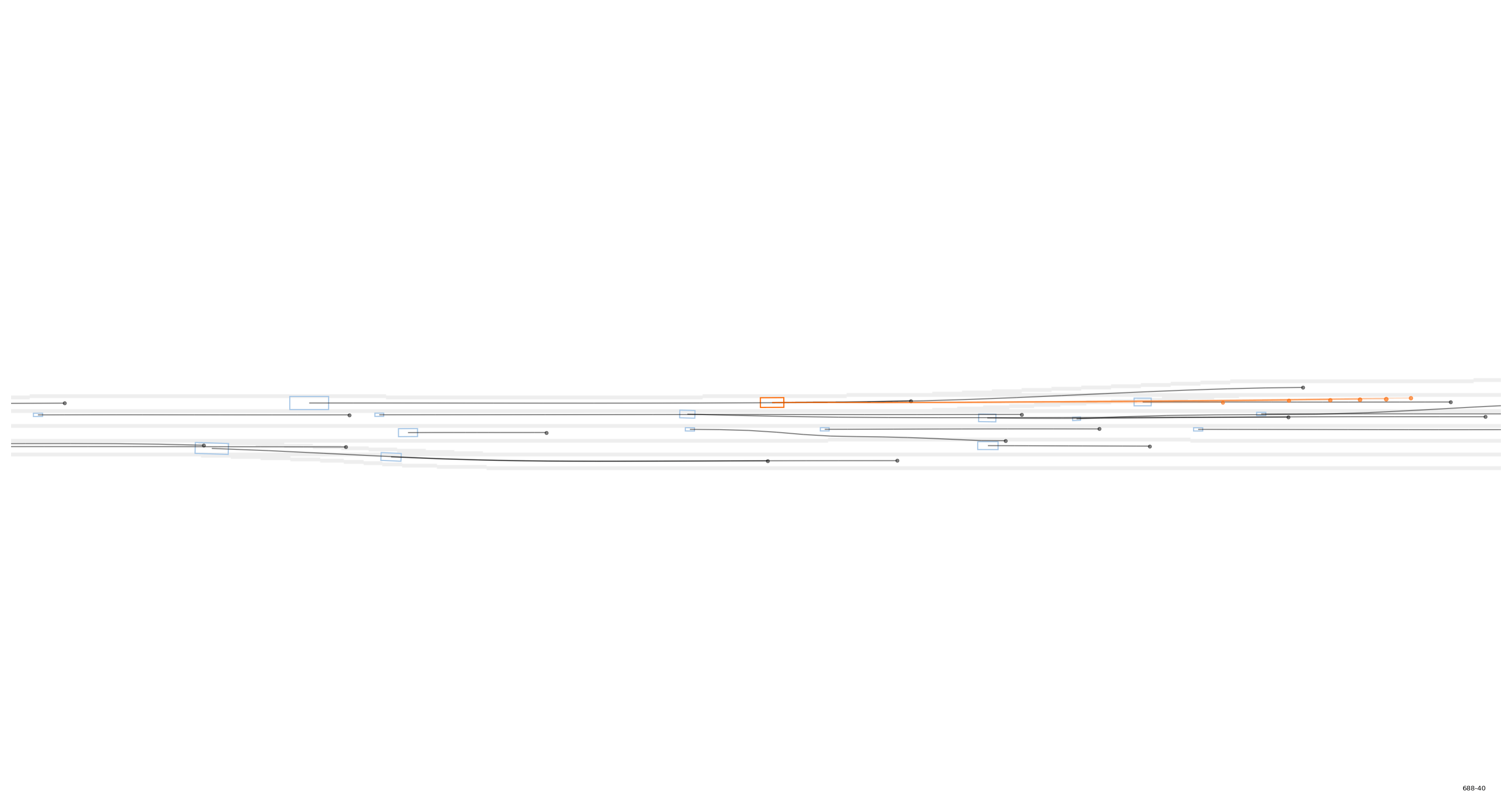}} \\
        \rotatebox[origin=c]{90}{GoRela} & \raisebox{-0.5\height}{\includegraphics[width=0.9\linewidth, trim={40cm, 21cm, 0cm, 24cm}, clip]{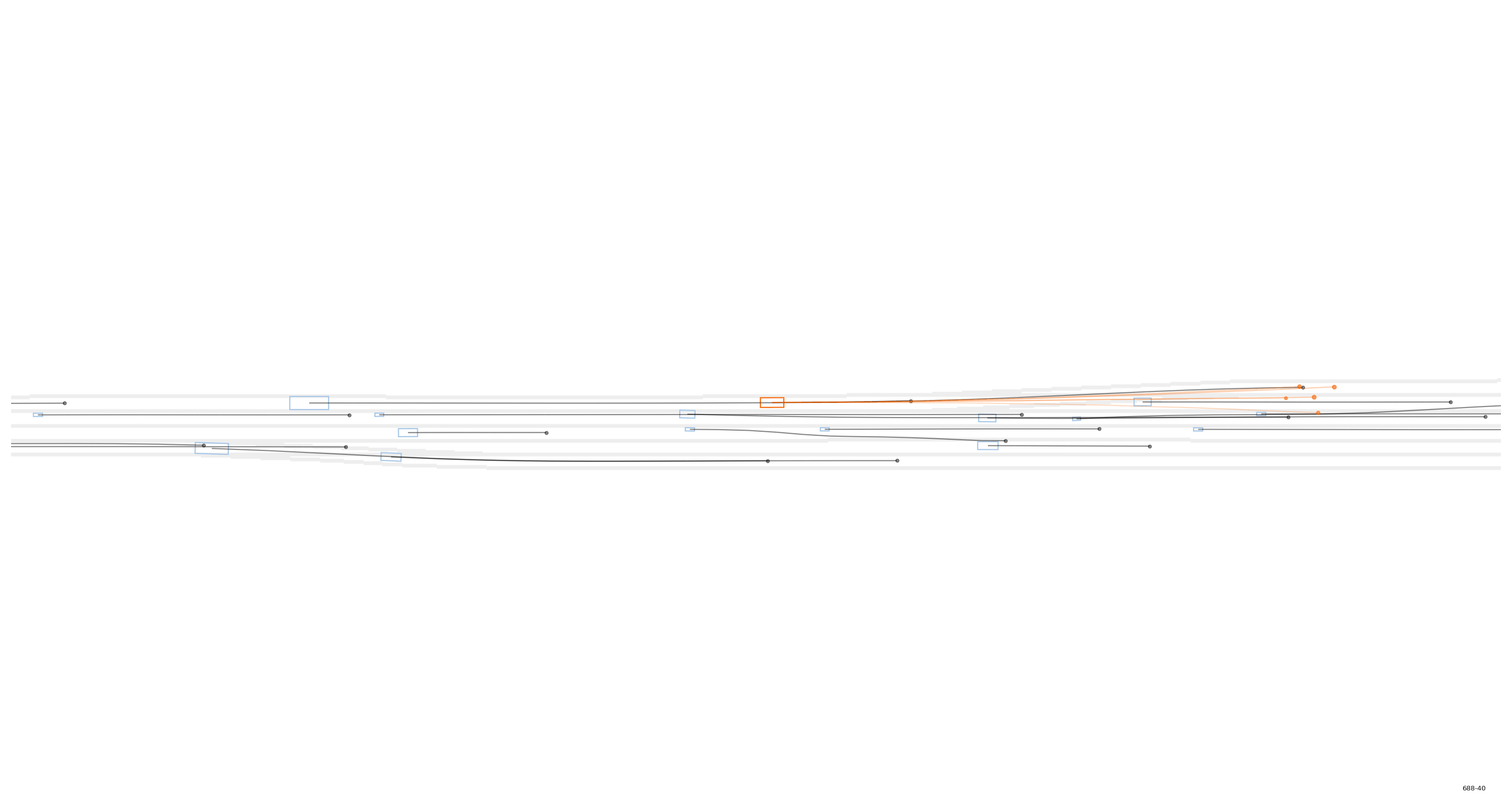}} \\
        \midrule
        \rotatebox[origin=c]{90}{MTP} & \raisebox{-0.5\height}{\includegraphics[width=0.9\linewidth, trim={30cm, 24cm, 10cm, 21cm}, clip]{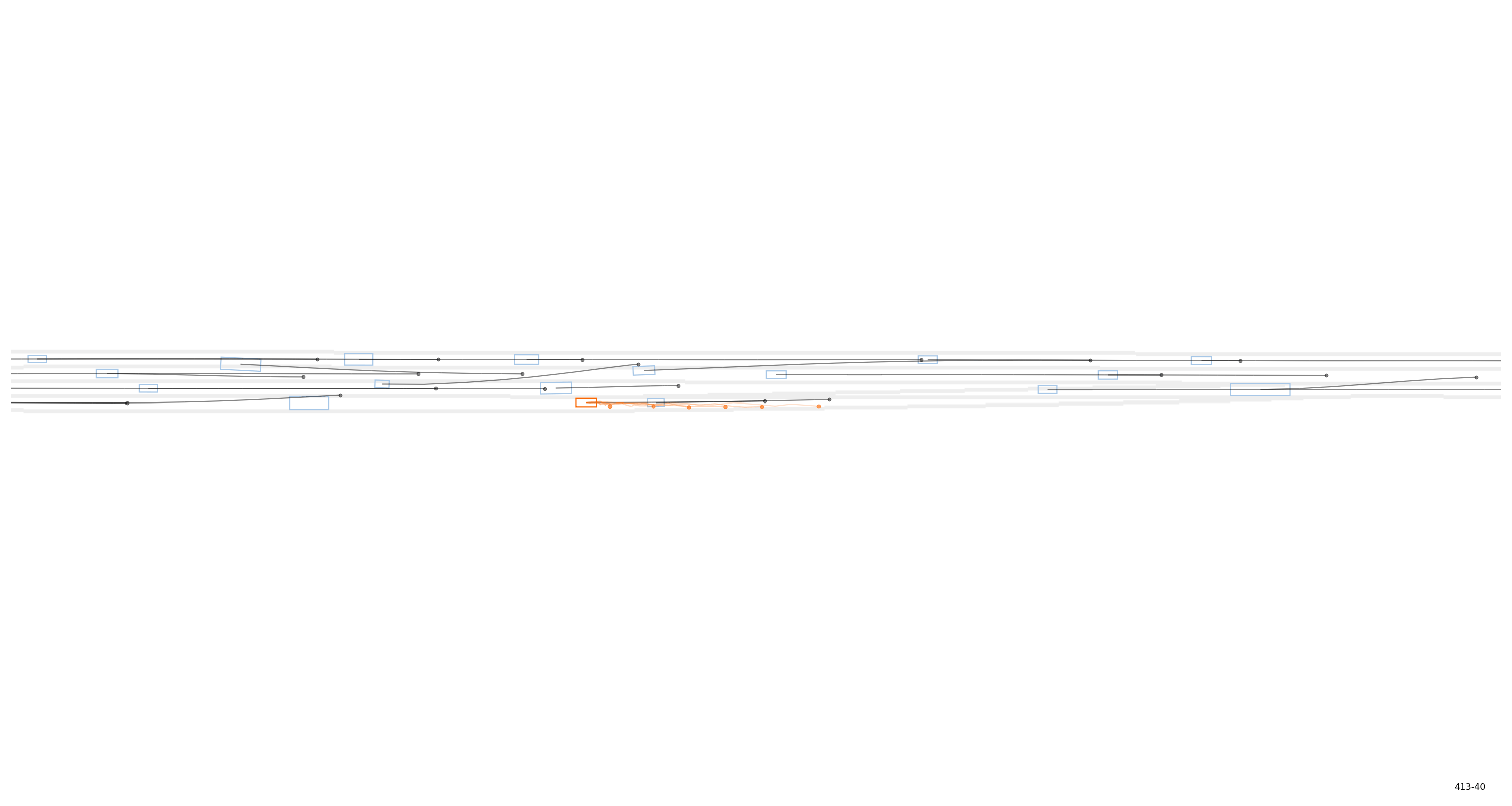}} \\
        \rotatebox[origin=c]{90}{LaneGCN} & \raisebox{-0.5\height}{\includegraphics[width=0.9\linewidth, trim={30cm, 24cm, 10cm, 21cm}, clip]{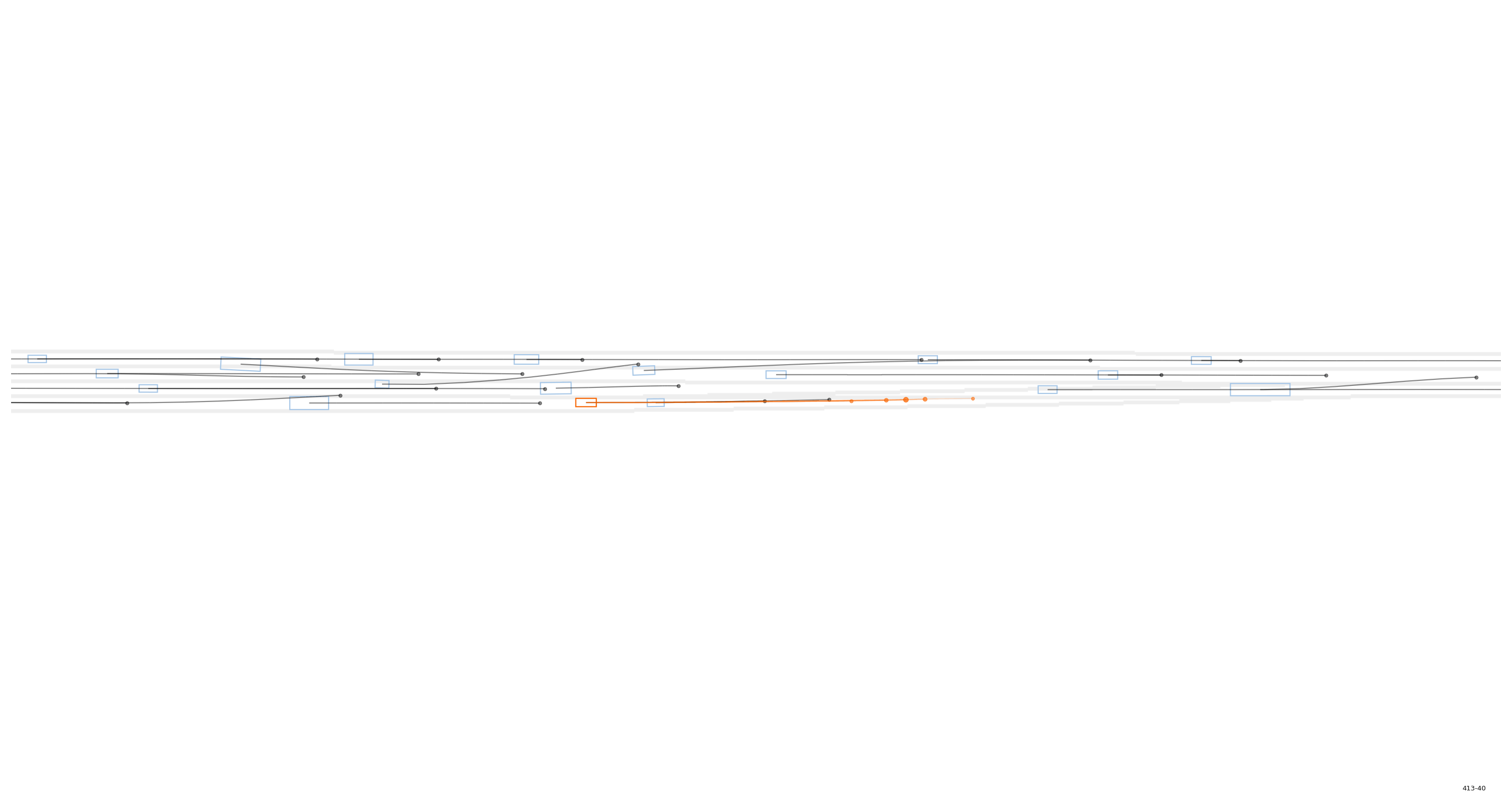}} \\
        \rotatebox[origin=c]{90}{SceneTransformer} & \raisebox{-0.5\height}{\includegraphics[width=0.9\linewidth, trim={30cm, 24cm, 10cm, 21cm}, clip]{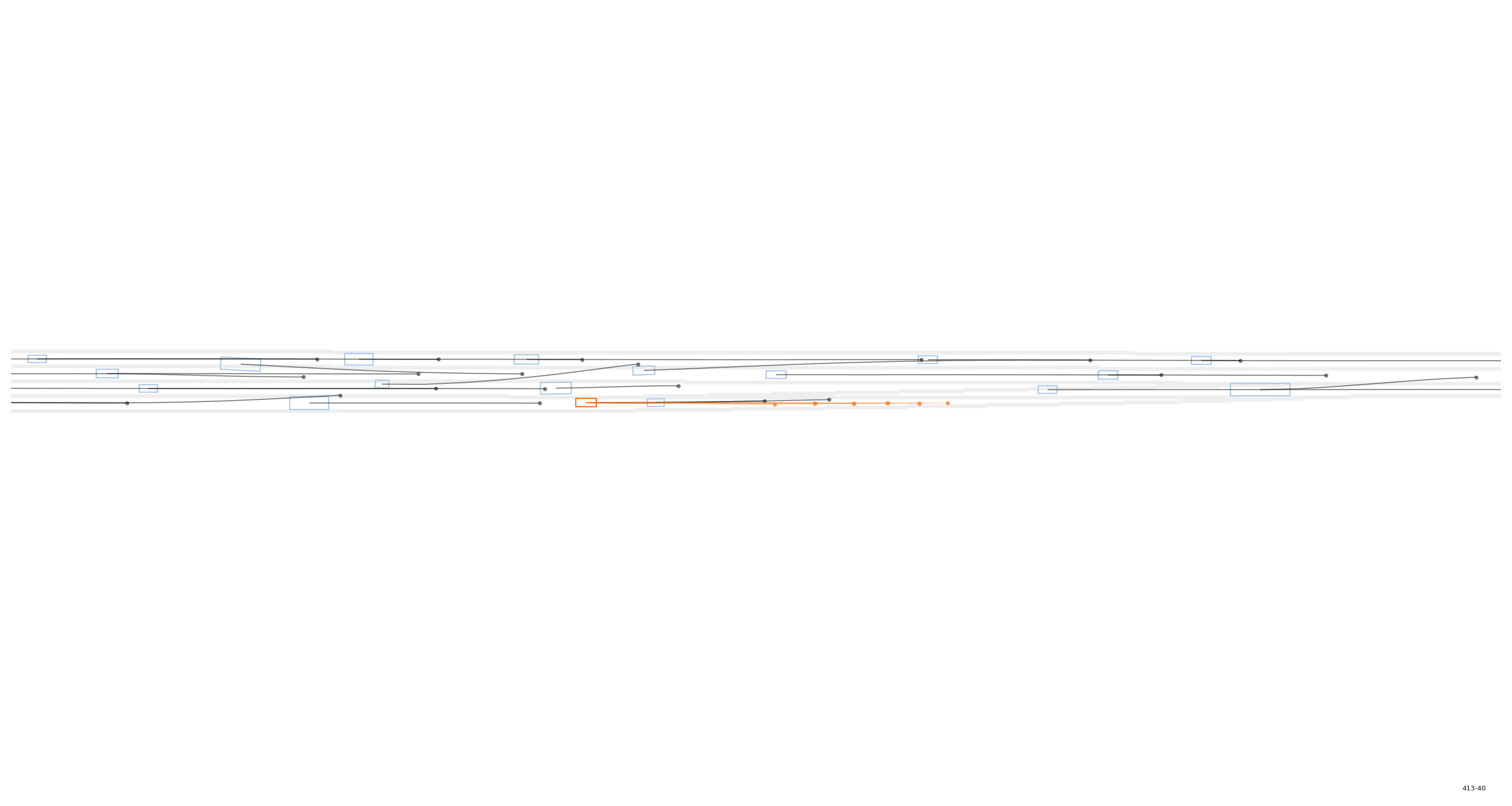}} \\
        \rotatebox[origin=c]{90}{GoRela} & \raisebox{-0.5\height}{\includegraphics[width=0.9\linewidth, trim={30cm, 24cm, 10cm, 21cm}, clip]{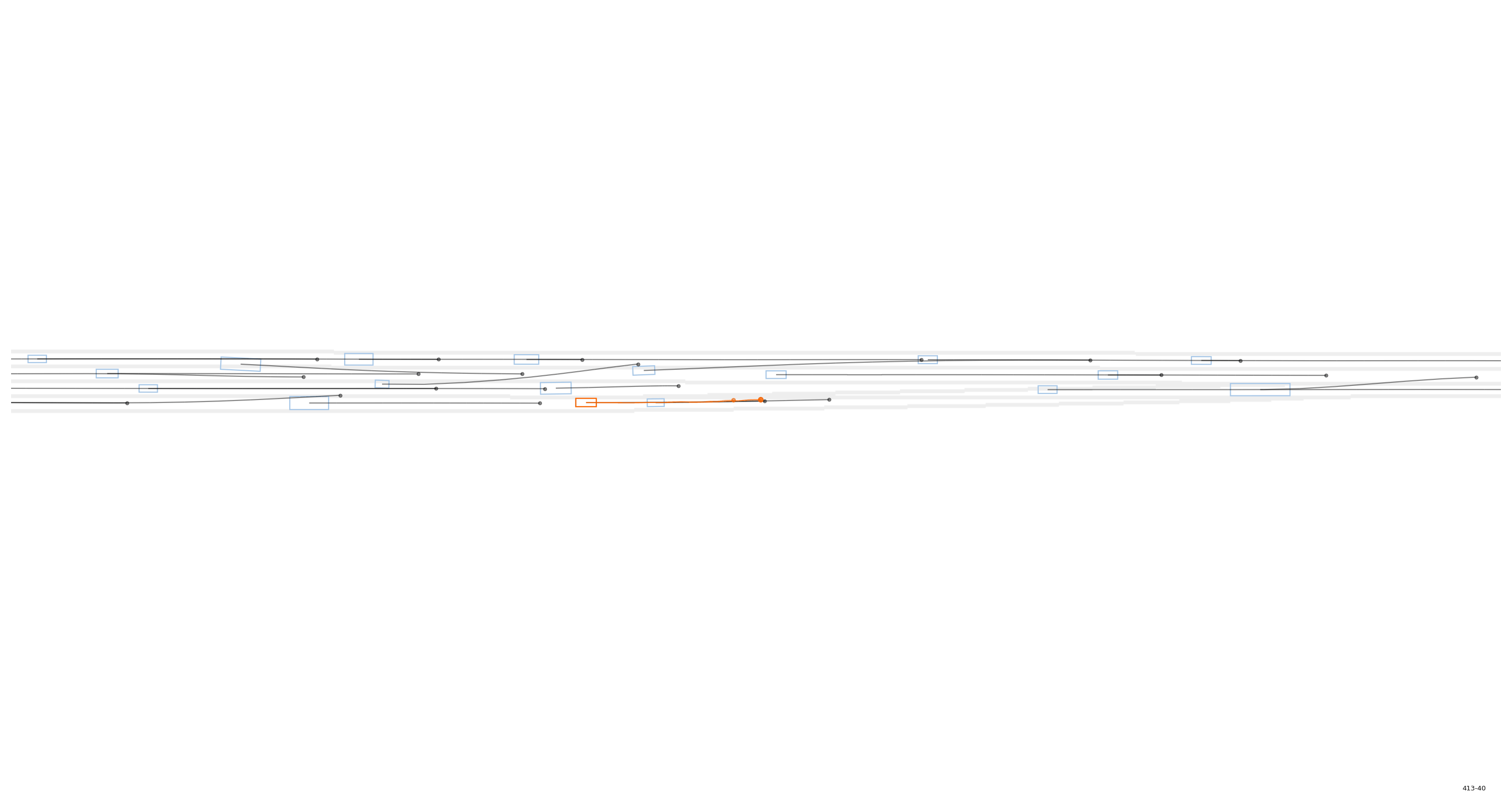}} \\

    \end{tabular}
    }
    \caption{
        \textbf{Qualitative comparison in HighwaySim}. 6 seconds long multi-modal motion forecasts. We only showcase predictions for 1 {\color{Orange} randomly sampled agent} in each scenario since otherwise the visualization is very cluttered due to highway high speeds. Ground-truth plotted in gray. Rows 1-4 correspond to one scenario, and rows 5-8 to another.
    }
    \vspace{-10pt}
    \label{fig:qualitative_comparison_hwysim}
\end{figure*}
\vspace{0.2cm}
\noindent{\bf Qualitative comparisons in HighwaySim:} Fig. \ref{fig:qualitative_comparison_hwysim} displays predictions in highway environments.
In the first scenario, the highlighted agent is taking the fork lane on its left as shown by the ground truth trajectory in gray. All baselines predict the agent is going to continue straight on its original lane. In contrast, \ourmodel{} is able to predict a bi-modal distribution where we believe the agent might lane change or continue straight.
The second scenario highlights an agent on an on-ramp (i.e., a lane merging into the highway), that is following another agent which is yielding to highway traffic.
All baselines do not understand this interaction well enough, and they predict trajectories that collide with the ground truth future motion of the lead vehicle. Our model understands this interaction and is able to predict that the highlighted agent will brake and stay behind its lead vehicle.
\section{Implementation details}
\label{sec:impl_details}

\vspace{0.2cm}
\noindent{\bf Input features:}
For the pair-wise relative positional encoding (PairPose), we use $16$ frequencies for each of sine and cosine.
For each timestep $t$ in agent history encoder inputs, we calculate the PairPose using the pose at $t$ and the pose at the present timestep. We also include the finite difference of the PairPose between $t$ and $t-1$, the velocity, bounding box size and boolean for whether the timestep is observed.
For each map node input in the map encoder, we include the node length, curvature, degree for each edge type, and boolean of whether it's in an intersection and whether it's part of a crosswalk. We also include left and right lane boundary color, type, and distance from the centerline.

\vspace{0.2cm}
\noindent{\bf Training:} For our multi-task objective, we use an equal weight of 1.0 for goal classification, goal regression and trajectory completion. We use $\gamma=2.0$ for the goal classification focal loss, and only supervise the closest node to the ground-truth goal in terms of goal regression.
For the goal classification and regression, our loss ignores those agents which ground-truth goal is not present in the data (labeled track is shorter than the prediction horizon).
For trajectory completion we use the ground-truth goal during training when this is available (teacher forcing). When it is not available in the data, we find it beneficial to use the most likely predicted goal as a weak teacher. 
For our multi-task loss, we do not supervise agents which trajectory does not stay within 10 meters of the lane-graph boundaries at all time steps within the prediction horizon. Agents outside this region are still taken into account in our heterogeneous graphs, but their corresponding predictions do not contribute to the loss. All agents are weighted equally in the loss, including the focal, scored and unscored agents in Argoverse 2.
Since our model is invariant to rotations and translations, we do not require such augmentations during training. However, we find useful to perform a scale augmentation by uniformly sampling the scaling factor between 0.8 and 1.2.
Our final model was trained for 17 epochs, using 16 GPUs for a total batch size of 64. We use Adam optimizer with a learning rate 1e-5, with a step-wise scheduler with 0.25 decay and 15 step-size. 

\vspace{0.2cm}
\noindent{\bf Greedy goal sampler:} 
We use hard threshold radius $\gamma = 2$, downweight radius $\nu = 4$, and downweight factor $\tau = 10$.

\end{document}